\documentclass{article}

\usepackage{arxiv}
\usepackage{t1enc}
\usepackage{hyperref} 
\usepackage{enumerate}
\usepackage{graphicx}
\usepackage{amsmath} 
\usepackage[english]{babel}
\usepackage[utf8]{inputenc,xcolor}
\usepackage{tcolorbox}
\usepackage{fancyhdr}
\usepackage{tikz}
\usepackage{multicol}
\usepackage{booktabs}
\usepackage{multirow}
\usepackage{siunitx}
\usepackage{adjustbox}
\usepackage{chngpage}
\usepackage{float}
\usepackage{caption}
\usepackage{subcaption}
\usepackage{comment}
\usepackage{textcomp}
\usepackage{siunitx}
\usepackage{abraces}
\usepackage[linesnumbered,ruled,vlined]{algorithm2e}
\usepackage{listings}

\usepackage[utf8]{inputenc} 
\usepackage[T1]{fontenc} 
\usepackage{hyperref} 
\usepackage{url} 
\usepackage{booktabs} 
\usepackage{amsfonts} 
\usepackage{nicefrac} 
\usepackage{microtype} 
\usepackage{xcolor} 

\usepackage[utf8]{inputenc} 
\usepackage[T1]{fontenc} 
\usepackage{hyperref} 
\usepackage{url} 
\usepackage{booktabs} 
\usepackage{amsfonts} 
\usepackage{nicefrac} 
\usepackage{microtype} 
\usepackage{amsfonts,mathtools}
\usepackage{array}

\usepackage{graphicx}
\usepackage{mathrsfs}

\usepackage{lipsum}

\usepackage{etoolbox}
\usepackage{amsmath,bm}
\usepackage{color}
\usepackage{float}
\usepackage{epstopdf}
\usepackage{amssymb}

\usepackage{graphicx}
\usepackage{sidecap}

\usepackage{amsthm}
\usepackage[utf8]{inputenc}
\usepackage{url}
\usepackage{breqn}

\newsavebox{\foobox}

\renewcommand{\b}[1]{\ensuremath{\mathbf{#1}}}		 							
\newcommand{\norm}[1]{\ensuremath{\left\|#1\right\|}}						

\providecommand{\tr}[1]{\text{tr}\left(#1\right)}

\newcommand{\arrowAS}[1]{\ensuremath{\stackrel{\text{a.s.}}{\longrightarrow}}}

\providecommand{\norm}[1]{\left \| #1 \right \|}

\providecommand{\norm}[1]{\left \| #1 \right \|}

\theoremstyle{plain} 
\newtheorem{thm}{Theorem}

\newtheorem{Lem1}{Lemma}
\newtheorem{def1}{Definition}

\theoremstyle{definition} 

\theoremstyle{remark} 

\newtheorem{rem}{\bf Remark}

\usepackage{blindtext}

\SetAlCapSkip{1em}

\SetKwInput{KwInput}{Input}
\SetKwInput{KwOutput}{Output}

\def \Rn {{\mathbb{R}}}
\def \x {{\b{x}}}

\def \y {{\b{y}}}

\def \C {{\b{C}}}

\def \G {{\mathcal{G}}}

\def \cX {{\mathcal{X}}}

\def \S {{\mathcal{S}}}

\def \bzero {{\b{0}}}
\def \bone {{\b{1}}}

\begin{document}
\title{A Unified Framework for Optimization-Based Graph Coarsening
}

\author{
  Manoj Kumar\\
  Department of Electrical Engineering \\
  Indian Institute of Technology Delhi \\
  New Delhi, 110016, India\\
  \texttt{eez208646@iitd.ac.in} \\
   \And
   Anurag Sharma\\
  Department of Mathematics \\
   Indian Institute of Technology Delhi \\
   New Delhi, 110016, India\\
  \texttt{mt6190745@iitd.ac.in} \\
  \And
   Sandeep Kumar\\
  Department of Electrical Engineering \\
  Yardi School of Artificial Intelligence\\
 Bharti School of Telecommunication Technology and Management\\
   Indian Institute of Technology Delhi \\
   New Delhi, 110016, India\\
  \texttt{ ksandeep@iitd.ac.in} \\
}



 \maketitle

\begin{abstract}
Graph coarsening is a widely used dimensionality reduction technique for approaching large-scale graph machine learning problems. Given a large graph, graph coarsening aims to learn a smaller-tractable graph while preserving the properties of the originally given graph. Graph data consist of node features and graph matrix (e.g., adjacency and Laplacian). The existing graph coarsening methods ignore the node features and rely solely on a graph matrix to simplify graphs. In this paper, we introduce a novel optimization-based framework for graph dimensionality reduction. The proposed framework lies in the unification of graph learning and dimensionality reduction. It takes both the graph matrix and the node features as the input and learns the coarsen graph matrix and the coarsen feature matrix jointly while ensuring desired properties. The proposed optimization formulation is a multi-block non-convex optimization problem, which is solved efficiently by leveraging block majorization-minimization, $\log$ determinant, Dirichlet energy, and regularization frameworks. The proposed algorithms are provably convergent and practically amenable to numerous tasks. It is also established that the learned coarsened graph is $\epsilon\in(0,1)$ similar to the original graph. Extensive experiments elucidate the efficacy of the proposed framework for real-world applications.



\end{abstract}

\section{Introduction}

Graph-based approaches with big data and machine learning are one of the strongest driving forces of the current research frontiers, creating new possibilities in a variety of domains from social networks to drug discovery and from finance to material science studies.
 Large-scale graphs are becoming increasingly common, which is exciting since more data implies more knowledge and more training sets for learning algorithms. However, the graph data size is the real bottleneck, handling large graph data involves considerable computational hurdles to process, extract, and analyze graph data. Therefore, graph dimensionality reduction techniques are needed. 
 
 In classical data analysis over Euclidean space, there exist a variety of data reduction techniques, e.g., compressive sensing\cite{yankelevsky2016dual}, low-rank approximation\cite{kishore2017literature}, metric preserving dimensionality reduction\cite{celik2014efficient}, but such techniques for graph data have not been well understood yet. Graph coarsening or graph summarization is a promising direction for scaling up graph-based machine learning approaches by simplifying large graphs. Simply, coarsening aims to summarize a very large graph into a smaller and tractable graph while preserving the properties of originally given graph. The core idea of coarsening comes from the algebraic multi-grid literature \cite{ruge1987algebraic}. Coarsening methods have been applied in various applications like graph partitioning \cite{hendrickson1995multi, karypis1998fast,kushnir2006fast,dhillon2007weighted}, machine learning \cite{lafon2006diffusion, gavish2010multiscale, shuman2015multiscale}, and scientific computing\cite{chen2022graph,hackbusch2013multi,ruge1987algebraic,briggs2000multigrid}. The recent work in \cite{loukas2019graph} developed a set of frameworks for graph matrix coarsening preserving spectral and cut guarantees, and \cite{cai2021graph} developed a graph neural network based method for graph coarsening.

A graph data consist of node features and node connectivity matrix also known as graph matrix e.g., adjacency or Laplacian Matrix \cite{kipf2017semi, zugner2019adversarial, wang2019heterogeneous}. The caveat of the existing graph coarsening methods is that they completely ignore the node features and rely solely on the graph matrix of given graph data \cite{loukas2018spectrally,loukas2019graph,bravo2019unifying,purohit2014fast,chen2022graph}. The quality of the graph is the most crucial aspect for any graph-based machine learning application. Ignoring the node features while coarsening the graph data would be inappropriate for many applications. For example, many real-world graph data satisfy certain properties, e.g., homophily assumption and smoothness \cite{unknown, pmlr-v51-kalofolias16}, that if two nodes are connected with stronger weights, then the features corresponding to these nodes should be similar. Implying, if the original graph satisfies any property, then that property should translate to the coarsen graph data. Current methods can only preserve spectral properties which indicates the property of the graph matrix but not the node features \cite{loukas2018spectrally,loukas2019graph}. And hence these are not suitable for many downstream real-world applications which require node features along with graph matrix information.

%


 We introduce a novel optimization-based framework lying at the unification of graph learning \cite{kumar2020unified,NEURIPS2019_90cc440b} and dimensionality reduction \cite{qiu2017deep, zhu2017subspace} for coarsening graph data, named as featured graph coarsening (FGC). It takes both the graph matrix and the node features as the input and learns the coarsen graph matrix and the coarsen feature matrix jointly while ensuring desired properties. The proposed optimization formulation is a multi-block non-convex optimization problem, which is solved efficiently by leveraging block majorization-minimization, $\log$ determinant, Dirichlet energy, and regularization frameworks. The developed algorithm is provably convergent and enforces the desired properties, e.g., spectral similarity and $\epsilon$-similarity in the learned coarsened graph. Extensive experiments elucidate the efficacy of the proposed framework for real-world applications.


\subsection{Summary and Contribution}

In this paper, we have introduced a novel optimization-based framework for graph coarsening, which considers both the graph matrix and feature matrix jointly. 
Our major contributions are summarized below:
\begin{itemize}
 \item[(1)] We introduce a novel optimization-based framework for graph coarsening by approaching it at the unification of dimensionality reduction and graph learning. We have proposed three problem formulations:
\begin{itemize}
 \item[(a)] \textbf{Featured graph coarsening}\\
This formulation uses both graph topology (graph matrix) and node features (feature matrix) jointly and learns a coarsened graph and coarsened feature matrix. The coarsened graph preserves the properties of the original graph.
 \item[(b)] \textbf{Graph coarsening without features}\\
 This formulation uses only a graph matrix to perform graph coarsening which can be extended to a two-step optimization formulation for featured graph coarsening. 
 \item[(c)] \textbf{Featured graph coarsening with feature dimensionality reduction}\\
 This formulation jointly performs the graph coarsening and also reduces the dimension of the coarsened feature matrix.
\end{itemize}
 \item[(2)] The first and the third proposed formulations are multi-block non-convex differentiable optimization problems. The second formulation is a strictly convex differentiable optimization. To solve the proposed formulations, we developed algorithms based on the block majorization-minimization (MM) framework, commonly known as block successive upper-bound minimization (BSUM). The convergence analyses of the proposed algorithms are also presented.

\item[(3)] The efficacy of the proposed algorithms are thoroughly validated through exhaustive experiments on both synthetic and real data sets. The results show that the proposed methods outperform the state-of-the-art methods under various metrics like relative eigen error, hyperbolic error, spectral similarity, etc. We also prove that the learned coarsened graph is $\epsilon$-similar to the original graph, where $\epsilon\in(0,1)$.

 \item[(5)] The proposed featured graph coarsening framework is also shown to be applicable for traditional graph-based applications, like graph clustering and stochastic block model identification. 
\end{itemize}

\subsection{Outline and Notation}
The rest of the paper is organized as follows. In Section 2, we present the related background of graphs and graph coarsening. All the proposed problem formulations are shown in Section 3. In Sections 4, 5, and 6, we introduce the development of the algorithms with their associated convergence results. In Section 7, we discussed how the proposed coarsening is related to clustering and community detection. Section 8 presents experimental results on both real and synthetic datasets for all the proposed algorithms.
\\
In terms of notation, lower case (bold) letters denote scalars (vectors) and upper case letters denote matrices. The dimension of a matrix is omitted whenever it is clear from the context. The $(i,j)$-th entry of a matrix $X$ is denoted by $X_{ij}$. $X^\dagger$ and $X^\top$ denote the pseudo inverse and transpose of matrix $X$, respectively. $X_i$ and $[X^T]_j$ denote the $i$-th column and $j$-th row of matrix X. The all-zero and all-one vectors or matrices of appropriate sizes are denoted by $\bzero$ and $\bone$, respectively. The $\norm{X}_1$, $\norm{X}_F$, $\norm{X}_{1,2}$ denote the $\ell_1$-norm, Frobenius norm and $\ell_{1,2}$-norm of $X$, respectively. The Euclidean norm of the vector $X$ is denoted as $\norm{X}_{2}$. $\text{det}(X)$ is defined as the generalized determinant of a positive definite matrix $X$, i.e., the product of its non-zero eigenvalues. The inner product of two matrices is defined as $\langle X, Y\rangle=\text{tr}(X^\top Y)$, where $\text{tr}(\cdot)$ is the trace operator. $\mathbb{R}_+$ represents positive real numbers. The inner product of two vectors is defined as $\langle X_i, X_j\rangle=X_i^TX_j$ where $X_i$ and $X_j$ are the $i$-th and $j$-th column of matrix $X$.

\section{Background }
In this section, we review the basics of graph and graph coarsening, the spectral similarity of the graph matrices, the $\epsilon$-similarity of graph matrices and feature matrices, the hyperbolic error and the reconstruction error of lifted graph.
\subsection{Graph}
A graph with features is denoted by $\mathcal{G}=(V,E,W, X)$ where $V = \{v^1, v^2, . . . , v^p\}$ is the vertex set, $E \subseteq V \times V$ is the edge set and $W$ is the adjacency (weight) matrix. We consider a simple undirected graph without self-loop: $W_{ij} >0$, if $(i, j) \in E $ and $W_{ij} =0$, if $(i, j) \notin E $. Finally, $X \in \mathbb{R}^{p\times n}=[X_1,X_2,\dots, X_p]^\top$ is the feature matrix, where each row vector $X_i\in \mathbb{R}^{n} $ is the feature vector associated with one of $p$ nodes of the graph $\mathcal{G}$. Thus, each of the $n$ columns of $X$ can be seen as a signal on the same graph. Graphs are conveniently represented by some matrix, such as Laplacian and adjacency graph matrices, whose positive entries correspond to edges in the graph. 

A matrix $\Theta \in \mathbb{R}^{p \times p}$ is a combinatorial graph Laplacian matrix if it belongs to the following set:
\begin{align}\label{Lap-set}
\mathcal{S}_{\Theta} =\left\{ \Theta \in \mathbb{R}^{p \times p}| \Theta_{ij} =\Theta_{ji} \leq 0 \ {\rm for} \ i\neq j; \Theta_{ii}=-\sum_{j\neq i}\Theta_{ij} \right\}.
\end{align}
The $W$ and the $\Theta$ are related as follows: $W_{ij}=-\Theta_{ij}\; \textsf{for}\; i\neq j$ and $W_{ij}=0\; \textsf{for}\; i=j$. Both $\Theta$ and $W$ represent the same graph, however, they have very different mathematical properties. The Laplacian matrix $\Theta$ is a symmetric, positive semidefinite matrix with zero row sum. The non-zero entries of the matrix encode positive edge weights as $-\Theta_{ij}$ and $\Theta_{ij}=0$ implies no connectivity between vertices $i$ and $j$. The importance of the graph Laplacian matrix has been well recognized as a tool for embedding, manifold learning, spectral sparsification, clustering, and semi-supervised learning. Owing to these properties, Laplacian matrix representation is more desirable for building graph-based algorithms.

\subsection{Graph Coarsening}
Given an original graph $\mathcal{G} = (V,E,W,X)$ with $p$ nodes, the goal of graph coarsening is to construct an appropriate "smaller" or coarsen graph $\mathcal{G}_c = (\tilde{V}, \tilde{E}, \tilde{W}, \tilde{X})$ with $k << p$ nodes, such that $\G_c$ and $\G$ are similar in some sense. Every node $\tilde{v}^j\in \tilde{V}$, where $j={1,2,...k}$, of the smaller graph with reference to the nodes of the larger graph is termed as a "super-node". In coarsening, we define a linear mapping $\pi : V \rightarrow \tilde{V}$ that maps a set of nodes in $\mathcal{G}$ having similar properties to a super-node in $\mathcal{G}_c$ i.e. for any super-node $\tilde{v} \in \tilde{V}$, all nodes $\pi^{-1}(\tilde{v}) \subset V$ have similar properties. Furthermore, the features of the super-node, $\tilde{v}$, should be based on the features of nodes $\pi^{-1}(\tilde{v}) \subset V$ in $\mathcal{G}$, and the edge weights of the coarse graph, $\tilde{W}$, should depend on the original graph's weights as well as the coarsen graph's features.\\
Let $P\in \mathbb{R}_+^{k\times p}$ be the coarsening matrix which is a linear map from $\pi : V \rightarrow \tilde{V}$ such that $\tilde{X}=PX.$ Each non-zero entry of $P$ i.e. $[P]_{ij}$, indicate the $i$-th node of $\mathcal{G}$ is mapped to $j$-th super node of $\mathcal{G}_c$. For example, non-zero elements of $j$-th row, i.e., $\mathbf{p}_j$ corresponds to the following nodes set $\pi^{-1}(\tilde{v}_j)\in V $. The rows of $P$ will be pairwise orthogonal if any node in $V$ is mapped to only a single super-node in $\tilde{V}_c$. This means that the grouping via super-node is disjoint.
 Let the Laplacian matrices of $\mathcal{G}$ and $\mathcal{G}_c$ be $\Theta \in \mathbb{R}^{p \times p}$ and $\Theta_c \in \mathbb{R}^{k \times k}$, respectively. The Laplacian matrices $\Theta$, $\Theta_c$, feature matrices $X$, $\tilde{X}$ and the coarsening matrix $P$ together satisfy the following properties\cite{loukas2019graph}:

\begin{equation}\label{coarsening}
 \Theta_c=C^T\Theta C, \quad \tilde X=PX, \quad X=P^{\dagger}\tilde{X}=C\tilde{X}
\end{equation}
where $C \in \mathbb{R}^{p \times k}$ is the tall matrix which is the pseudo inverse of $P$ and known as the loading matrix. The non-zero elements of $C$, i.e., $C_{ij}>0$ implies that the $i$-th node of $\mathcal{G}$ is mapped to the $j$-th supernode of $\mathcal{G}_c$. The loading matrix $C$ belongs to the following set:
\begin{align}\label{Loading matrix-set1}
\mathcal{C} =\left\{ C \in \mathbb{R}_+^{p \times k}|,\ \langle C_i, C_j \rangle=0 \ \forall \; i\neq j,\ \langle C_l, C_l \rangle=d_i,\ \norm{C_i}_0\geq 1 \ and \ \norm{[C^T]_i}_0= 1 \right\}
\end{align}
where $C_i$ and $C_j$ represent $i$-th and $j$-th column of loading matrix $C$ and they are orthogonal to each other, $[C^T]_i$ represents the $i$-th row of loading matrix $C$. There are a total of $k$ columns and $p$ rows in the $C$ matrix. Also, in each row of the loading matrix $C$, there is only one non zero entry and that entry is 1 which implies that $C \cdot \textbf{1}_k=\textbf{1}_p$, where $\textbf{1}_k$ and $\textbf{1}_p$ are vectors having all entry $1$ and having size of $k$ and $p$ respectively. Furthermore, as each row of loading matrix $C$ has only one non-zero entry, this also implies that $C^TC=\text{block}(\mathbf{d})$, where $\text{block}(\mathbf{d})$ is the diagonal matrix of size $k$ containing $d_i>0\;\forall\; i=1,2,\dots,k$ at it's diagonal. Furthermore, $d_i$ also indicates the number of nodes of the graph $\mathcal{G}$ mapped to $i$-th super-node of the coarsened graph $\mathcal{G}_c$.

\begin{figure}
 \centering
 \begin{subfigure}[b]{0.45\textwidth}
 \centering
 \includegraphics[width=\textwidth]{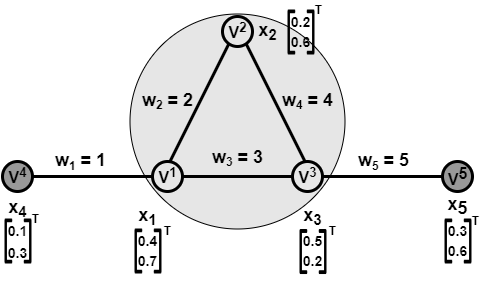}
 \caption{Original graph $\mathcal{G}$}
 \label{original graph}
 \end{subfigure}
 \hfill
 \begin{subfigure}[b]{0.45\textwidth}
 \centering
 \includegraphics[width=\textwidth]{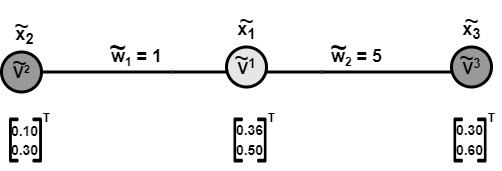}
 \caption{Coarsened graph $\mathcal{G}_c$}
 \label{coarsened graph}
 \end{subfigure}
 \hfill
 \caption{Toy example: graph coarsening.}
 \label{fig:my_label}
\end{figure}

In the toy example, nodes $(v^1, v^2, v^3)$ of $\mathcal{G}$ are coarsened into super-node $\tilde{v}^1$ of $\mathcal{G}_c$. The coarsening matrix $P$ and the loading matrix $C$ are

$$
P=
\begin{bmatrix}
 \frac{1}{3}& \frac{1}{3} & \frac{1}{3} & 0 & 0\\
 0 & 0 & 0 & 1 & 0\\
 0 & 0 & 0 & 0 & 1 
\end{bmatrix}
 \quad \text{and} \quad C=P^{\dagger}=
\begin{bmatrix}
 1 & 0 & 0 \\
 1 & 0 & 0\\ 
 1 & 0 & 0 \\
 0 & 1 & 0\\
 0 & 0 & 1
\end{bmatrix}
$$
For toy example, feature matrix $X$ of $\mathcal{G}$ is $X=\begin{bmatrix}
 0.4 & 0.2 & 0.5 & 0.1 & 0.3 \\
 0.7 & 0.6 & 0.2 & 0.3 & 0.6 \\
 \end{bmatrix}^T$. The feature matrix for $\mathcal{G}_c$ is calculated using $\tilde{X}=PX$ from \eqref{coarsening} and we get $\tilde{X}=\begin{bmatrix}
 0.36 & 0.10 & 0.30 \\
 0.50 & 0.30 & 0.60 \\
 \end{bmatrix}^T$. Also weight vector of $\mathcal{G}$ is $\textbf{w}=[w_1 \quad w_2\quad w_3\quad w_4 \quad w_5]^T=[1 \quad 2 \quad 3 \quad 4 \quad 5]^T$ . The Laplacian matrices $\Theta$ and $\Theta_c$ for $\mathcal{G}$ and $\mathcal{G}_c$ using (\ref{coarsening}) are
$$
\Theta=
 \begin{bmatrix}
 6 & -2 & -3 & -1 & 0 \\
 -2 & 6 & -4 & 0 & 0 \\
 -3 & -4 & 12 & 0 & -5 \\
 -1 & 0 & 0 & 1 & 0 \\
 0 & 0 & -5 & 0 & 5
 \end{bmatrix}
\quad \text{and} \quad \Theta_c=C^T\Theta C=
 \begin{bmatrix}
 6 & -1 & -5 \\
 -1 & 1 & 0 \\
 -5 & 0 & 5 \\
 \end{bmatrix}.
$$
The weights matrix of $\mathcal{G}_c$ is $\tilde{\textbf{w}}=[\tilde{w}_1 \quad \tilde{w}_2]^T=[1 \quad 5]^T$.

\subsection{Lifted Laplacian( \texorpdfstring{$\Theta_{\text{lift}}$}{})}
From the coarsened dimension of $k\times k$ one can go back to the original dimension, i.e., $p\times p$ by computing the lifted Laplacian matrix \cite{loukas2018spectrally} defined as
\begin{equation}
\Theta_{\text{lift}}=P^T\Theta_c P 
\end{equation}
where $P \in \mathbb{R}_+^{k \times p}$ is coarsening matrix and $\Theta_{C} \in \mathcal{S}_{\Theta}$ is the Laplacian of coarsened graph.
\subsection{Preserving properties of \texorpdfstring{$\mathcal{G}$ in $\mathcal{G}_c$}{}}

The coarsened graph $\mathcal{G}_c(\Theta_c, \tilde{X})$ should be learned such that the properties of $\mathcal{G}$ and $\mathcal{G}_c$ are similar. The widely used notions of similarities are (i) spectral similarity (ii) $\epsilon$-similarity \cite{loukas2018spectrally,loukas2019graph} (iii) hyperbolic error \cite{bravo2019unifying} (iv) Reconstruction error .

\begin{def1} \textbf{Spectral similarity} The spectral similarity is shown by calculating the relative eigen error (\textbf{REE}), defined as

\begin{align}\label{eig_error}
 \text{REE}(\Theta,\Theta_c,m)= \frac{1}{m}\sum_{i=1}^{m}\frac{|\tilde{\lambda}_i-\lambda_i|}{\lambda_i} 
\end{align}
 where ${\lambda_i}$ and ${\tilde{\lambda}_i}$ are the top $m$ eigenvalues corresponding to the original graph Laplacian matrix $\Theta$ and coarsened graph Laplacian matrix $\Theta _c$ respectively and m is the count of eigenvalue. 
 \end{def1}
 The REE value indicates that how well the eigen properties of the original graph $\mathcal{G}$ is preserved in the coarsened graph $\mathcal{G}_c$. A low REE will indicate higher spectral similarity, which implies that the eigenspace of the original graph matrix and the coarsen graph matrix are similar.

\begin{def1} \textbf{Hyperbolic error (HE)} For the given feature matrix $X$, the hyperbolic error between original Laplacian matrix $\Theta$ and lifted Laplacian matrix $\Theta_{\text{lift}}$ is defined as
\begin{equation}\label{HE}
HE=\text{arccosh}\Bigg(1+\frac{\|(\Theta-\Theta_{\text{lift}})X\|_F^2\|X\|_F^2}{2\text{tr}(X^T\Theta X)\text{tr}(X^T\Theta_{\text{lift}}X)}\Bigg).
\end{equation}

\end{def1}
\begin{def1} \textbf{Reconstructional Error (RE)}
 Let $\Theta$ be original Laplacian matrix and $\Theta_{\text{lift}}$ be the lifted Laplacian matrix,then the reconstruction error(\textbf{RE}) \cite{valle1999selection} is defined as
 \begin{equation}\label{RE}
 RE=\|\Theta-\Theta_{\text{lift}}\|_F^2 
 \end{equation}
 \end{def1}
 
 For a good coarsening algorithm lower values of these quantities are desired. Note that the above metrics only take into account the properties of graph matrices but not the associated features. To quantify how well a graph coarsening approach has performed for graphs with features, we propose to use the $\epsilon$-similarity measure, which considers both the graph matrix and associated features. It is also highlighted that the $\epsilon-$similarity in \cite{loukas2018spectrally,loukas2019graph} does not consist of features. In \cite{loukas2018spectrally,loukas2019graph} the eigenvector of the Laplacian matrix is considered while computing the $\epsilon$ similarity, which can only capture the properties of the graph matrix, not the associated features.
\begin{def1}\textbf{$\epsilon$-similarity} The coarsened graph data $\G_c(\Theta_c ,\tilde{X})$ is $\epsilon$ similar to the original graph data $\G(\Theta, X)$ if there exist an $\epsilon \geq 0$ such that
\begin{equation}\label{epsilon similar}
 (1-\epsilon)\|X\|_{\Theta} \leq \|\tilde{X}\|_{\Theta_c} \leq (1+\epsilon)\|X\|_{\Theta}
\end{equation}
\end{def1} 
where $\|X\|_{\Theta}=\sqrt{\text{tr}(X^T\Theta X)}$ and $\|\tilde{X}\|_{\Theta}=\sqrt{\text{tr}(\tilde{X}^T\Theta_c \tilde{X})}$.

\section{Problem Formulation}
The existing graph coarsening methods are not designed to consider the node features and solely rely on the graph matrix for learning a simpler graph \cite{loukas2018spectrally,loukas2019graph,bravo2019unifying,purohit2014fast,chen2022graph}, and thus, not suitable for graph machine learning applications. For example, many real-world graph data satisfy certain properties, e.g., homophily assumption and smoothness \cite{unknown, pmlr-v51-kalofolias16}, that if two nodes are connected with stronger weights, then the features corresponding to these nodes should be similar. Thus, if the original graph satisfies any property, then that property should translate to the coarsen graph data. Current methods can only ensure spectral properties which satisfy the property of the graph matrix but not the features \cite{loukas2018spectrally,loukas2019graph,chen2022graph}. This is slightly restrictive for graph-based downstream tasks, where both the nodal features and edge connectivity information are essential.

The aforementioned discussion suggests that the following graph coarsening method (i) should consider jointly both the graph matrix $\Theta$ and the node feature $X$ of the original graph and (ii) to ensure the desired specific properties on coarsened graph data, such as smoothness and homophily, the $\Theta_c$ and $\tilde{X}$ should be learned jointly depending on each other. This problem is challenging and it is not straightforward to extend the existing methods and make them suitable for considering both the node features and graph matrix jointly to learn coarsened graphs. We envision approaching this problem at the unification of dimensionality reduction \cite{qiu2017deep, zhu2017subspace} and graph learning \cite{kumar2020unified,NEURIPS2019_90cc440b}, where we solve these problems jointly, first we reduce the dimensionality and then learn a suitable graph on the reduced dimensional data. We propose a unique optimization-based framework that uses both the features $X$ and Laplacian matrix $\Theta$ of the original graph to learn loading matrix $C$ and coarsened graph's features $\tilde{X}$, jointly. Thus, firstly in this section, we briefly discuss how to learn graphs with features, and then we propose our formulation.

\subsection{Graph learning from data}
Let $X=[X_1,X_2,\ldots,X_p]^T$, where $X_i$ is $n-$dimensional feature vector associated with $i-$th node of an undirected graph. In the context of modeling signals or features with graphs, the widely used assumption is that the signal residing on the graph changes smoothly between connected nodes \cite{pmlr-v51-kalofolias16}. The Dirichlet energy (DE) is used for quantifying the smoothness of the graph signals which is defined by using graph Laplacian matrix $\Theta \in \mathcal{S}_\Theta$ matrix and the feature vector as follows: 
\begin{align}\label{DE}
\text{DE}(\Theta,X)= \text{tr}(X^T\Theta X)=-\sum_{i,j}\Theta_{ij}\norm{\mathbf{x}_i-\mathbf{x}_j}^2.
\end{align}
The lower value of Dirichlet energy indicates a more desirable configuration \cite{unknown}. Smooth graph signal methods are an extremely popular family of approaches for a variety of applications in machine learning and related domains \cite{dong2016learning}. When only the feature matrix $X=[X_1,X_2,\ldots,X_p]^T$, associated with an undirected graph is given then a suitable graph satisfying the smoothness property can be obtained by solving the following optimization problem:
\begin{align}\label{GMRF}
\begin{array}{ll}
\underset{\Theta \in \mathcal{S}_{\Theta} }{\textsf{minimize}} &
\begin{array}{c}
- \gamma \log(\det(\Theta + J))+\text{tr}(X^T\Theta X)+ \alpha h(\Theta) 
\end{array}
\end{array}
\end{align}
where $\Theta\in \mathbb{R}^{p\times p}$ denotes the desired graph matrix, $\mathcal{S}_{\Theta}$ is the set of Laplacian matrix $\eqref{Lap-set}$, $h(\cdot)$ is the regularization term, and $\alpha>0$ is the regularization parameter, and $J=\frac{1}{p}\textbf{1}_{p\times p}$ is a constant matrix whose each element is equal to $\frac{1}{p}$. The rank of $\Theta$ is $p-1$ for connected graph matrix having p nodes\cite{Chung:1997}, adding $J$ to $\Theta$ makes $\Theta+ J$ a full rank matrix without altering the row and column space of the matrix $\Theta$\cite{kumar2020unified, pmlr-v51-kalofolias16}.

When the data is Gaussian distributed $X \sim \mathcal{N}(\mathbf{0}, \Theta^\dagger)$, optimization in \eqref{GMRF} also corresponds to the penalized maximum likelihood estimation of the inverse covariance (precision) matrix also known as Gaussian Markov random field (GMRF) for $\gamma =1$ \cite{ying2020nonconvex}. The graph $\mathcal{G}$ inferred from $\Theta$ and the random vector $X$ follows the Markov property, meaning $\Theta_{ij}\neq 0 \iff \{i,j\} \in E\;\forall i\neq j$ implies $X_i$ and $X_j$ are conditionally dependent given the rest. Furthermore, in a more general setting with non-Gaussian distribution, \eqref{GMRF} can be related to the log-determinant Bregman divergence regularized optimization problem, which ensures nice properties on the learned graph matrix, e.g., connectedness and full rankness.
\newline
In the next subsections, we introduce three optimization frameworks for graph coarsening i) Graph coarsening for nodes with Features, ii) Graph coarsening for nodes without features, and iii) Graph coarsening for nodes with features and feature dimensionality reduction.

\subsection{A General Framework for Graph Coarsening with Features}
We introduce a general optimization-based framework for graph coarsening with features as follows
\begin{align}\label{original}
\begin{array}{ll}
\underset{\Theta_c,\tilde{X}, {C}}{\textsf{minimize}} &
\begin{array}{c}
- \gamma \log(\det(\Theta_c + J))+\text{tr}(\tilde{X}^{T}\Theta_c\tilde{X}) + \beta h(\Theta_c)+ \frac{\lambda}{2} g(C)
\end{array}\\
\textsf{subject to} & \begin{array}[t]{l}
 \ C \geq 0, \ \Theta_c=C^T\Theta C, \ X= C\tilde{X}, \ \Theta_c \in \S_{\Theta}, C \in \mathcal{C}
\end{array}
\end{array}
\end{align}
where $\Theta$ and $X$ are the given Laplacian and feature matrix of a large connected graph, and ${\tilde{X}} \in \mathbb{R}^{k \times n}$ and $\Theta_c \in \mathbb{R}^{k \times k}$ are the feature matrix and the Laplacian matrix of the learned coarsened graph, respectively, $C \in \mathbb{R}^{p \times k}$ is the loading matrix, $h(\cdot)$ and $g(\cdot)$ are the regularization functions for $\Theta_c$ and the loading matrix $C$, while $\beta>0$ and $\lambda>0$ are the regularization parameters. Fundamentally, the proposed formulation \eqref{original} aims to learn the coarsened graph matrix $\Theta_c$, the loading matrix $C$, and the feature matrix $\tilde{X}$, jointly. This constraint $X=C\tilde{X}$ coarsens the feature matrix of larger graph $X \in \mathbb{R}^{p\times n}$ to a smaller graph's feature matrix $\tilde{X} \in \mathbb{R}^{k\times n}$. Next, the first two-term of the objective function is the graph learning term, where the $\log \det(\cdot)$ term ensures the coarsen graph is connected while the second term imposes the smoothness property on the coarsened graph, and finally third and fourth term act as a regularizer. The regularizer $g(C)$ ensures the mapping such that one node $v^i \in V$ does not get mapped to two different super-nodes $\tilde{v}^j, \tilde{v}^k \in \tilde{V}$ and mapping of nodes to super-nodes be balanced such that not all or majority of nodes get mapped to the same super-node. This simply implies that only one element of each row of $C$ be non-zero and the columns of $C$ be sparse. An $\ell_{1,2}$-based group penalty is suggested to enforce such structure \cite{yuan2006model, ming2019probabilistic}.
\subsection{Graph coarsening without Features}

In a variety of network science dataset, we are only provided with the adjacency matrix without any node features \cite{preis1997party, gleich2008matlabbgl, turk1994zippered}. Ignoring the feature term $X$ in \eqref{original}, the proposed formulation for graph coarsening without node features is
\begin{align}\label{Main formulation without X1}
\begin{array}{ll}
\underset{{\Theta_c , C}}{\textsf{minimize}} &
\begin{array}{c}
\hspace{-1em}-\gamma \text{log det}(\Theta_c +J) +\beta h(\Theta_c)+ \frac{\lambda}{2} g(C) 
\end{array}\\
\textsf{subject to} & \begin{array}[t]{l}
 C \geq 0,\ \Theta_c=C^T\Theta C, \ \Theta_c \in \S_{\Theta},\ C \in \mathcal{C}
\end{array}
\end{array}
\end{align}
where $\Theta$ is the given Laplacian of a large connected graph, $\Theta_c$ is the Laplacian matrix of the learned coarsened graph, $C \in \mathbb{R}^{p \times k}$ is the loading matrix, $h(\cdot)$ and $g(\cdot)$ are the regularization functions for $\Theta_c$ and the loading matrix $C$, while $\beta>0$ and $\lambda>0$ are the regularization parameters. Fundamentally, the proposed formulation \eqref{Main formulation without X1} aims to learn the coarsened graph matrix $\Theta_c$ and the loading matrix $C$. The first term i.e. $\text{log det}(\cdot)$ ensures the coarsen graph is connected, second term i.e. $h(\cdot)$ is the regularizer on coarsening Laplacian matrix $\Theta_c$ which imposes sparsity in the resultant coarsen graph and finally the third term i.e. $g(C)$ is the regularizer on loading matrix $C$ which ensures the mapping of node-supernode should be balanced such that a node of the original graph does not get mapped to two supernodes of coarsened graph, also not all majority of nodes of original graph get mapped to the same supernode. In this formulation also, an $\ell_{1,2}$-based group penalty is suggested to enforce such structure \cite{yuan2006model, ming2019probabilistic}.

\subsection{Graph Coarsening with Feature Dimensionality Reduction}
In the FGC algorithm, the dimension of the feature of each node of the original graph $\mathcal{G}$ and $\mathcal{G}_c$ are the same i.e both are in $\mathbb{R}^n$ dimension. As we reduce the number of nodes, it may be desirable to reduce the dimension of the features as well associated with each supernode. However, the proposed FGC algorithm can be adapted to reduce the dimension of features of each node, by combining various feature dimensionality techniques. Here, we propose to integrate the matrix factorization technique on the feature matrix within the FGC framework, we name it FGC with dimensionality reduction (FGCR). Using matrix factorization \cite{fu2019nonnegative}, the feature dimension of each node of coarsened graph $\mathcal{G}_c$ reduces from $\mathbb{R}^n$ to $\mathbb{R}^d$ using 
\begin{equation}
 \tilde{X}=WH
\end{equation}
where $W \in \mathbb{R}^{k \times d}$ be the feature matrix in reduced dimension, $H \in \mathbb{R}^{d \times n}$ be the transformation matrix and always $d<<n$.\\
In FGC, we learn the coarsened graph with coarsened graph feature matrix $\tilde{X} \in \mathbb{R}^{k \times n}$, where each node has features in $\mathbb{R}^n$ dimension. Now using matrix factorization $\tilde{X}=WH$, we reduce the feature of each node from $\mathbb{R}^{n}$ to $\mathbb{R}^d$ and learn the coarsened graph with reduced feature matrix $W \in \mathbb{R}^{k \times d}$.\\
The proposed formulation for learning a coarsened graph while reducing the dimension of the feature of each node simultaneously is
\begin{align}\label{Main formulation with reduced features12}
\begin{array}{ll}
\underset{{W},{H}, {C}, \Theta_c, \tilde{X}}{\textsf{minimize}} &
\begin{array}{c}
\hspace{-1em}-\gamma \text{log det}(\Theta_c +J)+\text{tr}(W^TC^T\Theta CW) + \beta h(\Theta_c)+ \frac{\lambda}{2} g(C)
\end{array}\\
\textsf{subject to} & \begin{array}[t]{l}
 C \geq 0,\ \Theta_c=C^T\Theta C,\ X=C\tilde{X},\ \tilde{X}=WH,\ \Theta_c \in \S_{\Theta},\ C \in \mathcal{C}
\end{array}
\end{array}
\end{align}
where $\Theta$ and $X$ are the given Laplacian and feature matrix of a large connected graph, $W \in \mathbb{R}^{k \times d}$ and $\Theta_c$ are the reduced dimension feature matrix and Laplacian matrix of coarsened graph, respectively, $C \in \mathbb{R}^{p \times k}$ be the loading matrix, $H \in \mathbb{R}^{d \times n}$ be the transformation matrix, $h(\cdot)$ and $g(\cdot)$ are the regularization function for $\Theta_c$ and the loading matrix $C$, while $\beta > 0$ and $\lambda >0$ are the regularization parameters. Fundamentally the problem formulation \eqref{Main formulation with reduced features12} aims to learn the coarsened graph matrix $\Theta_c$ , reduced feature matrix $W$, and transformation matrix $H$, jointly. This constraint $X=C\tilde{X}$ coarsen the feature matrix of the large graph but will not reduce the dimension of the feature of each supernode and the constraint $\tilde{X}=WH$
reduces the dimension of each supernode of the coarsened graph from $\mathbb{R}^n$ to $\mathbb{R}^d$. Next, the first two-term of the objective function is the graph learning term, where the $\log\det(\cdot)$ term ensures the coarsen graph is connected while the second term imposes the smoothness property on the coarsened graph with reduced feature matrix $W$, and finally the third and fourth act as a regularizer which is same as in the FGC algorithm.

\subsection{Some properties of \texorpdfstring{$C^T\Theta C$}   {} matrix}

Before we move forward toward algorithm development, some of the properties and intermediary Lemmas are presented below.

\begin{Lem1}\label{Lemma 1}
If $\Theta$ be the Laplacian matrix for a connected graph with $p$ nodes, and $C$ be the loading matrix such that $C\in\mathbb{R}_+^{p\times k}$ and $C\in\mathcal{C}$ as in \eqref{Loading matrix-set1}
, then the coarsened matrix $\Theta_c=C^T \Theta C$ is a connected graph Laplacian matrix with $k$ nodes.

\end{Lem1}

\begin{proof}

 The matrix $\Theta \in \mathbb{R}^{p \times p}$ is the Laplacian matrix of a connected graph having $p$ nodes.
From \eqref{Lap-set} it is implied that $\Theta=\Theta^T$, $\Theta$ is positive semi-definite matrix with rank$(\Theta)=p-1$ and $\Theta\cdot t\mathbf{1}_p=\mathbf{0}_p$, where $t\in \mathbf{R}$ and $\mathbf{1}_p$ and $\mathbf{0}_p$ are the all one and zero vectors of size $p$. In addition, we also have $\Theta\cdot \mathbf{u}_p\neq\mathbf{0}_p$ for $\mathbf{u}_p\neq t\mathbf{1}_p$, this means that there is only one zero eigenvalues possible and the corresponding eigenvector is a constant vector. In order to establish $\Theta_c$ is the connected graph Laplacian matrix of size $p$, we need to prove that $\Theta_c\in \eqref{Lap-set} $ and rank$(\Theta_c)=k-1$.


We begin by using the Cholesky decomposition of the Laplacian matrix $\Theta$, as $\Theta=S^TS$. Next, we can write $C^T\Theta C$ as 
\begin{align}
\Theta_c=C^T\Theta C
&=C^TS^TSC\\
&= Z^TZ
\end{align}
where $Z=SC$ and $C^T\Theta C=Z^TZ$ imply that $\Theta_c$ is a symmetric positive semidefinite matrix. Now, using the property of $C$, i.e, $C\cdot t\mathbf{1}_k=t\mathbf{1}_p$ as in \eqref{Loading matrix-set1}. In each row of the loading matrix $C$, there is only one non zero entry and that entry is 1 which implies that $C\cdot\textbf{1}_k=\textbf{1}_p$ and $C^T\Theta C\cdot \textbf{1}_k=C^T\Theta\cdot \textbf{1}_p=\textbf{0}_k$ which imply that the row sum of $C^T\Theta C$ is zero and constant vector is the eigenvector corresponding to the zero eigenvalue. Thus $\Theta_c$ is the Laplacian matrix.

Next, we need to prove that $\Theta_c$ is a connected graph Laplacian matrix for that we need to prove that rank$(\Theta_c)=k-1$. Note that, $C\cdot \textbf{u}_k = \textbf{u}_p$ if and only if $\textbf{u}_k=t\textbf{1}_k$ and $\textbf{u}_p=t\textbf{1}_p$ and $C\cdot\textbf{u}_k \neq \textbf{u}_p\; \forall\; \textbf{u}_k \neq t\textbf{1}_k$ and $\textbf{u}_p \neq t\textbf{1}_p$ where $t \in \mathbb{R}$ which implies that $C\cdot \textbf{u}_k = \textbf{u}_p$ holds only for a constant vector $\textbf{u}_k$. And thus, $C^T\Theta C\cdot \mathbf{u}_k=\mathbf{0}_k $, for constant vector $\mathbf{u}_k$. This concludes that the constant vector is the only eigenvector spanning the nullspace of $\Theta_c$ which concludes that the rank of $C^T\Theta C$ is $k-1$ which completes the proof.
\end{proof}

\begin{Lem1}\label{convex}
If $\Theta$ be the Laplacian matrix for a connected graph with $p$ nodes, $C$ be the loading matrix, and $J=\frac{1}{k}\textbf{1}_{k\times k}$ is a constant matrix whose each element is equal to $\frac{1}{p}$.\\ The function $f(C)=-\gamma \text{log det}(C^T\Theta C +J)$ is a convex function with respect to the loading matrix $C$.
\end{Lem1}

\begin{proof}
We prove the convexity of $-\gamma \text{log det}(C^T\Theta C +J)$ using restricting function to line i.e. A function $f:\mathbb{R}^n\rightarrow \mathbb{R}$ is convex if $g:\mathbb{R}\rightarrow \mathbb{R}$ is convex where, 
\begin{equation}
 g(t)=f(z+tv), \{z \in dom(f), t \in dom(g), v \in \mathbb{R}^n\}
\end{equation}
Since $\Theta$ is the Laplacian of connected original Graph $\mathcal{G}$ and Laplacian of coarsened graph $\mathcal{G}_c$ is $\Theta_c= C^T\Theta C$ which also represents a connected graph and proof is given in Lemma \ref{Lemma 1}. Using the property of the connected graph Laplacian matrix, $\Theta_c$ is a symmetric positive semi-definite matrix and has a rank $k-1$. Adding $J=\frac{1}{k}\textbf{1}_{k \times k}$ which is a rank $1$ matrix in $\Theta_c$ increases rank by 1. $\Theta_c+J$ becomes symmetric and positive definite matrix and we can rewrite $\Theta_c+J=C^T\Theta C+J=Y^TY$. Now, we can rewrite $-\gamma \text{log det}(C^T\Theta C +J)$ as
\begin{equation}\label{f(X)}
 f(Y)= -\gamma\text{log det}(C^T\Theta C +J)=-\gamma \text{log det}(Y^TY)
\end{equation}
Consider Y=Z+tV and put it in \eqref{f(X)}. However, Z and V are constant so function in Y becomes function in t i.e. $g(t)$ is
\begin{align}
g(t)
&=-\gamma \text{log det}((Z+tV)^T(Z+tV)) \\
&=-\gamma \text{log det}(Z^TZ+t(Z^TV+V^TZ)+t^2V^TV) \\\label{proofZ}
&= -\gamma \text{log det}(Z^T(I+t(VZ^{-1}+(VZ^{-1})^T)+t^2(Z^{-1})^TV^TVZ^{-1})Z)\\\label{proof1}
&=-\gamma (\text{log det}(Z^TZ)+\text{log det}(I+t(P+P^T)+t^2P^TP))\\\label{proof}
&=-\gamma (\text{log det}(Z^TZ)+\text{log det}(QQ^T+2tQ \Lambda Q^T+t^2Q\Lambda^2Q^T))\\\label{proof2}
&=-\gamma (\text{log det}(Z^TZ)+\text{log det}(Q(I+2t\Lambda+t^2\Lambda^2)Q^T))\\
&=-\gamma \text{log det}(Z^TZ)-\gamma\sum_{i=1}^n\text{log}(1+2t\lambda_i+t^2\lambda_i^2)
\end{align}
On putting $P=VZ^{-1}$ in \eqref{proofZ} to get \eqref{proof1}. Using eigenvalue decomposition of P matrix i.e. $P=Q \Lambda Q^T$ and $QQ^T=I$ and putting the values of $P$ and $I$ in \eqref{proof1} to get \eqref{proof}. The second derivative of g(t) with respect to $t$ is
\begin{equation}
 g^{''}(t)=\sum_{i=1}^n\frac{2\lambda_i^2(1+t\lambda_i)^2}{(1+2t\lambda_i+t^2\lambda_i^2)^2}
\end{equation}
It is clearly seen that $g^{''}(t)\geq 0$, $ \forall t \in \mathbb{R}$ so it is a convex function in $t$. Now, using the restricting function to line property if $g(t)$ is convex in $t$ then $f(Y)$ is convex in $Y$. Consider $Y=\Theta^{\frac{1}{2}}C+\frac{1}{\sqrt{kp}}\textbf{1}_{P \times k}$ so,
\begin{align}
Y^TY
&=(\Theta^{\frac{1}{2}}C+\frac{1}{\sqrt{kp}}\textbf{1}_{P \times k})^T(\Theta^{\frac{1}{2}}C+\frac{1}{\sqrt{kp}}\textbf{1}_{P \times k})\\
&=C^T\Theta C+\frac{1}{kp}(p\textbf{1}_{k \times k})+\frac{1}{\sqrt{kp}}\textbf{1}_{P \times k}^T\Theta^{\frac{1}{2}}C+\frac{1}{\sqrt{kp}}C^T(\Theta^{\frac{1}{2}})^T\textbf{1}_{P \times k} \label{Laplacianproperty1}\\
&=C^T\Theta C+\frac{1}{k}\textbf{1}_{k \times k}\label{final}\\
&=C^T\Theta C+J
\end{align}
$\Theta$ is a Laplacian matrix so $\Theta^{\frac{1}{2}}$ is also Laplacian matrix and using the property of Laplacian matrix i.e. $\Theta^{\frac{1}{2}}.\textbf{1}_{p \times k}=\textbf{0}_{p \times k}$ and $\textbf{1}_{p \times k}^T.\Theta^{\frac{1}{2}}=\textbf{0}_{k \times p}$ in \eqref{Laplacianproperty1}, we get \eqref{final}.
Since $Y=\Theta^{\frac{1}{2}}C+\frac{1}{\sqrt{pk}}\textbf{1}_{p \times k}$ and $f(Y)$ is convex in $Y$ and $C$ is a linear function of $Y$ so $-\gamma \text{log det}(C^T\Theta C +J)$ is a convex function in $C$.
\end{proof}

\subsection{ The \texorpdfstring{$\ell_{1,2}^2$}{} norm regularizer for balanced mapping}
The choice of regularizer on the loading matrix $C$ is important to ensure that the mapping of the node to the super node should be balanced, such that any node should not get mapped to more than one super node and there should be at least one node mapped to a supernode. This implies that the row of the $C$ matrix should have strictly one non-zero entry and the columns should not be all zeros. To ensure the desired properties in the $C$ matrix, i.e., $C_{ij}\geq 0$ and $C^TC=\text{block}(\mathbf{d})$, where $\text{block}(\mathbf{d})$ is the diagonal matrix of size $k$ containing $d_i>0\;\forall\; i=1,2,\dots,k$, at it's diagonal, we will use the $\ell_{1,2}^2$ norm regularization for $C$, i.e., $\|C^T\|_{1,2}^2$ \cite{article1}. Below lemma add more details to the property induced by this regularization.


\begin{Lem1}
For $C\geq 0$, $||C^T||_{1,2}^2$ regularizer is a differentiable function.
\end{Lem1}
\begin{proof}
 It is easy to establish by the fact that $C\geq 0$ and hence each entry of loading matrix is $C_{ij} \geq 0$. Using this,
We have $\|C^T\|_{1,2}^2=\sum_{i=1}^p\big(\sum_{j=1}^kC_{ij}\big)^2=\|C\|_F^2+\sum_{i \neq j}\langle C_i , C_j \rangle$ ${\rm for}$ $i,j=1,2,...k$ which is differentiable and it's differentiation with respect to loading matrix $C$ is $C.\textbf{1}_{k \times k}$ where $\textbf{1}_{k \times k}$ is matrix of size $k \times k$ having all entries $1$.

\end{proof}

\subsection{Block Majorization-Minimization Framework}\label{bsum-section}
The resulting optimization problems formulated in \eqref{original}, \eqref{Main formulation without X1}, and \eqref{Main formulation with reduced features12}
are non-convex problems.
Therefore we develop efficient optimization methods based on block MM
framework~\cite{article, 7547360}. First, we present a general scheme of the block MM framework
\begin{align}\label{bsum}
\begin{array}{ll}
\underset{\x}	{\textsf{minimize}} &
\begin{array}{c}
f(\x)
\end{array}\\
\textsf{subject to} & \begin{array}[t]{l}
\x \in \mathcal{X},
\end{array}
\end{array}
\end{align}
where the optimization variable $\x$ is partitioned into $m$ blocks as $\x=(\x_1,\x_2,\dots,\x_m)$,
with $\x_i \in \mathcal{X}_i$, $\mathcal{X}=\prod_{i=1}^{m}\mathcal{X}_i$ is a closed convex set, and $f: \mathcal{X}\rightarrow \Rn$ is a continuous function.
At the $t$-th iteration, each block $\x_i$ is updated in a cyclic order by solving the following:
\begin{align}\label{bsum-maj}
\begin{array}{ll}
\underset{\x_i}	{\textsf{minimize}} &
\begin{array}{c}
 g_i\left(\x_i\vert \x_1^{(t)}, \dots,\x_{i-1}^{(t)},\x_{i+1}^{(t-1)},\dots, \x_m^{(t-1)}\right),
\end{array}\\
\textsf{subject to} & \begin{array}[t]{l}
\x_i \in \mathcal{X}_i,
\end{array}
\end{array}
\end{align}
where $g_i\left(\x_i\vert\y_i^{(t)}\right)$ with $\y_i^{(t)}\triangleq \left(\x_1^{(t)}, \dots, \x_{i-1}^{(t)}, \x_{i}^{(t-1)}, \x_{i+1}^{(t-1)},\dots, \x_m^{(t-1)}\right)$
is a majorization function of $f(\x)$ at $\y_i^{(t)} $ satisfying
\begin{subequations}\label{bsum-maj-0}
	\begin{align}
&g_i\left(\x_i\vert \y_i^{(t)}\right)\; \text{is continuous in } \; \left(\x_i, \y_i^{(t)}\right), \; \forall\; i, \label{bsum-maj-4}\\
&g_i\left(\x^{(t)}_i\vert \y_i^{(t)}\right)=	f\left( \x_1^{(t)}, \dots,\x_{i-1}^{(t)},\x_{i}^{(t)},\x_{i+1}^{(t-1)},\dots, \x_m^{(t-1)}\right),\label{bsum-maj-1}\\
&g_i\left(\x_i\vert \y_i^{(t)}\right)\geq	f\left( \x_1^{(t)}, \dots,\x_{i-1}^{(t)},\x_{i},\x_{i+1}^{(t-1)},\dots, \x_m^{(t-1)}\right),\;\forall \; \x_i\in \cX_i,\forall \; \y_i\in \cX, \forall \; i,\label{bsum-maj-2} \\
&{ g'_i\left(\x_i;\b d_i\vert \y_i^{(t)}\right)\vert_{\x_i=\x_i^{(t)}} = f'\left( \x_1^{(t)}, \dots,\x_{i-1}^{(t)},\x_{i},\x_{i+1}^{(t-1)},\dots, \x_m^{(t-1)};\b d\right)},\nonumber\\
& \hspace{3cm}{ \forall\;\; \b d=(\b 0, \dots, \b d_i, \dots, \b 0)\;\text{such that}\; \x_i^{(t)}+\b d_i \in \cX_i,\; \forall\; i,} \label{bsum-maj-3}
	\end{align}
\end{subequations}
where $ f'(\x;\b d)$ {stands for the directional derivative at $\x$ along $ \b d$}~\cite{article}.
In summary, the framework is based on a sequential inexact block coordinate approach, which updates the variable
in one block keeping the other blocks fixed. If the surrogate functions $g_i$ is properly chosen, then the solution
to~\eqref{bsum-maj} could be easier to obtain than solving \eqref{bsum} directly.

\subsection{Majorization Function for \texorpdfstring{$\textbf{L}$}{}-smooth and Differentiable Function}
Consider a function $f(x)$ is $\textbf{L}$-smooth ($\textbf{L} > 0$) \cite{paulavivcius2006analysis} on $\mathbb{R}^n$, meaning that
\begin{equation}
 \|\nabla f(\textbf{x})-\nabla f(\textbf{y})\|\leq \textbf{L}\|\textbf{x}-\textbf{y}\|
\end{equation}
There are various set of functions which can satisfies properties \eqref{bsum-maj-4}-\eqref{bsum-maj-2}. The first order Taylor series expansion of $f(x)$ is \cite{paulavivcius2006analysis}:
\begin{equation}
f(\textbf{y}) \leq f(\textbf{x})+ \langle \nabla f(\textbf{x}), \textbf{y}-\textbf{x}\rangle + \frac{\textbf{L}}{2}\|\textbf{x}-\textbf{y}\|^2
\end{equation}
Thus the function 
\begin{equation}\label{majorization}
 h(\textbf{y},\textbf{x})=f(\textbf{x})+ \langle \nabla f(\textbf{x}), \textbf{y}-\textbf{x}\rangle + \frac{\textbf{L}}{2}\|\textbf{x}-\textbf{y}\|^2
\end{equation}
is the majorized function of $f(\textbf{x})$ at $\textbf{x}$.

\section{Proposed Featured Graph Coarsening (FGC) Algorithm}
In this section, we developed a block MM-based algorithm for featured graph coarsening (FGC). By using $\Theta_c=C^T\Theta C$, the three variable optimization problem \eqref{original} is equivalent to two variable optimization problem as:
\begin{align}\label{Main formulation FGC1}
\begin{array}{ll}
\underset{{\tilde{X}}, {C}}{\textsf{minimize}} &
\begin{array}{c}
\hspace{-1em}-\gamma \text{log det}(C^T\Theta C +J)+\text{tr}(\tilde{X}^{T}C^T\Theta C\tilde{X}) +\frac \lambda 2 \|C^T\|_{1,2}^2
\end{array}\\
\textsf{subject to} & \begin{array}[t]{l}
 C \geq 0,\ X=C \tilde{X}, \ \|[C^T]_i\|_2^2 \leq 1 \ \forall \ i=1,2,3,\ldots, p
\end{array}
\end{array}
\end{align}
where $\|C^T\|_{1,2}^2= \sum \limits_{i=1}^p\|[C^T]_i\|_1^2=\sum_{i=1}^p\big(\sum_{j=1}^kC_{ij}\big)^2$ is the $\ell_{1,2}$ norm of $C^T$ which ensures group sparsity in the resultant $C$ matrix and $[C^T]_i$ is the $i$-th row of matrix $C$. For a high value of $\lambda$, the loading matrix is observed to be orthogonal, more details are presented in the experiment section. We further relax the problem by introducing the term $\frac{\alpha}{2}||C \tilde X-X||_F^2$ with $\alpha > 0$, instead of solving the constraint $X=C \tilde{X}$. Note that this relaxation can be made tight by choosing sufficiently large or iteratively increasing $\alpha$. Now the original problem can be approximated as:
\begin{align}\label{Main formulation FGC}
\begin{array}{ll}
\underset{{\tilde{X}}, {C}}{\textsf{minimize}} &
\begin{array}{c}
\hspace{-1em}-\gamma \text{log det}(C^T\Theta C +J)+\text{tr}(\tilde{X}^{T}C^T\Theta C\tilde{X}) +\frac{\alpha}{2}||C \tilde X-X||_F^2+\frac \lambda 2 \|C^T\|_{1,2}^2
\end{array}\\
\textsf{subject to} & \begin{array}[t]{l}
 C \geq 0, \ \|[C^T]_i\|_2^2 \leq 1 \ \forall \ i=1,2,3,\ldots, p
\end{array}
\end{array}
\end{align}


The problem \eqref{Main formulation FGC} is a multi-block non-convex optimization problem. We develop an iterative algorithm based on the block successive upper bound minimization (BSUM) technique \cite{article, 7547360}. Collecting the variables as $(C\in \mathbb{R}_{+}^{p\times k},\tilde{X}\in \mathbb{R}^{k\times n})$, we develop a block MM-based algorithm which updates one variable at a time while keeping the other fixed.

\subsection{Update of \texorpdfstring{$C$}{}}
Treating $C$ as a variable while fixing $\tilde X$, and ignoring the term independent of $C$, we obtain the following sub-problem for $C$:

\begin{align}\label{UpdateC123}
\begin{array}{ll}
\underset{{C}}{\textsf{minimize}} &
\begin{array}{c}
 f(C) =-\gamma \text{log det}(C^T\Theta C +J)+\frac \alpha 2 \|C\tilde{X} - X\|^2_F +\text{tr}(\tilde{X}^{T}C^T\Theta C\tilde{X})+ \frac \lambda 2 \|C^T\|_{1,2}^2
\end{array}\\
\textsf{subject to} & \begin{array}[t]{l}
 C\geq 0,\ \|[C^T]_i\|_2^2 \leq 1 \ \forall \ i=1,2,3,\ldots, p
\end{array}
\end{array}
\end{align}
 To rewrite the problem \eqref{UpdateC123} simply, we have defined a set $\mathcal{S}_c$ as:
 \begin{align}\label{Loading matrix-set}
\mathcal{S}_C =\left\{ C \in \mathbb{R}^{p \times k}|\ C \geq 0, \ \|[C^T]_i\|_2^2 \leq 1 \ \forall \ i=1,2,3,\ldots, p \right\}
\end{align}
where $[C^T]_i$ is the $i$-th row of loading matrix $C$. Note that the set $\mathcal{S}_C$ is a closed and convex set. Using the set $\mathcal{S}_c$, problem \eqref{UpdateC123} can be rewritten as:
 \begin{align}\label{UpdateC1234}
\hspace {-2em}\underset{C \in \mathcal{S}_c}{\text{\text{minimize}}} \hspace{1em}f(C)=-\gamma \text{log det}(C^T\Theta C +J)+\frac \alpha 2 \|C\tilde{X} - X\|^2_F +\text{tr}(\tilde{X}^{T}C^T\Theta C\tilde{X})+ \frac \lambda 2 \|C^T\|_{1,2}^2
\end{align}
\begin{Lem1} \label{trace1}
 $\text{tr}(\tilde{X}^{T}C^T\Theta C\tilde{X})$ is a convex function in loading matrix $C$. 
\end{Lem1}

\begin{proof}
 Since $\Theta$ is a positive semi-definite matrix and using Cholesky decomposition, we can write $\Theta=M^TM$. Now, consider the term: 
\begin{align}
\text{tr}(\tilde{X}^{T}C^T\Theta C\tilde{X})
&= \text{tr}(Y^T \Theta Y) = \text{tr}(Y^T M^TM Y)= \|MY\|_F^2 
\end{align}
Frobenius norm is a convex function so, $\|MY\|_F^2$ is convex function in $Y$ and $Y=C \tilde{X}$ which is a linear function of $C$ so it is a convex function in $C$ also.
\end{proof}

\begin{Lem1}\label{strictly convex}
 The function $f(C)$ in \eqref{UpdateC1234} is strictly convex.
\end{Lem1}

\begin{proof}
$\text{log det}(\cdot)$ and $\text{trace}(\cdot)$ are convex functions and proof are in Lemma \ref{convex} and \ref{trace1} respectively, also Frobenius and $\ell_{1,2}^2$ norm are convex functions. Consider the term
$ \|C^T\|_{1,2}^2=\sum \limits_{i=1}^p\|[C^T]_i\|_1^2 \ > \ 0 $
which implies that $f(C)$ in \eqref{UpdateC1234} is strictly convex function.
\end{proof}
\begin{Lem1}\label{LipshitzforC}
The function $f(C)$ is $L$-Lipschitz continuous gradient function where $L=\max(L_1,L_2, L_3$, $L_4)$ with $L_1,L_2, L_3$, $L_4$ the Lipschitz constants of $-\gamma \text{log det}(C^T\Theta C +J)$, $\text{tr}(\tilde{X}^{T}C^T\Theta C\tilde{X})$, $\|C\tilde{X} - X\|^2_F$, $\frac \lambda 2 \|C^T\|_{1,2}^2$ respectively.
\end{Lem1}

\begin{proof}
The detailed proof is deferred to Appendix \ref{Lipschitz}.
\end{proof}

The function $f(C)$ in \eqref{UpdateC1234} is $L$-smooth, strictly convex and differentiable function. Also, the constraint $C \in \mathcal{S}_c$ together makes the problem \eqref{UpdateC1234} a convex optimization problem. By using \eqref{majorization}, the majorised function for $f(C)$ at $C^{(t)}$ is:
\begin{equation}\label{majorizedfunction}
 g(C|C^{(t)})=f(C^{(t)})+(C-C^{(t)})\nabla f(C^{(t)})+\frac{L}{2}||C-C^{(t)}||^2 
\end{equation}
After ignoring the constant term, the majorized problem of \eqref{UpdateC1234} is
\begin{align}\label{UpdateC1FGC}
 \underset{C \in \mathcal{S}_c}{\text{\text{minimize}}} \quad \frac{1}{2}C^TC-C^TA
\end{align}

\noindent where $A=C^{(t)}-\frac{1}{L}\nabla f(C^{(t)})$ and
$ \nabla f(C^{(t)})= -2\gamma \Theta C^{(t)}(C^{(t)^T}\Theta C^{(t)}+J)^{-1}+\alpha \left(C^{(t)}\tilde{X} - X\right)\tilde{X}^{T} +2\Theta C^{(t)}\tilde{X}\tilde{X}^{T}+\lambda C^{(t)}\pmb{1}_{k \times k}$ where $ \pmb 1_{k \times k} $ is all ones matrix of dimension ${k \times k}$. 
\begin{Lem1}\label{KKTcondition}
By using the KKT optimality condition we can obtain the optimal solution of \eqref{UpdateC1FGC} as
\begin{equation}\label{eqn:C}
 C^{(t + 1)} = \left(C^{(t)} - \frac{1}{L}\nabla f\left(C^{(t)}\right)\right)^+
 \end{equation}
where $(X_{ij})^{+}=\max(\frac{X_{ij}}{\|[X_{T}]^{i}\|_{2}},0)$ and $[X_{T}]^{i}$ is the $i$-th row of matrix $X$.
\end{Lem1}

\begin{proof}
The detailed proof is deferred to Appendix \ref{LagrangianKKT} .
\end{proof}

\subsection{Update of \texorpdfstring{$\pmb{\tilde{X}}$}{} }
By fixing $C$, we obtain the following sub-problem for $\tilde{X}$:
\begin{align}\label{updatetildeX}
 \underset{{\tilde{X}}}{\text{\text{minimize}}} \qquad & f(\tilde{X})=\text{tr}(\tilde{X}^TC^T\Theta C \tilde{X}) + \frac \alpha 2 \|C\tilde{X} - X\|^2_F
\end{align}

 


\begin{Lem1}
 Problem \eqref{updatetildeX} ia a convex optimization problem.
\end{Lem1}

\begin{proof}
 The first term in objective function of \eqref{updatetildeX} $\text{trace}(\cdot)$ is convex function in $\tilde{X}$ and proof is similar to proof of Lemma \ref{trace1}. Also, Frobenius norm is convex function so overall objective function of \eqref{updatetildeX} is convex function and overall problem is convex optimization problem.
\end{proof}

Problem \eqref{updatetildeX} is a convex optimization problem, we get the closed form solution by setting the gradient to zero
\begin{equation}
2C^T\Theta C \tilde{X} + \alpha C^T(C\tilde{X} - X) = 0,
\end{equation}

we get
\begin{equation}\label{eqn:X12}
 \tilde{X}^{t+1} =\left(\frac 2 \alpha C^{T}\Theta C + C^{T}C\right)^{-1} C^TX
\end{equation}

\begin{rem}
In the update of $\tilde{X}$, if taking the inverse is demanding, one can use gradient descent type update for finding $C$. Using gradient descent, the update rule of $\tilde{X}$ is
\begin{equation}\label{eqn:X1}
 \tilde{X}^{t+1} =\tilde{X}^{t}-\eta \nabla f(\tilde{X})
\end{equation}
where, $\eta$ is the learning rate and $\nabla f(\tilde{X})=2C^T\Theta C \tilde{X} + \alpha C^T(C\tilde{X} - X)$
\end{rem}

\begin{algorithm}[H]\label{Algorithm1}
\SetAlgoLined
\SetAlCapFnt{\footnotesize}
\SetAlCapNameFnt{\footnotesize}
 \caption{\textsf{FGC Algorithm}} 
 \KwIn{$\mathcal{G}(X,\Theta), \alpha, \gamma, \lambda$}

  $t \leftarrow 0;$ \\
		\While {stopping criteria not met}{
          Update $C^{t+1}$ and $\tilde{X}^{t+1}$ as in \eqref{eqn:C} and \eqref{eqn:X12} respectively.\\
   $t \leftarrow t+1;$ \\
  }
   \KwOut{$C$, $\Theta_c$, and $\tilde{X}$}
\end{algorithm}

Algorithm \ref{Algorithm1} summarizes the implementation of feature graph coarsening (FGC) method. The worst case computational complexity is $\mathcal{O}(p^{3})$ which is due to the matrix multiplication in the gradient of $f(C)$ in \eqref{eqn:C}.
\begin{thm}\label{convergence}
 The sequence $\{ C^{(t)}, \tilde{X}^{(t)}\}$ generated by Algorithm \ref{Algorithm1}
 converges to the set of Karush–Kuhn–Tucker (KKT) points of Problem \eqref{Main formulation FGC}.
\end{thm}

\begin{proof}
The detailed proof is deferred to Appendix \ref{convergence proof}.
\end{proof}

\begin{thm}\label{epsilon-thm}
 The coarsened graph data $\G_c(\Theta_c ,\tilde{X})$ learned from the FGC algorithm is $\epsilon$ similar to the original graph data $\G(\Theta, X)$, i.e., there exist an $ 0 \leq \epsilon \leq 1$ such that 

\begin{equation}
 (1-\epsilon)\|X\|_{\Theta} \leq \|\tilde{X}\|_{\Theta_c} \leq (1+\epsilon)\|X\|_{\Theta}
 \end{equation}
\end{thm}

Note that $\epsilon$- similarity also indicate similarity in the Dirichlet energies of the $\G(\Theta ,{X})$ and $\G_c(\Theta_c ,\tilde{X})$, as $\|X\|_{\Theta}^2=\text{tr}(X^T\Theta X)$ and $\|\tilde{X}\|_{\Theta_c}^2=\text{tr}(\tilde{X}^T\Theta_c\tilde{X})$.

\begin{proof}
The detailed proofs of Theorem \eqref{epsilon-thm} are deferred to the Appendix \ref{Dritchletenergy}.
\end{proof}

\subsection{Interpretation of the proposed formulation and the FGC Algorithm}

The proposed FGC algorithm \ref{Algorithm1} summarizes a larger graph $\mathcal{G}$ into a smaller graph $\mathcal{G}_c$. The loading matrix variable is simply a mapping of nodes from the set of nodes in $G$ to nodes in $G_c$, i.e., $\pi : V \rightarrow \tilde{V}$. In order to have a balanced mapping the loading matrix $C$ should satisfy the following properties:
\begin{enumerate}
 \item Each node of original graph $\mathcal{G}$ must be mapped to a supernode of coarsened graph $\mathcal{G}_c$ implying that the cardinality of rows should not be zero, $\|[C^T]_i\|_0 \neq 0;\forall\; i=1,2,\dots,p$ where $[C^T]_i$ is the $i$-th row of loading matrix $C$.
 \item In each supernode, there should be
 at least one node of the original graph $G$ should be mapped to a super-node of the coarsened graph, also known as a supernode of $G_c$. Which requires the cardinality of columns of $C$ be greater than equal to 1, i.e., $\|C_i\|_0 \geq 1$ where $C_i;\forall\; i=1,2,\dots,k$ is the $i$-th column of loading matrix $C$.
 \item A node of the original graph should not be mapped to more than one supernode implying the columns of $C$ be orthogonal, i.e., $\langle C_i, C_j \rangle=0$. Furthermore, the orthogonality of columns $\langle C_i, C_j \rangle=0$ clubbed with positivity of elements of $C$, implies that the rows $[C^T]_i\;\forall\; i=1,2,\dots,p$ should have only one nonzero entry, i.e., $\|[C^T]_i\|_0=1$. And in order to make sure that the $\C^T\Theta\C$ is a Laplacian matrix, we need the nonzero elements of $C$ to be $1$.
\end{enumerate}

The proposed formulation and the FGC algorithm manage to learn a balanced mapping, i.e., the loading matrix $C$ satisfies the aforementioned properties. Let us have a re-look at the proposed optimization formulation $C$:
\begin{align}\label{originalC}
\begin{array}{ll}
\underset{{C}}{\textsf{minimize}} &
\begin{array}{c}
\frac{\lambda}{2}\|C^T\|_{1,2}^2 -\gamma \log \det (C^T\Theta C+J)+\frac{\alpha}{2}||C \tilde X-X||_F^2
\end{array}\\
\textsf{subject to} & \begin{array}[t]{l}
 \ C \in \mathcal{S}_c
\end{array}
\end{array}
\end{align}
Using $\|C^T\|_{1,2}^2=\sum_{i=1}^p\big(\sum_{j=1}^kC_{ij}\big)^2=\|C\|_F^2+\sum_{i \neq j}\langle C_i , C_j \rangle$ ${\rm for}$ $i,j=1,2,...k$, we can rewrite problem \eqref{originalC} as
\begin{align}\label{original12}
\begin{array}{ll}
\underset{{C}}{\textsf{minimize}} &
\begin{array}{c}
\frac{\lambda}{2}(\|C\|_F^2+\sum_{i \neq j}\langle C_i , C_j \rangle) -\gamma \log \det (C^T\Theta C+J)+\frac{\alpha}{2}||C \tilde X-X||_F^2
\end{array}\\
\textsf{subject to} & \begin{array}[t]{l}
 \ C \in \mathcal{S}_c
\end{array}
\end{array}
\end{align}

Note that $\Theta \in \mathcal{S}_{\Theta}$ is the Laplacian of a connected graph with rank $p-1$. We aim here to learn $C\in \mathbb{R}_+^{p\times k}$ which maps a set of $p$ nodes to $k$ nodes, such that $C^T\Theta C$ is a Laplacian matrix of a connected graph with $k$ nodes, which implies that rank$(C^T\Theta C+J_{k\times k}) =k$. The $\log \det (\cdot)$ requires that the matrix $C^T\Theta C+J\in \mathbb{R}^{k\times k}$ is always be a full rank matrix, i.e., $k$. Now, we will investigate the importance of each term in the optimization problem \eqref{original12} below:
\begin{enumerate}
 \item The trivial solution of all zero $C=\textbf{0}_{p \times k}$ will make the term $(C^T\Theta C +J)$ rank deficient and the $\log\det(\cdot)$ term become infeasible and thus ruled out and in toy example, $C1$ is ruled out, for example, $C1$ in Fig \ref{loading matrix}. 
 \item Next, any $C$ with zero column vector i.e. $C_i=0$ $\forall i=1,2,\ldots p$ will lead to a coarsened graph of size less than $k$, and thus again $(C^T\Theta C+J)$ will be rank deficient, so this solution is also ruled out, for example, $C2$ in Fig \ref{loading matrix}. 
 \item The minimization of $\|C \tilde X-X\|_F^2=\sum_{i=1}^{p}( [C^T]_i\tilde{X}-X_i ) ^2$ ensures that no row of $C$ matrix will be zero, for example $C3$ in Fig \ref{loading matrix} is ruled out. 
 \item Next, as $C \geq 0$, $C\neq \textbf{0}_{p \times k} $, and from the property of Frobenius norm it implies that $\|C\|_F^2 \neq 0$, thus the only possibility to minimize \eqref{original12} is to get $C\geq 0$ such that
$\sum_{i \neq j}\langle C_i , C_j \rangle = 0$. This implies that columns of loading matrix $C$ are orthogonal to each other, and $C^TC=\text{block}(\mathbf{d})$ is a block diagonal matrix which implies that $\langle C_i, C_j \rangle=0$, for example C4 in Fig \ref{loading matrix} is ruled out.
 \item The orthogonality of columns combined with $C \geq 0$ implies that in each row there is only one non-zero entry and the rest entries are zero which finally implies that $\|[C^T]_i\|_0=1$.

\end{enumerate}
Summarizing, the solution of \eqref{original12} is $C\in \mathbb{R}_+^{p\times k}$ of rank $k$ with orthogonal columns, and rows and columns are having maximum and minimum cardinality 1, respectively, i.e., $||C_i||_0\geq 1$ and $||[C^T]_i||_0= 1$ which satisfies all the properties for a balanced mapping. Finally, each row of the loading matrix has cardinality 1, and $\|[C^T]_i\|_2 \leq 1$ ensures that each row of the loading matrix has only one non-zero entry and, i.e., 1 and the rest of entries in each row is zero.

\begin{figure}{!htb}
 $$
C1= \begin{bmatrix}
 0 & 0 & 0 \\
 0 & 0 & 0\\ 
 0 & 0 & 0 \\
 0 & 0 & 0\\
 0 & 0 & 0
\end{bmatrix},
C2= \begin{bmatrix}
 0 & 1 & 0 \\
 0 & 1 & 0\\ 
 0 & 1 & 0 \\
 0 & 0 & 1\\
 0 & 0 & 1
\end{bmatrix}, 
C3= \begin{bmatrix}
 0 & 0 & 0 \\
 1 & 0 & 0\\ 
 1 & 0 & 0 \\
 0 & 1 & 0\\
 0 & 0 & 1
\end{bmatrix},
C4= \begin{bmatrix}
 1 & 0 & 1 \\
 1 & 0 & 0\\ 
 1 & 0 & 0 \\
 0 & 1 & 0\\
 0 & 0 & 1
\end{bmatrix},
C5= \begin{bmatrix}
 1 & 0 & 0 \\
 1 & 0 & 0\\ 
 1 & 0 & 0 \\
 0 & 1 & 0\\
 0 & 0 & 1
\end{bmatrix}
$$
\caption{Some possible realisations of the loading matrix for the toy example: $C1,C2, C3$, and $C4$ are not balanced mapping, while $C5$ is an example of balanced mapping.}
 \label{loading matrix}
\end{figure}



\section{Proposed Graph Coarsening without Features (GC) Algorithm}
In this section, we learn the coarsened graph without considering the feature matrix. By using $\Theta_c=C^T\Theta C$, the two variable optimization problem \eqref{Main formulation without X1} is equivalent to a single variable optimization problem as:
\begin{align}\label{Main formulation without X}
\begin{array}{ll}
\underset{{C}}{\textsf{minimize}} &
\begin{array}{c}
\hspace{-1em}f(C)=-\gamma \text{log det}(C^T\Theta C +J) +\frac \lambda 2 \|C^T\|_{1,2}^2
\end{array}\\
\textsf{subject to} & \begin{array}[t]{l}
 C \in \mathcal{S}_c
\end{array}
\end{array}
\end{align}
where the set $\mathcal{S}_c$ is defined in \eqref{Loading matrix-set}. 
\begin{Lem1}
The function $f(C)$ in \eqref{Main formulation without X} is strictly convex function.
\end{Lem1}

\begin{proof}
The proof is similar to the proof of lemma \ref{strictly convex}. 
\end{proof}
\begin{Lem1}
The function $f(C)$ is $L$-Lipschitz continuous gradient function where $L=\max(L_1,L_2)$ with $L_1,L_2$ the Lipschitz constants of $-\gamma \text{log det}(C^T\Theta C +J)$, $\sum \limits_{i=1}^p \|[C^T]_i\|_1^2$ respectively.
\end{Lem1}

\begin{proof}
The proof is similar to the proof of lemma \ref{LipshitzforC}.
\end{proof}

We solve this problem using the BSUM framework. The function $f(C)$ is strictly convex, differentiable and $L$-smooth. By using \eqref{majorization}, the majorised function for $f(C)$ at $C^{(t)}$ 
\begin{equation}
 g(C|C^{(t)})=f(C^{(t)})+(C-C^{(t)})\nabla f(C^{(t)})+\frac{L}{2}||C-C^{(t)}||^2 
\end{equation}
 After ignoring the constant term, the majorized problem of \eqref{Main formulation without X} is
\begin{align}\label{UpdateC12}
 \underset{C \in \mathcal{S}_c}{\text{\text{minimize}}} \quad \frac{1}{2}C^TC-C^TA
\end{align}
where $A=C^{(t)}-\frac{1}{L}\nabla f(C^{(t)})$ and
$ \nabla f(C^{(t)})= -2\gamma \Theta C^{(t)}(C^{(t)^{T}}\Theta C^{(t)}+J)^{-1}+ \lambda C^{(t)}\pmb{1}_{k \times k}$ where $ \pmb 1_{k \times k} $ is all ones matrix of dimension ${k \times k}$. 
\begin{Lem1}
By using the KKT optimality condition we can obtain the optimal solution of \eqref{UpdateC12} as
\begin{equation}\label{eqn:C1}
 C^{(t + 1)} = \left(C^{(t)} - \frac{1}{L}\nabla f\left(C^{(t)}\right)\right)^+
 \end{equation}
where $(X_{ij})^{+}=\max(\frac{X_{ij}}{\|[X^T]_i\|_{2}},0)$ and $[X^T]_i$ is the $i$-th row of matrix $X$. 
\end{Lem1}

\begin{proof}
The proof is similar to the proof of Lemma \ref{KKTcondition}.
\end{proof}

\begin{algorithm}[H]\label{Algorithm21}
\SetAlgoLined
\SetAlCapFnt{\footnotesize}
\SetAlCapNameFnt{\footnotesize}
 \caption{\textsf{Graph Coarsening (GC) Algorithm}} 
 \KwIn{$\mathcal{G}(X,\Theta)$, $\gamma$, $\lambda$}
  $t \leftarrow 0;$ \\
		\While {stopping criteria not met}{
           Update $C^{t+1}$ as in \eqref{eqn:C1}\\
   $t \leftarrow t+1;$ \\
  }
   \KwOut{$C$ and $\Theta_c$}
\end{algorithm}

\subsubsection*{Two stage optimization problem}
In graph coarsening, when we encounter graph data containing features, we can compute coarsened feature matrix $\tilde{X}$ using $\tilde{X}=PX$ directly as mentioned by \cite{loukas2019graph} where $P$ are coarsening matrix. This learns a coarsened graph with high Dirichlet energy. As we know that for a smooth graph, its Dirichlet energy should be low. So to impose smoothness property in our reduced-graph, we can use the following optimization problem: 
\begin{align}\label{GC with feature}
\begin{array}{ll}
\underset{{X_c}}{\textsf{minimize}} &
\begin{array}{c}
\hspace{-1em} f(X_c)=\|X_c-\tilde{X}\|_F^2+\text{tr}(X_c^{T} \Theta_c X_c) 
\end{array}
\end{array}
\end{align}	
where, $X_c$ is new learned smooth feature of coarsened graph and $\text{tr}(X_c^{T} \Theta_c X_c)$ is smoothness of resulting coarsened graph.
\\
Problem \eqref{GC with feature} is a convex and differentiable function. We get the closed form solution by setting the gradient w.r.t $X_c$ to zero. 
\begin{equation}
2\Theta_c X_c + 2(X_c-\tilde{X}) = 0,
\end{equation}
we get,
\begin{equation}\label{eqn:X}
 X_c=(\Theta_c+I)^{-1}\tilde{X}
\end{equation}
Now, we can extend GC (proposed) to a two-stage optimization problem where we first compute $C$ and $\Theta_c$, then use it to compute $X_c$, the feature matrix of a coarsened graph. 


\begin{algorithm}[H]\label{Algorithm3}
\SetAlgoLined
\KwInput{$\mathcal{G}(X,\Theta), \gamma, \lambda$}
 Set $C \leftarrow GC(\mathcal{G}(X,\Theta), \gamma, \lambda)$;\\
 Set $X_c \leftarrow (\Theta_c+I)^{-1}\tilde{X}$;\\
 \textbf{Output} $C$ and $X_c$
\caption{Two-Stage Optimization Algorithm}
\end{algorithm}

\section{Proposed Featured Graph Coarsening with Feature Dimensionality Reduction(FGCR)}
In this section, we develop a block MM-based algorithm for graph coarsening with feature reduction. In particular, we propose to solve \eqref{Main formulation with reduced features12} by introducing the quadratic penalty for $X=C\tilde{X}$ and $\tilde{X}=WH$, we aim to solve the following optimization problem:

\begin{align}\label{Main formulation with reduced features}
\begin{array}{ll}
\underset{{W},{H}, {C}}{\textsf{minimize}} &
\begin{array}{c}
\hspace{-1em}-\gamma \text{log det}(C^T\Theta C +J)+\text{tr}(W^TC^T\Theta CW) +\frac{\alpha}{2}||CWH-X||_F^2 +\frac \lambda 2 \|C^T\|_{1,2}^2
\end{array}\\
\textsf{subject to} & \begin{array}[t]{l}
 C \in \mathcal{S}_c
\end{array}
\end{array}
\end{align}

where the set $\mathcal{S}_c$ is defined in \eqref{Loading matrix-set}. 
The problem \eqref{Main formulation with reduced features} is a multi-block non-convex optimization problem. We develop an iterative algorithm based on the block successive upper bound minimization (BSUM) technique \cite{article, 7547360}. Collecting the variables as $(C\in \mathbb{R}_{+}^{p\times k},W \in \mathbb{R}^{k\times d}, H \in \mathbb{R}^{d \times n})$, we develop a block MM-based algorithm which updates one variable at a time while keeping the other fixed.

\subsection{Update of C}
Treating $C$ as a variable and fixing $W$ and $H$, we obtain the following sub-problem for $C$:
\begin{align}\label{UpdateC}
\hspace {-2em}\underset{C \in \mathcal{S}_c}{\text{\text{minimize}}} \hspace{1em}f(C)=-\gamma \text{log det}(C^T\Theta C +J)+\text{tr}(W^TC^T\Theta CW) +\frac{\alpha}{2}||CWH-X||_F^2 +\frac \lambda 2 \|C^T\|_{1,2}^2
\end{align}
\begin{Lem1}
The function $f(C)$ in \eqref{UpdateC} is strictly convex function.
\end{Lem1}

\begin{proof}
The proof is similar to the proof of lemma \ref{strictly convex}. 
\end{proof}

\begin{Lem1}
The function $f(C)$ is $L$-Lipschitz continuous gradient function where $L=\max(L_1,L_2, L_3,L_4)$ with $L_1,L_2, L_3,L_4$ the Lipschitz constants of $-\gamma \text{log det}(C^T\Theta C +J)$, $\text{tr}(W^TC^T\Theta CW)$ , $\|CWH - X\|^2_F$, $\sum \limits_{i=1}^p \|[C^T]_i\|_1^2$ respectively.
\end{Lem1}

\begin{proof}
The proof is similar to the proof of Lemma \ref{LipshitzforC}.
\end{proof}
The function $f(C)$ in \eqref{UpdateC} is convex, differentiable and $L$-smooth. Also, the constraint $C \in \mathcal{S}_c$ together makes the problem \eqref{UpdateC} a convex optimization problem. By using \eqref{majorization}, the majorised function for $f(C)$ at $C^{(t)}$ is
\begin{equation}
 g(C|C^{(t)})=f(C^{(t)})+(C-C^{(t)})\nabla f(C^{(t)})+\frac{L}{2}||C-C^{(t)}||^2 
\end{equation}
 After ignoring the constant term, the majorised problem of \eqref{UpdateC} is
\begin{align}\label{UpdateC1}
 \underset{C \geq 0}{\text{\text{minimize}}} \quad \frac{1}{2}C^TC-C^TA
\end{align}
where $A=C^{(t)}-\frac{1}{L}\nabla f(C^{(t)})$ and
$ \nabla f(C^{(t)})= -2\gamma\Theta C^{(t)}(C^{(t)^{T}}\Theta C^{(t)}+J)^{-1}+\alpha \left(C^{(t)}WH - X\right)H^TW^T +2\Theta C^{(t)}WW^T+ \lambda C^{(t)}\pmb{1}_{k \times k}$ where $ \pmb 1_{k\times k} $ is all ones matrix of dimension ${k \times k}$.

\begin{Lem1}
By using KKT optimality condition we can obtain the optimal solution of \eqref{UpdateC1} as
\begin{equation}\label{eqn:C3}
 C^{(t + 1)} = \left(C^{(t)} - \frac{1}{L}\nabla f\left(C^{(t)}\right)\right)^+
 \end{equation}
where $(X_{ij})^{+}=\max(\frac{X_{ij}}{\|[X^T]_i\|_{2}},0)$ and $[X^T]_i$ is the $i$-th row of matrix $X$.
\end{Lem1}

\begin{proof}
The proof is similar to the proof of Lemma \ref{KKTcondition}.
\end{proof}
\subsection{Update of \texorpdfstring{$\pmb{W}$}{}}
By fixing $C$ and $H$, we obtain the following sub-problem for $W$:
\begin{align}\label{W}
 \underset{W}{\text{\text{minimize}}} \qquad & f(W)=\text{tr}(W^TC^T\Theta CW) +\frac{\alpha}{2}||CWH-X||_F^2
\end{align}

\begin{Lem1}
The function $f(W)$ in \eqref{W} is $L$-Lipschitz continuous gradient function where $L=\max(L_1,L_2)$ with $L_1,L_2$ the Lipschitz constants of $\text{tr}(W^TC^T\Theta CW)$ , $\|CWH - X\|^2_F$ respectively.
\end{Lem1}

\begin{proof}
The proof is similar to the proof of Lemma \ref{LipshitzforC}.
\end{proof}

The function $f(W)$ is convex, differentiable and $L$-smooth. By using \eqref{majorization}, the majorised function for $f(W)$ at $W^{(t)}$ is
\begin{equation}
 g(W|W^{(t)})=f(W^{(t)})+(W-W^{(t)})\nabla f(W^{(t)})+\frac{L}{2}||W-W^{(t)}||^2 
\end{equation}
 After ignoring the constant term, the majorised problem of \eqref{W} is
\begin{align}\label{UpdateC.1}
 \underset{W}{\text{\text{minimize}}} \quad \frac{1}{2}W^TW-W^TA
\end{align}
where $A=W^{(t)}-\frac{1}{L}\nabla f(W^{(t)})$ and
$ \nabla f(W^{(t)})= 2C^T\Theta CW^{(t)}HH^T+\alpha C^T \left(CW^{(t)}H - X\right)H^T $.
\begin{Lem1}
By using the KKT optimality condition we can obtain the optimal solution of \eqref{UpdateC.1} as
\begin{equation}\label{eqn:W}
 W^{(t + 1)} = \left(W^{(t)} - \frac{1}{L}\nabla f\left(W^{(t)}\right)\right)
 \end{equation}
\end{Lem1}

\subsection{Update of \texorpdfstring{$\pmb{H}$}{} }
By fixing $C$ and $H$, we obtain the following sub-problem for $W$:
\begin{align}\label{H}
 \underset{H}{\text{\text{minimize}}} \qquad & f(H)=\frac{\alpha}{2}||CWH-X||_F^2
\end{align}
\begin{Lem1}
The function $f(H)$ is $L$-Lipschitz continuous gradient function where $L$ is the Lipschitz constants of $\|CWH - X\|^2_F$.
\end{Lem1}

\begin{proof}
The proof is similar to the proof of Lemma \ref{LipshitzforC}.
\end{proof}
The function $f(H)$ is convex, differentiable and $L$-smooth. By using \eqref{majorization}, the majorised function for $f(H)$ at $H^{(t)}$ is
\begin{equation}
 g(H|H^{(t)})=f(H^{(t)})+(H-H^{(t)})\nabla f(H^{(t)})+\frac{L}{2}||H-H^{(t)}||^2 
\end{equation}
 After ignoring the constant term, the majorised problem of \eqref{H} is
\begin{align}\label{UpdateC.12}
 \underset{H}{\text{\text{minimize}}} \quad \frac{1}{2}H^TH-H^TA
\end{align}
where $A=H^{(t)}-\frac{1}{L}\nabla f(H^{(t)})$ and
$ \nabla f(H^{(t)})= \alpha W^TC^T \left(CWH^{(t)} - X\right) $.
\begin{Lem1}
By using the KKT optimality condition we can obtain the optimal solution of \eqref{UpdateC.12} as
\begin{equation}\label{eqn:H}
 H^{(t + 1)} = \left(H^{(t)} - \frac{1}{L}\nabla f\left(H^{(t)}\right)\right)
 \end{equation}
\end{Lem1}
\begin{proof}
The proof is similar to the proof of Lemma \ref{KKTcondition}
\end{proof}
\begin{algorithm}[H]\label{Algorithm2}
\SetAlgoLined
\SetAlCapFnt{\footnotesize}
\SetAlCapNameFnt{\footnotesize}
 \caption{\textsf{Featured Graph Coarsening with Reduction (FGCR) Algorithm}} 
 \KwIn{$\mathcal{G}(X,\Theta), \alpha, \gamma, \lambda$}
  $t \leftarrow 0;$ \\
		\While {stopping criteria not met}{
          Update $C^{t+1}$ as in \eqref{eqn:C3};\\
		Update $W^{t+1}$ as in \eqref{eqn:W};\\
  Update $H^{t+1}$ as in \eqref{eqn:H};\\
   $t \leftarrow t+1;$ \\
  }
   \KwOut{$C$, $W$, and $H$}
\end{algorithm}

\begin{thm}\label{convergence1}
 The sequence $\{ C^{(t)}, W^{(t)}, H^{(t)}\}$ generated by Algorithm \ref{Algorithm2}
 converges to the set of Karush–Kuhn–Tucker (KKT) points of Problem \eqref{Main formulation with reduced features}.
\end{thm}

\begin{proof}
The proof is similar to the proof of Theorem \ref{convergence}.
\end{proof}

\section{Connection of graph coarsening with clustering and community detection}

It is important here to highlight the distinction between (i) Clustering \cite{ng2001spectral, dhillon2007weighted} (ii) Community detection \cite{fortunato2010community}, and (iii) graph coarsening approaches. Given a set of data points, the clustering and community detection algorithm aim to segregate groups with similar traits and assign them into clusters. For community detection, the data points are nodes of a given network. But these methods do not answer how these groups are related to each other. On the other hand coarsening segregates groups with similar traits and assigns them into supernodes, in addition, it also establishes how these supernodes are related to each other. It learns the graph of the supernodes, the edge weights, and finally the effective feature of each supernode. Thus, the scope of the coarsening method is wider than the aforementioned methods.



The proposed coarsening algorithm can be used for graph clustering and community detection problems. For grouping $p$ nodes into $c$ clusters, we need to perform coarsening from $p$ nodes to $c$ supernodes. For the FGC algorithm, it implies learning a loading matrix of size $p\times c$ and the node supernode mapping reveals the clustering and community structure present in the graph.


Furthermore, we also believe that the overarching purpose of coarsening method goes beyond the clustering and partitioning types of algorithms. Given a large graph with nodes, the FGC method can learn a coarse graph with nodes. A good coarsening algorithm will be a significant step in addressing the computational bottleneck of graph-based machine learning applications. Instead of solving the original problem, solve a coarse problem of reduced size at a lower cost; then lift (and possibly refine) the solution. The proposed algorithms archive this goal by approximating a large graph with a smaller graph while preserving the properties of the original graph. In the experiment section, we have shown the performance of the FGC method by evaluating it with different metrics indicating how well the coarsened graph has preserved the properties of the original graph. The FGC framework is also tested for clustering tasks on real datasets, e.g., Zachary's karate club and the Polblogs dataset. In all the experiments the superior performance of the FGC algorithm against the benchmarks indicates the wider applicability and usefulness of the proposed coarsening framework.

\section{Experiments}
In this section, we demonstrate the effectiveness of the proposed algorithms by a comprehensive set of experiments conducted on both real and synthetic graph data sets. We compare the proposed algorithms by benchmarking against the state-of-the-art methods, Local Variation Edges (LVE) and Local Variation Neighbourhood (LVN), proposed in \cite{loukas2019graph} along with some other pre-existing famous graph coarsening methods like Kron reduction (Kron) \cite{https://doi.org/10.48550/arxiv.1102.2950} and heavy edge matching (HEM) \cite{karypis1998fast}. The baseline method only uses adjacency matrix information for performing coarsening. Once the coarsening matrix is learned, which is the node supernode mapping. It is further used for coarsening the feature matrix as well. The main difference is that the baseline methods do not consider the graph feature matrix to learn coarsening matrix, while the proposed FGC algorithm considers both the graph matrix and the graph feature matrix jointly for it. Throughout all the experiments the proposed algorithms have shown outstanding and superior performance. \\
\noindent \textbf{Datasets:} The graph datasets (p,m,n) where p is the number of nodes, m is the number of edges and n is the number of features, used in the following experiments are mentioned below.
(i) The details of real datasets are as follows:
\begin{itemize}
 \item Cora. This dataset consists of p=2708, m=5278, and n=1433. Hyperparameters ($\lambda$=500, $\alpha$=500, $\gamma$=716.5) used in FGC algorithm. DE of $\mathcal{G}$ is 160963.
 \item Citeseer. This dataset consists of p=3312, m=4536, and n=3703. Hyperparameters ($\lambda$=500, $\alpha$=500, $\gamma$=1851.5) used in FGC algorithms. DE of $\mathcal{G}$ is 238074.
 \item Polblogs. This dataset consists of p=1490, m=16715, and n=5000. Hyperparameters ($\lambda$=500, $\alpha$=500, $\gamma$=2500) used in FGC algorithms. DE of $\mathcal{G}$ is 6113760.
 \item ACM. This dataset consists of p=3025, m=13128, and n=1870. Hyperparameters ($\lambda$=500, $\alpha$=500, $\gamma$=935) used in FGC algorithms. DE of $\mathcal{G}$ is 1654444.
 \item Bunny. This dataset consists of p=2503, m=78292, and n=5000. Hyperparameters ($\lambda$=450, $\alpha$=500, $\gamma$=2500) used in FGC algorithm. DE of $\mathcal{G}$ is 12512526.
 \item Minnesota. This dataset consists of p=2642, m=3304, and n=5000. Hyperparameters ($\lambda$=500, $\alpha$=550, $\gamma$=2500) used in FGC algorithms. DE of $\mathcal{G}$ is 13207844.
 \item Airfoil. This dataset consists of p=4253, m=12289, and n=5000. Hyperparameters ($\lambda$=2000, $\alpha$=600, $\gamma$=2500) used in FGC algorithms. DE of $\mathcal{G}$ is 21269451.
\end{itemize}
(ii) The details of synthetic datasets are as follows:
\begin{itemize}
 \item Erdos Renyi (ER). It is represented as $\mathcal{G}(n,p)$, where $n=1000$ is the number of nodes and $p=0.1$ is probability of edge creation. Hyperparameters ($\lambda$=500, $\alpha$=500, $\gamma$=10) used in FGC algorithms. DE of $\mathcal{G}$ is 4995707.
 \item Barabasi Albert (BA). It is represented as $\mathcal{G}(n,m)$, where $n=1000$ is the number of nodes and $m=20$ edges are preferentially linked to existing nodes with a higher degree. Hyperparameters($\lambda$=500, $\alpha$=500, $\gamma$=1000) used in FGC algorithms . DE of $\mathcal{G}$ is 4989862.
 \item Watts Strogatz (WS). It is represented as $\mathcal{G}(n,k,p)$, where $n=1000$ is the number of nodes, $k=20$ is nearest neighbors in ring topology connected to each node, $p=0.1$ is probability of rewiring edges. Hyperparameters($\lambda$=500, $\alpha$=500, $\gamma$=1000) used in FGC algorithm. DE of $\mathcal{G}$ is 4997509.
 \item Random Geometric Graph (RGG). It is represented as $\mathcal{G}(n,radius)$, where $n=1000$ is number of nodes and $radius=0.1$ is the distance threshold value for an edge creation. Hyperparameters($\lambda$=500, $\alpha$=500, $\gamma$=1000) used in FGC algorithm. DE of $\mathcal{G}$ is 4989722.
\end{itemize}
\vspace{-0.15cm}
The features of Polblogs, Bunny, Minnesota, Airfoil, Erdos Renyi (ER), Watts Strogatz (WS), Barabasi Albert (BA) and Random Geometric Graph (RGG) are generated using $X \sim \mathcal{N}(\mathbf{0}, \Theta^\dagger)$ \eqref{GMRF}, where $\Theta$ is the Laplacian matrix of the given graph as these graphs has no features. Weights for synthetic datasets are generated randomly and uniformly from a range of (1,10).

\subsection{Performance Evaluation for the FGC algorithm}
\noindent \textbf{REE, DE, HE and RE analysis:} We use relative eigen error (REE) defined in \eqref{eig_error}, Dirichlet energy (DE) of $\mathcal{G}_c$ defined in \eqref{DE}, hyperbolic error(HE) defined in \eqref{HE} and reconstruction error(RE) defined in \eqref{RE} as the evaluation metrics to measure spectral similarity, smoothness and $\epsilon$ similarity of coarsened graph $\mathcal{G}_c$. The baseline method only uses adjacency matrix information for performing coarsening. Once the coarsening matrix $P=C^{\dagger}$ is learned, which establishes the linear mapping of the nodes to the super-nodes. The matrix $P$ is used further for the coarsening of the feature matrix as $\tilde{X}=PX$. It is evident in Table \ref{tabel:FGC on DE REE} and \ref{tabel: FGC HE and RE} that the FGC outperforms state-of-the-art algorithms.
\begin{table}
\begin{center}
\begin{tabular}{ |m{1.5cm} | m{2.0em} | m{2.5em}| m{2.1em} |m{2.1em} |m{2.1em} |m{2.4em} | m{2.3em}|m{2.1em}|m{2.1em}|m{2.1em} |m{2.4em} |} 
\hline
\multirow{2}{*}{Dataset} & \multirow{2}{*} {$r$=$\frac{k}{p} $} & 
 \multicolumn{5}{c|}{$\text{REE}(\Theta,\Theta_c,100)$} & \multicolumn{5}{c|}{$\text{DE in $10^4$}$}\\
 \cline{3-12}
 & & FGC & LVN & LVE & Kron & HEM & FGC & LVN & LVE & Kron & HEM\\ 
 \hline
 Cora & 0.7 0.5 0.3 & \textbf{0.04} \textbf{0.051} \textbf{0.058} & 0.33 0.51 0.65 & 0.29 0.53 0.71 & 0.38 0.57 0.74 & 0.38 0.58 0.77 & \textbf{0.75 0.69 0.66} & 10.0 6.10 3.2 & 9.9 5.81 2.8 & 9.1 \textbf{\newline}5.5 \textbf{\newline} 2.7 & 9.1 \textbf{\newline} 5.4\textbf{\newline} 2.4 \\ 
 \hline
 Citeseer & 0.7 0.5 0.3 & \textbf{0.012} \textbf{0.04} \textbf{0.05} & 0.32 0.54 0.72 & 0.29 0.55 0.76 & 0.31 0.54 0.77 & 0.31 0.54 0.80 & \textbf{0.71 0.69 0.59} & 13.0 7.50 3.1 & 14.0 7.10 2.9 & 12.9 \textbf{\newline}7.0 \textbf{\newline}2.7 & 12.9\textbf{\newline} 7.0\textbf{\newline} 2.5\\ 
 \hline
 Poblogs & 0.7 0.5 0.3 & \textbf{0.001} \textbf{0.007} \textbf{0.01} & 0.50 0.73 0.86 & 0.35 0.67 0.96 & 0.42 0.67 0.96 & 0.44 0.70 0.92 & \textbf{3.2 3.0 2.6} & 607 506 302 & 656 468 115 &752 513 132 & 761 373 183\\ 
 \hline
 ACM & 0.7 0.5 0.3 & \textbf{0.002} \textbf{0.034} \textbf{0.036} & 0.38 0.66 0.92 & 0.14 0.42 0.88 & 0.15 0.40 0.85 & 0.15 0.41 0.93 & \textbf{1.7 1.5 0.5} & 72.0 30.0 5.7 & 93.4 43.0 7.5 &94.5 49 \textbf{\newline}8.9 & 94.5 46.1 5.4 \\ 
 \hline
\end{tabular}
\end{center}
 \caption{ This table summarizes the REE and DE values obtained by FGC (proposed), LVN, LVE, Kron and HEM on different coarsening ratios ($r$) for standard real graph datasets. It is evident that FGC (proposed) outperforms state-of-the-art methods significantly.}
\label{tabel:FGC on DE REE}
\end{table}

\begin{table}
\begin{center}
\begin{tabular}{ |m{1.5cm} | m{2.2em} | m{2.2em}| m{2.2em} |m{2,2em} | m{2.2em}|m{2.2em}|m{2.2em}|} 
 \hline
 \multirow{2}{*}{Dataset} & \multirow{2}{*}{$r$=$\frac{k}{p} $} & 
 \multicolumn{3}{c|}{$\text{HE}$} & \multicolumn{3}{c|}{RE in $\log(\cdot)$}\\
 \cline{3-8}
 & & FGC & LVN & LVE & FGC & LVN & LVE\\ 
 \hline
 Cora & 0.7 0.5 0.3 &\textbf{0.72} \textbf{1.18} \textbf{1.71} & 1.39 2.29 2.94 & 1.42 2.37 3.08 & \textbf{1.91} \textbf{2.78} \textbf{3.28} & 2.92 3.63 3.77 & 2.95 3.67 3.79\\ 
 \hline
 Citeseer & 0.7 0.5 0.3 & \textbf{0.85} \textbf{1.05} \textbf{1.80} & 1.68 2.43 3.25 & 1.63
2.40
3.41 & \textbf{1.32} \textbf{1.61} \textbf{2.41}& 2.56 2.87 3.04 & 2.51
2.90
3.04 \\ 
 \hline
 Polblogs & 0.7 0.5 0.3 & \textbf{1.73} \textbf{2.70} \textbf{2.89} & 2.33 2.73 3.07 & 2.39
2.58
3.69 & \textbf{5.1} \textbf{6.2} \textbf{6.3} & 7.27 7.42 7.50 & 7.11
7.42
7.51 \\ 
 \hline
 ACM & 0.7 0.5 0.3 & \textbf{0.45} \textbf{0.98} \textbf{1.86} & 2.13 3.10 4.867 & 1.63
2.55 4.43 & \textbf{2.42} \textbf{3.78} \textbf{4.77} & 5.05 5.35 5.44 & 4.66
5.18
5.42\\ 
 \hline
\end{tabular}
\end{center}
\caption{ This table summarizes the HE and RE values obtained by FGC (proposed), LVN and LVE on different coarsening ratios ($r$) for standard real graph datasets. It is evident that FGC (proposed) outperforms state-of-the-art methods significantly.}
\label{tabel: FGC HE and RE}
\end{table}

\noindent \textbf{Comparison with Deep Learning based Graph Carsening method (GOREN)\cite{cai2021graph}}: We have compared FGC (proposed) against the GOREN, a deep learning-based graph coarsening approach, on real datasets. Due to the unavailability of their code, we compared only REE because it is the only metric they have computed in their paper among REE, DE, RE, and HE. Their results of REE are taken directly from their paper. It is evident in Table \ref{tabel: FGC wit GOREN} that FGC outperforms GOREN.
\begin{table}
\begin{center}
\begin{tabular}{ |m{1.7cm} | m{2.6em} | m{4em}| m{4em} |m{4em} | m{4em}|} 
 \hline
 \multirow{2}{*}{Dataset} & \multirow{2}{*}{$r$=$\frac{k}{p} $} & 
 \multicolumn{4}{c|}{$\text{REE}(\Theta,\Theta_c,100)$} \\
 \cline{3-6}
 & & FGC & G.HEM & G.LVN & G.LVE \\ 
 \hline
 Bunny & 0.7 0.5 0.3 & 0.0167 \textbf{ 0.0392 0.0777 } & 0.258 0.420 0.533 & 0.082 0.169 0.283 & \textbf{0.007} 0.057 0.094 \\ 
 \hline
 Airfoil & 0.7 0.5 0.3 & 0.103 \textbf{ 0.105 0.117 } & 0.279 0.568 1.979 & 0.184 0.364 0.876 & \textbf{0.102} 0.336 0.782 \\ 
 \hline
 Yeast & 0.7 0.5 0.3 & \textbf{0.007 0.011 0.03 } & 0.291 1.080 3.482 & 0.024 0.133 0.458 & 0.113 0.398 2.073\\
 \hline
Minnesota & 0.7 0.5 0.3 & \textbf{0.0577 0.0838 0.0958 }& 0.357 0.996 3.423 & 0.114 0.382 1.572 & 0.118 0.457 2.073 \\
\hline
\end{tabular}
\end{center}
 \caption{This table summarizes the REE values obtained by FGC (proposed), GOREN(HEM), GOREN(LVN), and GOREN(LVE) on different coarsening ratios ($r$) for real graph datasets. It is evident that FGC (proposed) outperforms state-of-the-art methods significantly.}
\label{tabel: FGC wit GOREN}
\end{table}

\begin{figure}
 \centering
 \begin{subfigure}[b]{0.24\textwidth}
 \centering
 \includegraphics[width=\textwidth]{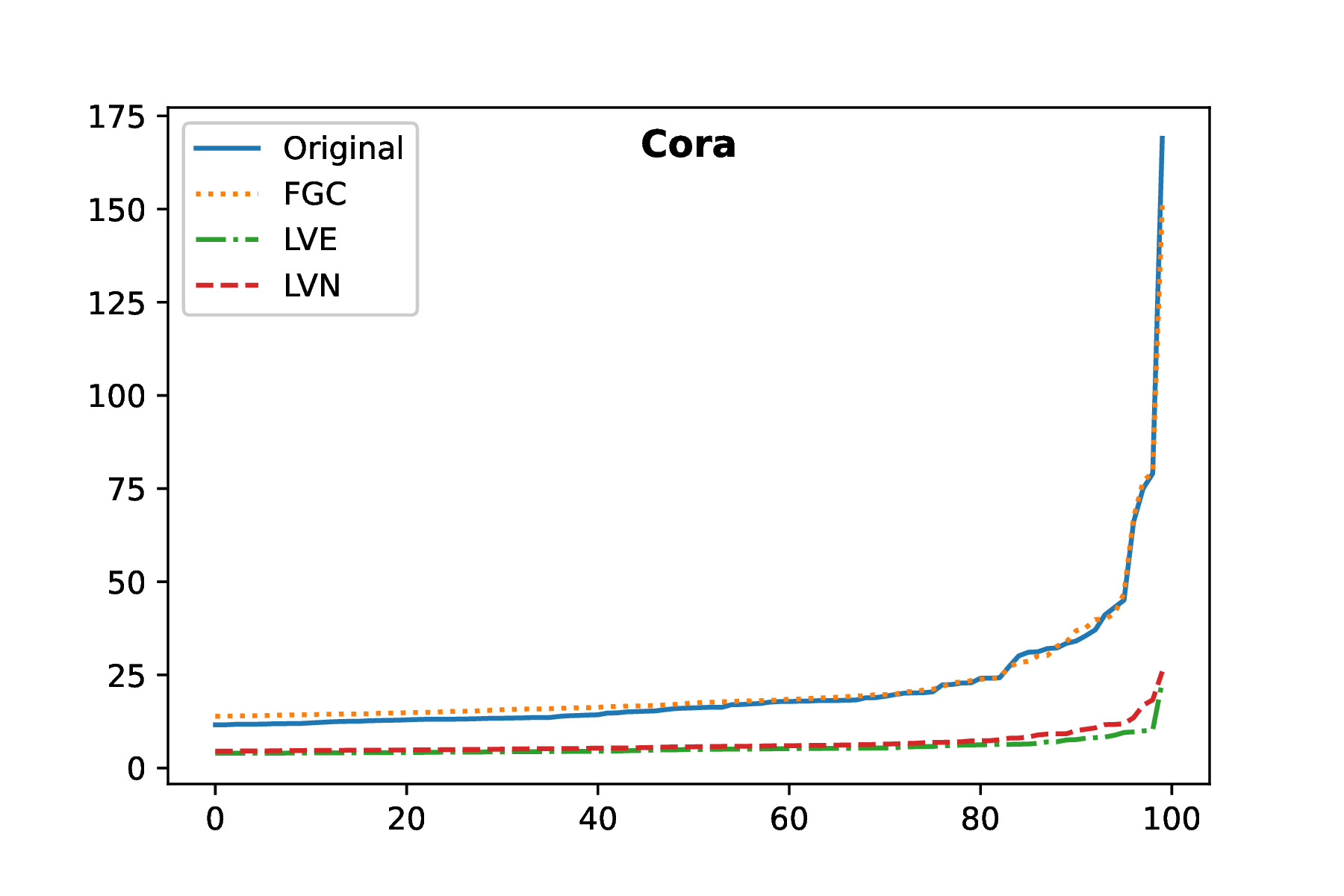}
 \label{FGC eigen values Coraeig}
 \end{subfigure}
 \begin{subfigure}[b]{0.24\textwidth}
 \centering
 \includegraphics[width=\textwidth]{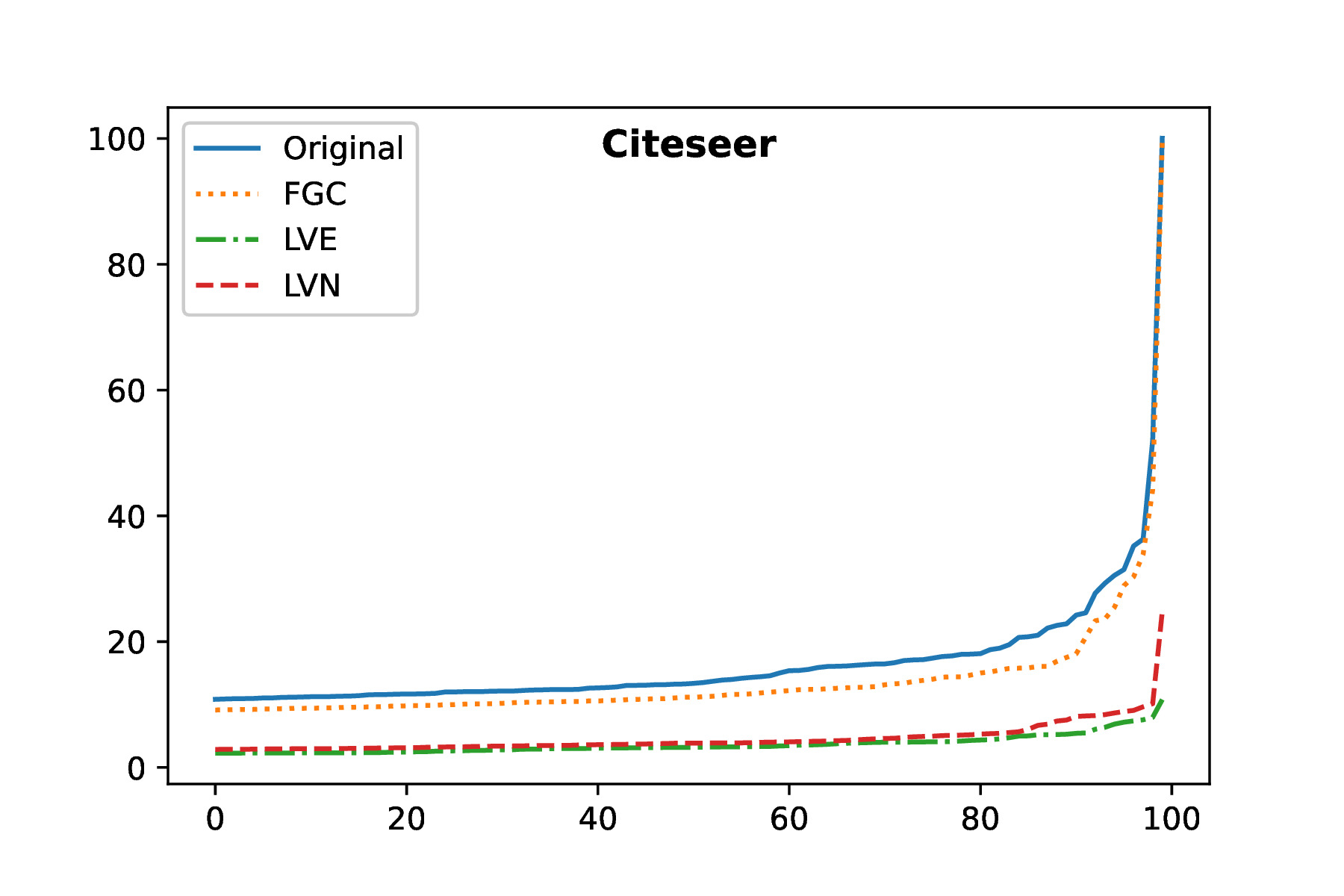}
 \label{FGC eigen values citeseereig}
 \end{subfigure}
 \begin{subfigure}[b]{0.24\textwidth}
 \centering
 \includegraphics[width=\textwidth]{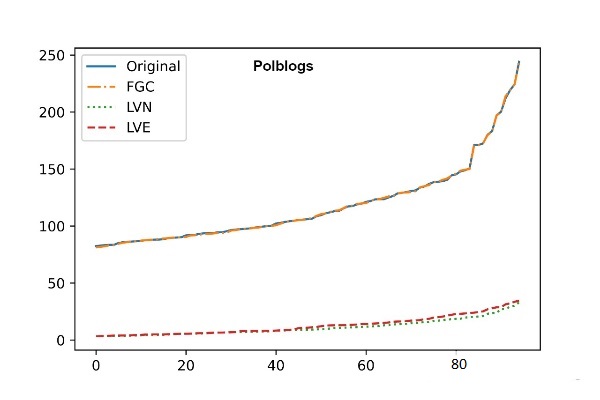}
 \label{FGC eigen values polblogsig}
 \end{subfigure}
 \begin{subfigure}[b]{0.24\textwidth}
 \centering
 \includegraphics[width=\textwidth]{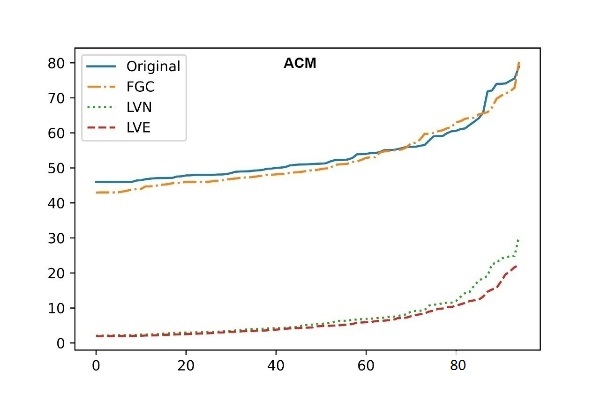}
 \label{FGC eigen values acmeig}
 \end{subfigure}
 \caption{This figure plots the top-100 eigenvalues of the coarsened Laplacian matrix against the original Laplacian matrix for the Cora, Citeseer, Polblogs, and ACM datasets with coarsening ratio r=0.3. The eigenvalues for the coarsened matrix obtained by the FGC algorithm are almost similar to the original graph Laplacian matrix, highlighting that FGC is superior in preserving the spectral properties in the coarsened graph matrix in comparison to the existing state-of-the-art.}
 \label{FGC eigen values}
\end{figure}

\noindent \textbf{Spectral similarity:}
Here we evaluate the FGC algorithm for spectral similarity. The plots are obtained for three coarsening methods FGC (proposed), LVE, and LVN and the coarsening ratio is chosen as $r=0.3$. It is evident in Figure \ref{FGC eigen values} that the top 100 eigenvalues plot of the original graph and coarsened graph learned from the proposed FGC algorithm are similar as compared to other state-of-the-art algorithms.

Moreover, \cite{loukas2019graph} has already shown that the local variation methods outperform other pre-existing graph coarsening methods. So, we have compared FGC (proposed) only with the local variation methods \cite{loukas2019graph} in our experiments.\\

\begin{figure}
 \centering
 \includegraphics[scale=.2]{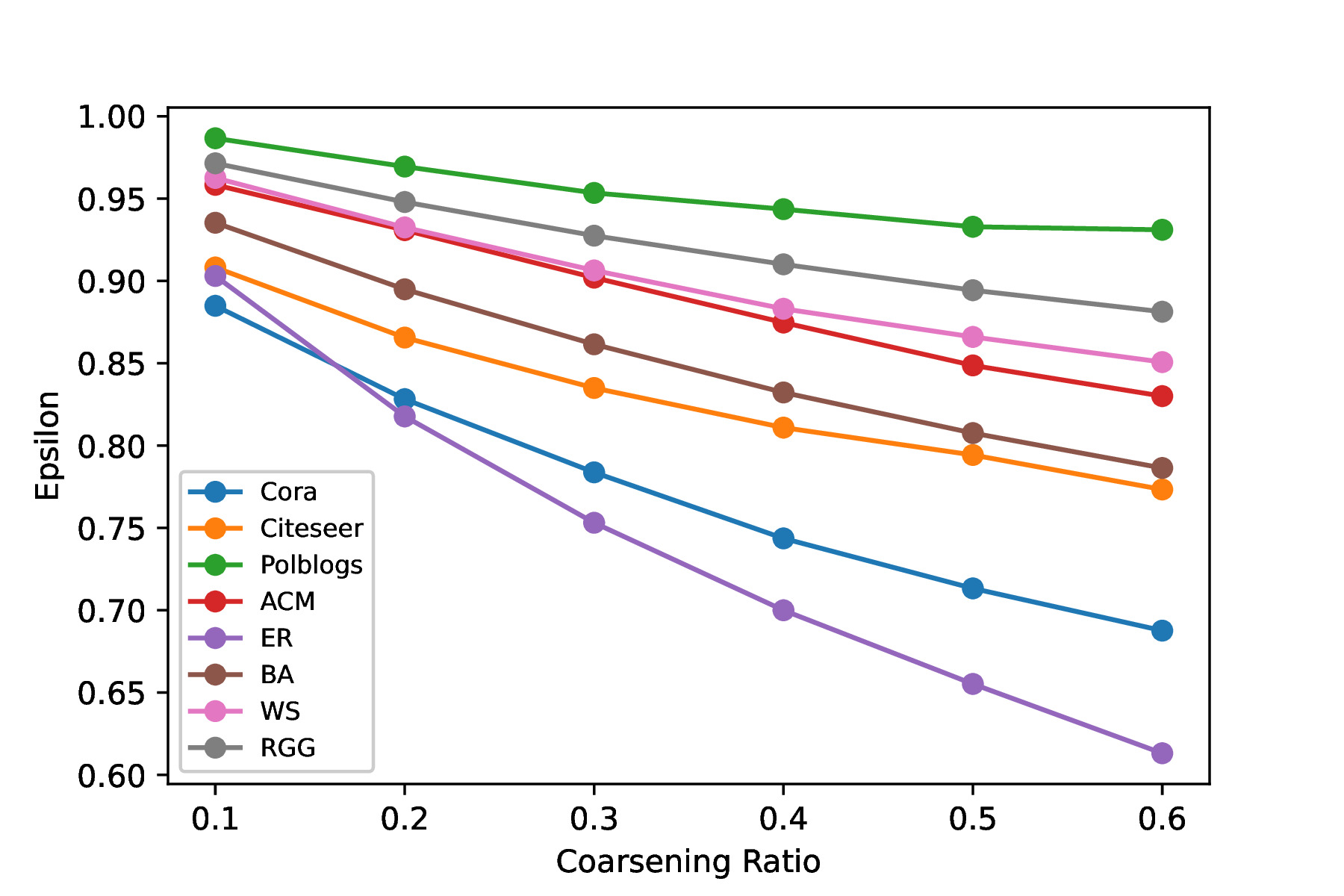}
 \caption{This figure plots the $\epsilon$ values for a variety of real and synthetic datasets. The $\epsilon $ values lying between $(0,1)$ indicate that the coarsened graph $G_c$ learned by the proposed FGC method and $\mathcal{G}$ are similar.} 
 \label{fig: epsilon similarity FGC}
\end{figure}

\noindent \textbf{$\epsilon$-Similarity:}
Here we evaluate the FGC algorithm for $\epsilon-$similarity as discussed in \eqref{epsilon similar}. Note that the similarity definition \eqref{epsilon similar} considers the properties of both the graph matrix and its associated features, while in \cite{loukas2018spectrally} it is restricted to just the graph matrix property. It is evident in Figure \ref{fig: epsilon similarity FGC} that the range of $\epsilon$ is $(0,1)$ which implies that original graph $\mathcal{G}$ and coarsened graph $\mathcal{G}_c$ are $\epsilon$ similar.

\begin{figure}
 \centering
 \begin{subfigure}[b]{0.22\textwidth}
 \centering
 \includegraphics[width=\textwidth]{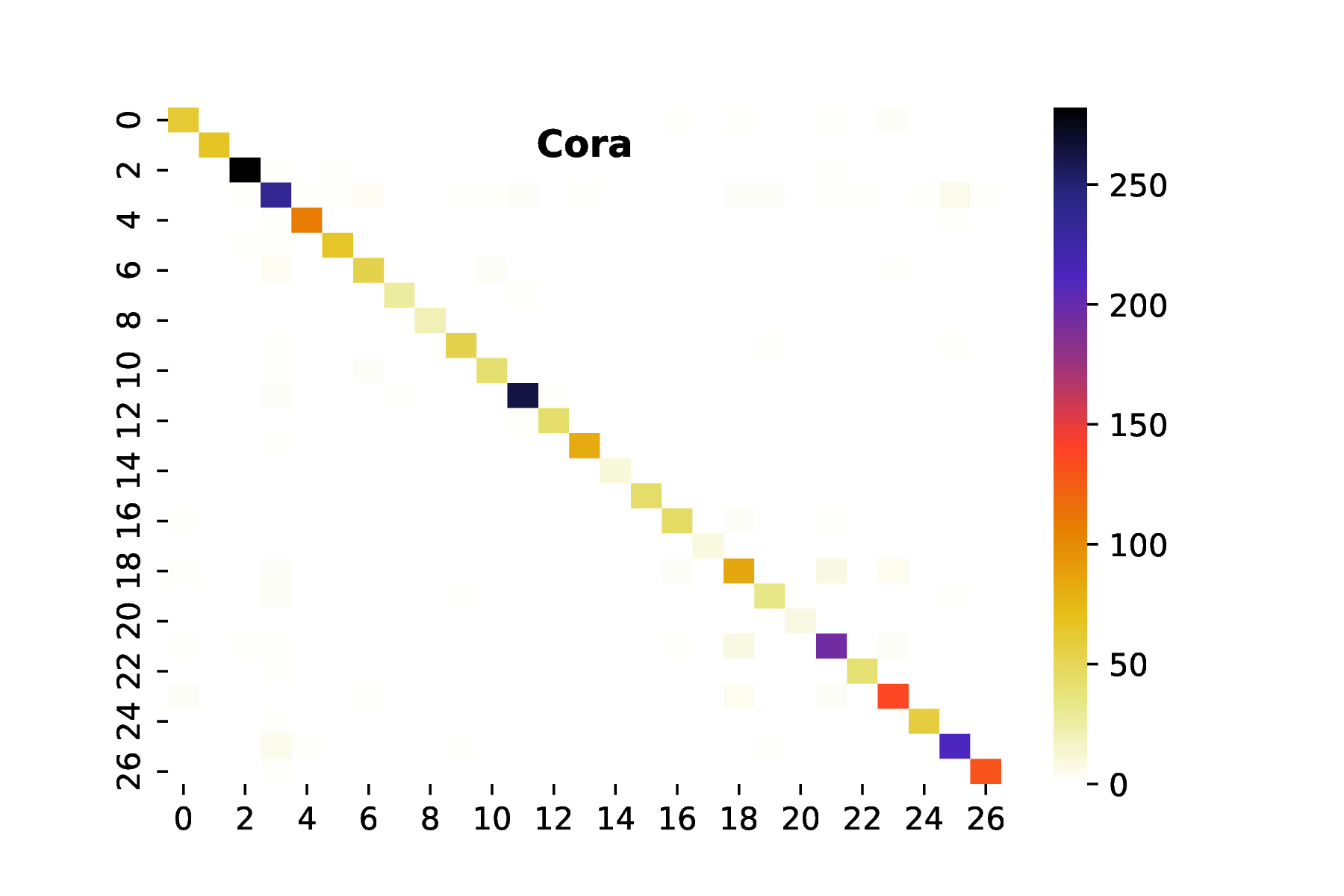}
 \label{FGC Heat Map Cora heatmap}
 \end{subfigure}
 \hfill
 \begin{subfigure}[b]{0.22\textwidth}
 \centering
 \includegraphics[width=\textwidth]{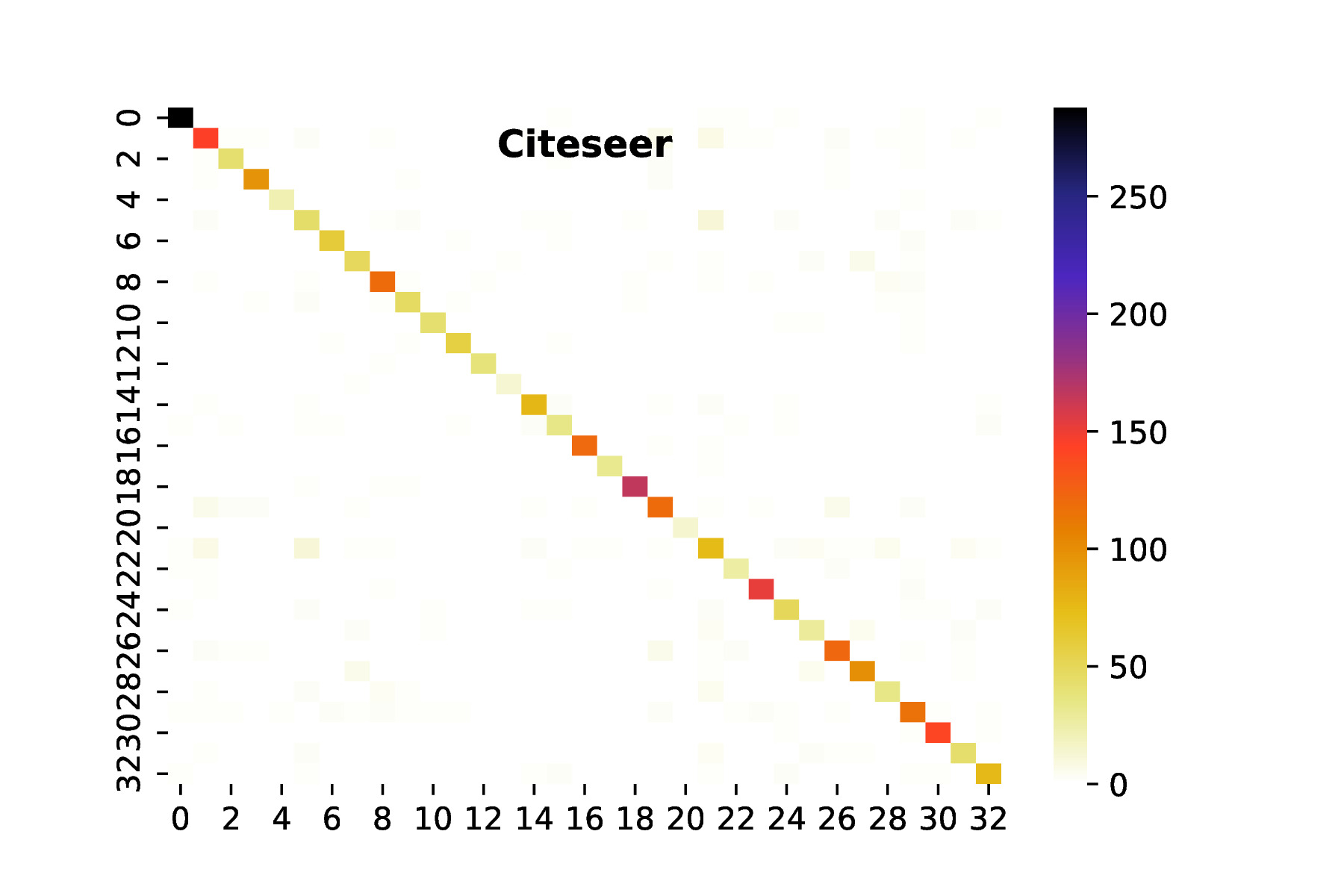}
 \label{FGC Heat Map citeseer heatmap}
 \end{subfigure}
 \hfill\begin{subfigure}[b]{0.22\textwidth}
 \centering
 \includegraphics[width=\textwidth]{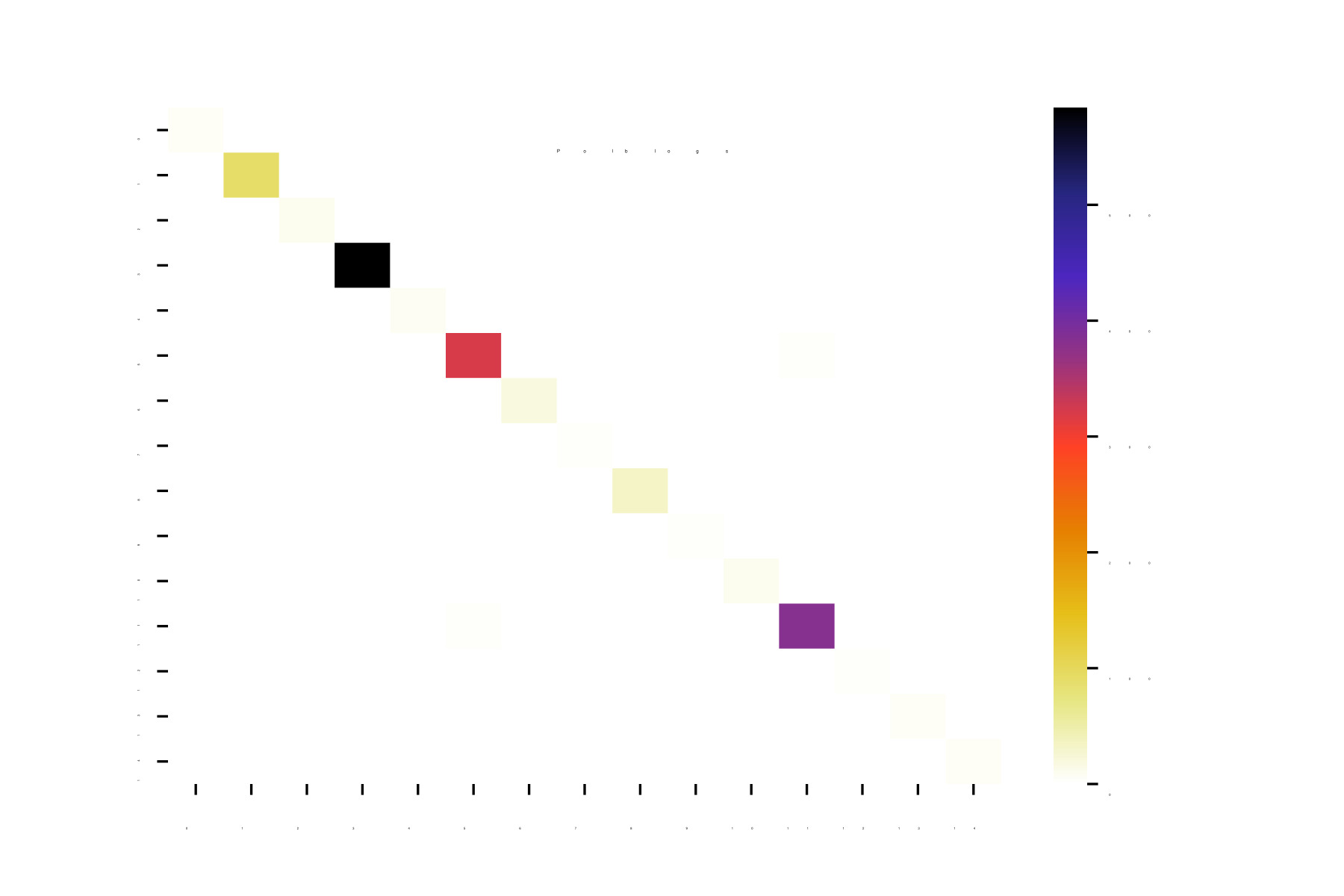}
 \label{FGC Heat Map polblogs heatmap}
 \end{subfigure}
 \hfill\begin{subfigure}[b]{0.22\textwidth}
 \centering
 \includegraphics[width=\textwidth]{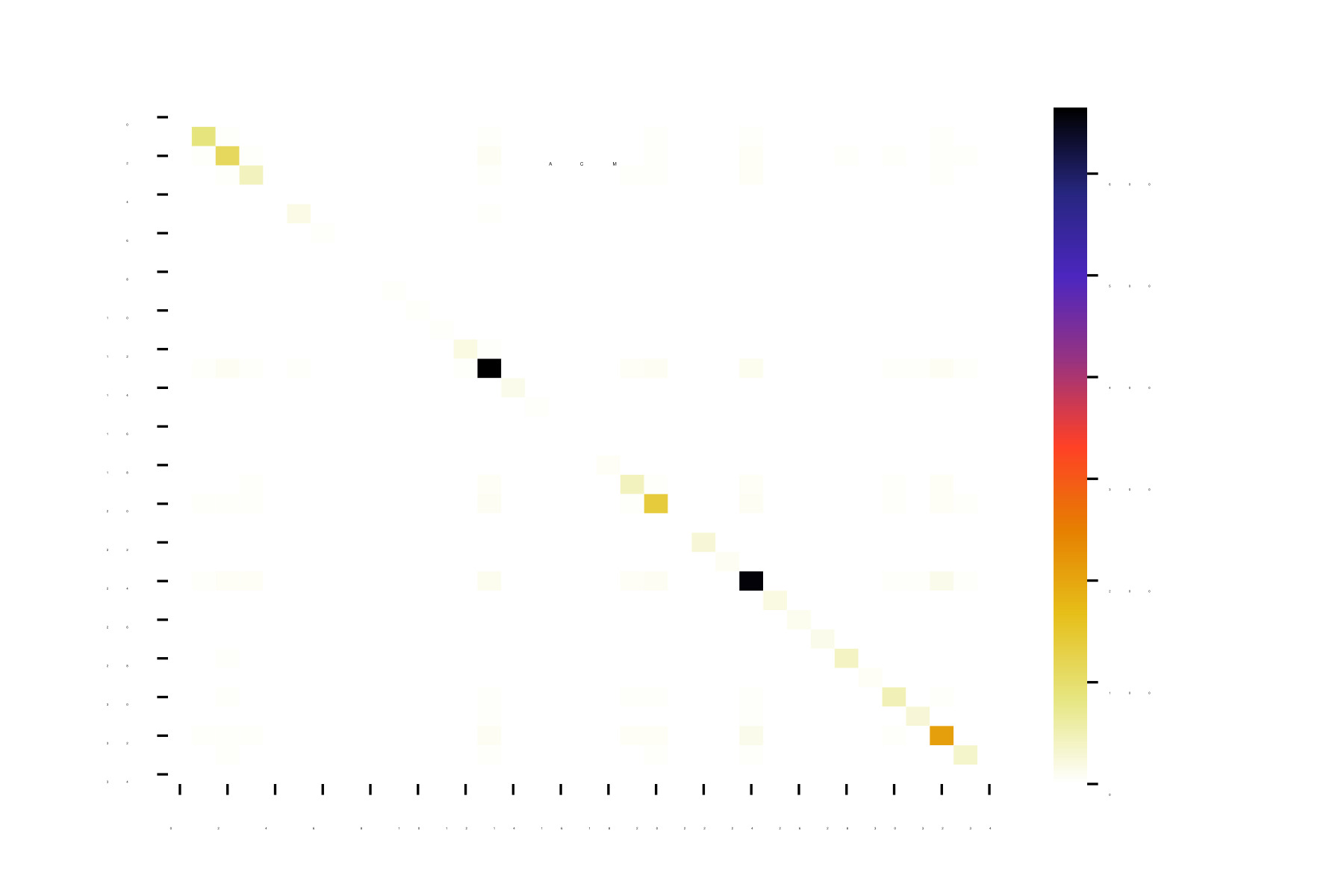}
 \label{FGC Heat Map acm heatmap}
 \end{subfigure}
 \hfill
 \caption{ This figure plots the heat maps of the loading matrices $C^TC$ obtained by FGC algorithm for Cora, Citeseer, Polblogs and ACM datasets for the coarsening ratio $r=0.01$. Even for the extreme coarsening, where the size of the graph is reduced by 100, the $C^TC$ is almost diagonal, which indicates that the $C$ matrix is also almost orthogonal. For moderate coarsening ($r=0.3,0.5)$ we are observing $C$ to be perfectly orthogonal. The strength of the values of the diagonal entries signifies the number of nodes from the set $V$ mapped to a super-node. As indicated from the vertical color bar the mapping is balanced, such that not all or the majority of nodes are mapped to one single supernode in the coarsened graph. These observations also validate that the $\|C^T\|_{1,2}^2$ norm penalty is effective in enforcing desired grouped sparsity structure. Finally, the good results for the experiments with extreme coarsening also suggest that the proposed method can be utilized for doing clustering and stochastic block model identification.}
\label{FGC Heat Map}
\end{figure}

\noindent \textbf{Heat Maps of $C^TC$:}
Here we aim to show the grouping and structural properties ensured by the FGC algorithm. We aim to evaluate the properties of loading $C$ as discussed in \eqref{Loading matrix-set1} which is important for ensuring that the mapping of nodes to super-node should be balanced. It is evident in Figure \ref{FGC Heat Map} that the loading matrix $C$ learned from the proposed FGC algorithm satisfies all the properties of set $\mathcal{S}_c$ in \eqref{Loading matrix-set1}.

\begin{figure}
\centering
\begin{subfigure}[b]{0.24\textwidth}
\centering
\includegraphics[width=\textwidth]{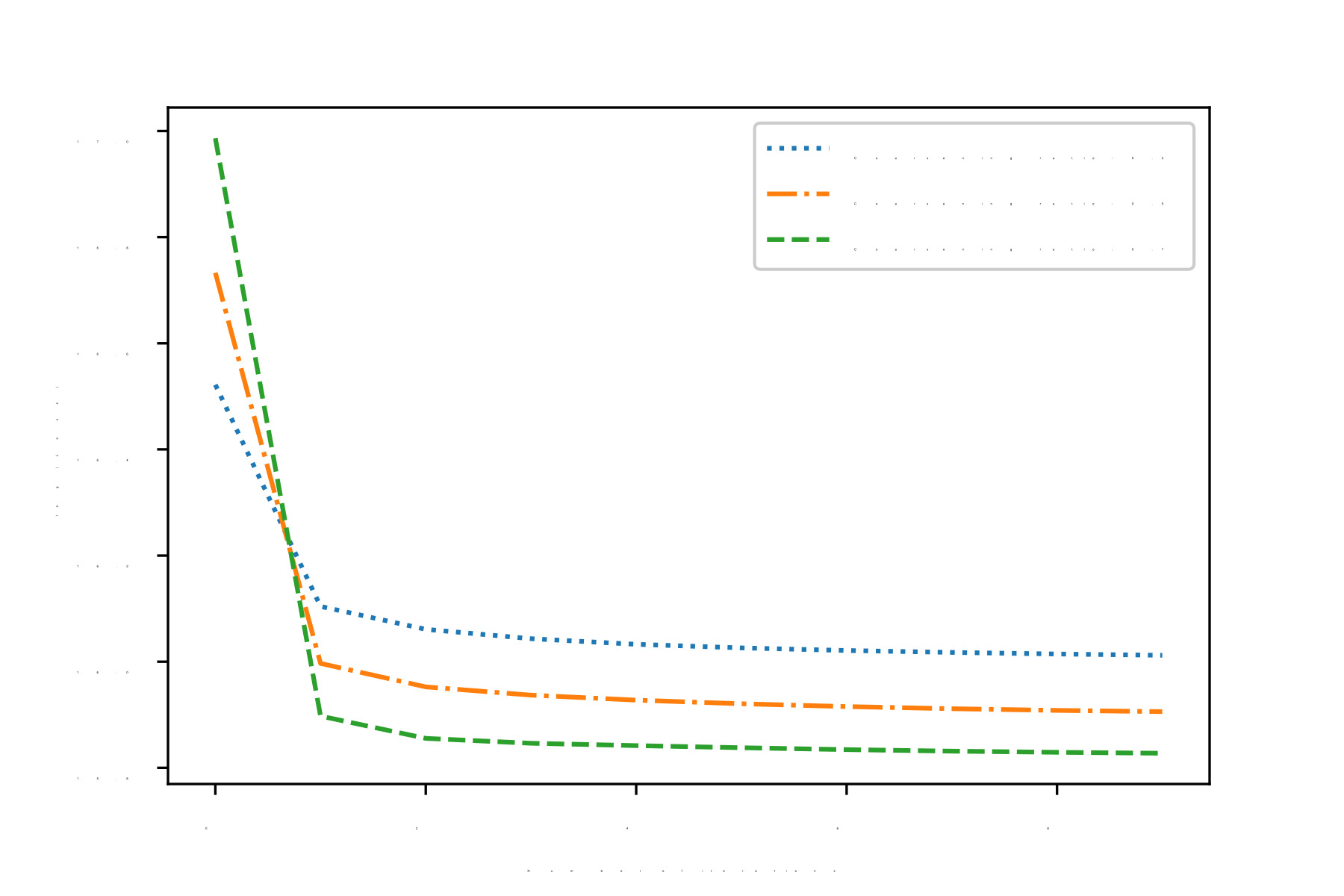}
\label{FGC Loss Curve real cora loss}
\caption{Cora}
\end{subfigure}
\begin{subfigure}[b]{0.24\textwidth}
\centering
\includegraphics[width=\textwidth]{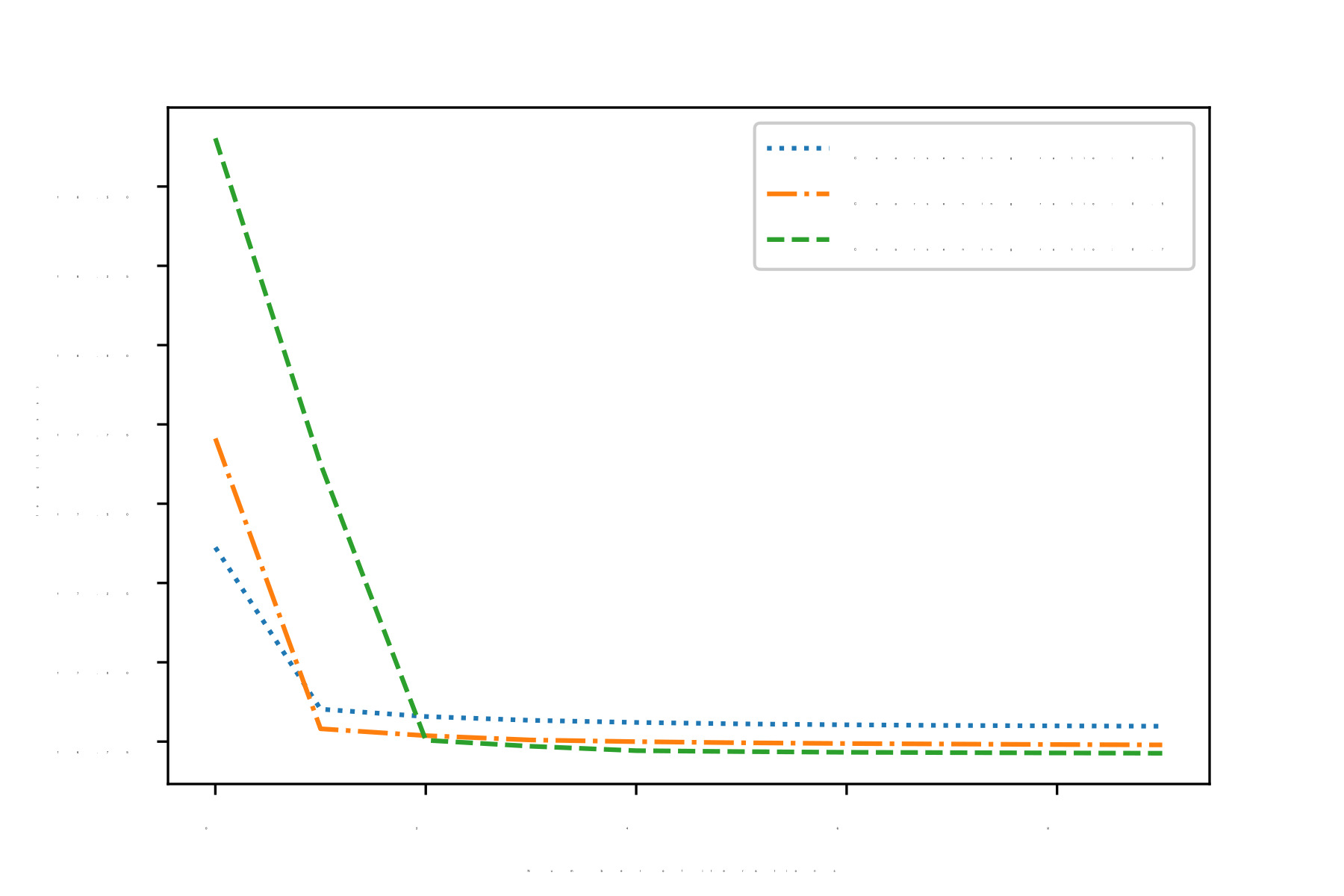}
\label{FGC Loss Curve real citeseer loss}
\caption{Citeseer}
\end{subfigure}
\begin{subfigure}[b]{0.24\textwidth}
\centering
\includegraphics[width=\textwidth]{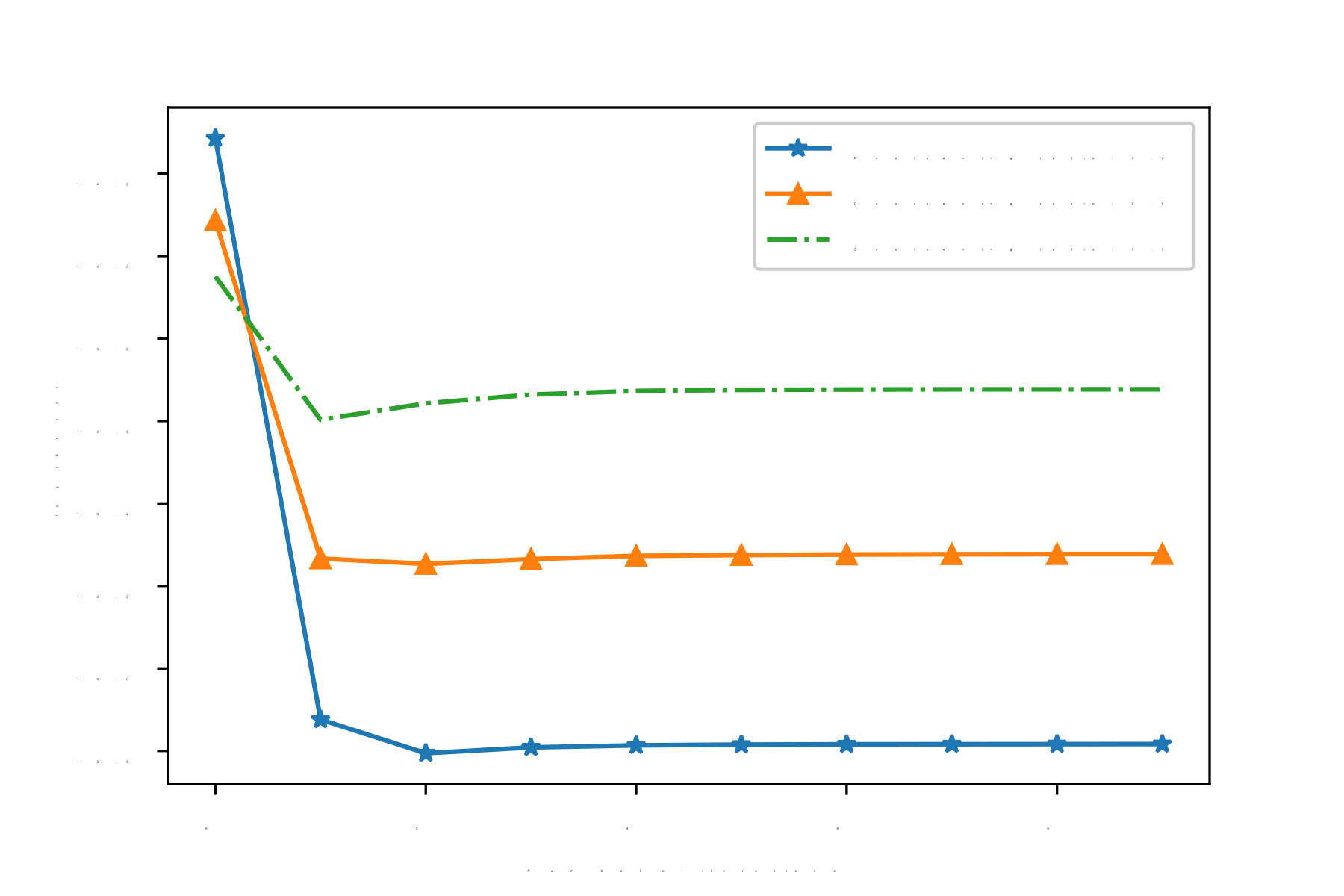}
\label{FGC Loss Curve real polblogs loss}
\caption{Polblogs}
\end{subfigure}
\begin{subfigure}[b]{0.24\textwidth}
\centering
\includegraphics[width=\textwidth]{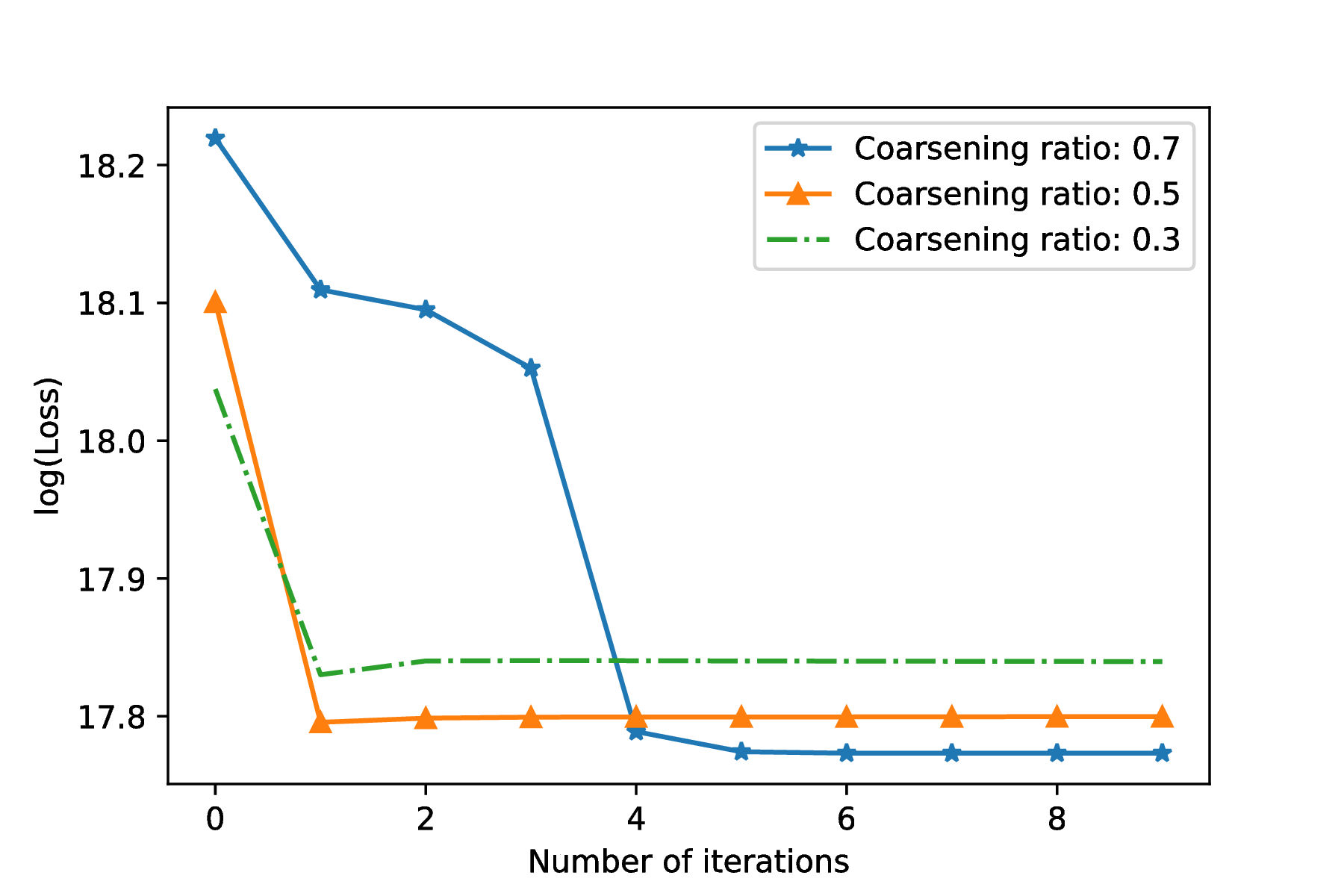}
\label{FGC Loss Curve real acm loss}
\caption{ACM}
\end{subfigure}
\caption{This figure shows the loss curves of FGC for (a) Cora, (b) Citeseer, (c) Polblogs and (d) ACM on different coarsening ratios.}
\label{FGC Loss Curve real}
\end{figure}

\noindent \textbf{Loss Curves:}
Here we plot the loss curves for proposed FGC on 10 iterations for different coarsening ratios $r=$ 0.3, 0.5, and 0.7 respectively where in each iteration, $C$ is updated 100 times having a learning rate $\frac{1}{k}$ on real datasets. The plots in Figure \ref{FGC Loss Curve real} show the convergence properties of the FGC algorithm. 

\subsection{Performance Evaluation for the GC Algorithm}
\noindent \textbf{REE, HE and RE Analysis:} We present REE, HE, and RE on Bunny, Minnesota, and Airfoil datasets validating that GC (proposed) performs better than the state-of-the-art methods, i.e., local variation methods. The aforementioned are popularly used graph datasets that only consist of graph matrices. The HE values for graph data having no feature matrix can be obtained using \cite{bravo2019unifying},
\begin{equation}
\textbf{HE}=\text{arccosh}\Bigg(1+\frac{\|(\Theta-\Theta_{\text{lift}})x\|_F^2\|x\|_F^2}{2(x^T\Theta x)(x^T\Theta_{\text{lift}}x)}\Bigg).
\end{equation}
 where $x$ is the eigenvector corresponding to the smallest non-zero eigenvalue of the original Laplacian matrix. It is evident from Table \ref{tabel: GC on real datasets} and \ref{tabel: GC on syntetic REE, HE and RE.} that the proposed GC algorithms have the lowest value of $HE$ and $DE$ on both real and synthetic datasets as compared to another state of the art algorithms which indicate that $\Theta_{\text{lift}}$ learned from coarsened graph matrix $\Theta_c$ is closer to original graph matrix $\Theta$.
\begin{table}
\begin{center}
{\renewcommand{\arraystretch}{1.7}
\begin{tabular}{ |m{1.6cm} | m{1.7em} | m{2.4em}| m{2.4em} |m{2.4em} | m{2.4em}|m{2.4em}|m{2.4em} | m{2.1em}|m{2.1em}|m{2.1em}|} 
 \hline
 \multirow{2}{*}{Dataset} & \multirow{2}{*}{$r$=$\frac{k}{p} $} & 
 \multicolumn{3}{c|}{$\text{REE}$} &
 \multicolumn{3}{c|}{$\text{HE}$} & \multicolumn{3}{c|}{RE in $\log(\cdot)$}\\
 \cline{3-11}
 & & FGC & LVN & LVE & FGC & LVN & LVE & FGC & LVN & LVE\\ 
 \hline
 Minnesota & 0.7 0.5 0.3 & \textbf{0.013} \textbf{0.015 } \textbf{0.026} & 0.058 0.164 0.473 & 0.093 0.283 0.510 & 0.847 \textbf{ 1.300 1.808} & \textbf{0.707} 1.530 2.407 & 1.120 1.770 2.390 & \textbf{1.16 1.64 1.95} & 1.37 1.88 2.14 & 1.51 1.97 2.15 \\ 
 \hline
 Airfoil & 0.7 0.5 0.3 & \textbf{0.013 } \textbf{0.014 } \textbf{0.032 } & 0.027 0.092 0.192 & 0.019 0.092 0.186 & \textbf{0.742 1.073 1.520} & 0.787 1.220 1.930 & 0.799 1.266 1.908 & \textbf{1.17 1.66 1.98} & 2.65 3.15 3.50 & 2.68 3.22 3.51 \\ 
 \hline
 Bunny & 0.7 0.5 0.3 & \textbf{0.024 } \textbf{0.015 } \textbf{0.020} & 0.092 0.200 0.286 & 0.046 0.050 0.128 & \textbf{0.703 0.923 1.482} & 0.816 1.240 1.673 & 0.703 1.045 1.542 & \textbf{7.14 7.65 7.99} & 7.36 7.85 8.11 & 7.27 7.65 8.01\\ 
 \hline
\end{tabular}
}
\end{center}
 \caption{This table summarizes the REE, HE and RE values obtained by GC (proposed), LVN, and LVE on different coarsening ratios ($r$) for standard real graph datasets. The proposed GC algorithm outperforms the state-of-the-art methods significantly.}
\label{tabel: GC on real datasets}
\end{table}
\begin{table}
\begin{center}
\begin{tabular}{ |m{1.3cm} | m{2.0em} | m{2.3em}| m{2.4em} |m{2.4em} | m{2.5em}|m{2.5em}|m{2.1em}| m{2.1em}|m{2.1em}|m{2.1em}|} 
 \hline
 \multirow{2}{*}{Dataset} & \multirow{2}{*}{ $r$=$\frac{k}{p} $} &
 \multicolumn{3}{c|}{$\text{REE}$} &
 \multicolumn{3}{c|}{$\text{HE}$} & \multicolumn{3}{c|}{RE in $\log(\cdot)$}\\
 \cline{3-11}
 & &GC &LVN &LVE & GC &LVN &LVE & GC &LVN &LVE\\ 
 \hline
 BA & 0.7 0.5 0.3 & \textbf{0.038 0.055 0.068} & 0.164 0.445 0.644 & 0.290 0.427 0.504 &
 \textbf{0.73 1.03 1.44} & 0.81 1.17 1.67 & 0.95 1.09 1.60 & \textbf{6.50 7.13 7.50} & 6.65 7.40 7.67 & 7.07 7.39 7.53\\ 
 \hline
 WS & 0.7 0.5 0.3 & 0.026 \textbf{ 0.053 0.091} & 0.038 0.070 0.120 & \textbf{0.025} 0.063 0.107 & \textbf{0.62 1.03 1.48 } & 0.75 1.13 1.59 & 0.68 1.03 1.52 & \textbf{4.87 5.37 5.70} & 4.92 5.43 5.76 & 4.92 5.42 5.75\\ 
 \hline
 ER & 0.7 0.5 0.3 & 0.053 0.078 \textbf{0.101} & \textbf{0.035 0.059} 0.109 & 0.055 0.066 0.113 & 0.71 1.05 \textbf{ 1.40} & 0.77 1.06 1.49 & \textbf{0.65 0.97} 1.45 & \textbf{8.02 8.53 8.86} & 8.07 8.56 8.89 & 8.15 8.55 8.89\\ 
 \hline
RGG & 0.7 0.5 0.3 & \textbf{0.024 0.050 0.086} & 0.069 0.160 0.234 & 0.033 0.052 0.127 & \textbf{0.58 0.93 1.50} & 0.68 1.20 2.07 & 0.76 1.02 1.57 & \textbf{5.64 6.16 6.50} & 5.88 6.38 6.63 & 5.82 6.20 6.54\\ 
 \hline
\end{tabular}
\end{center}
 \caption{REE, HE, and RE in $log(\cdot)$ values on Barabasi Albert (BA), Watts Strogatz (WS), Erdos Renyi (ER), and Random Geometric Graph (RGG) datasets. The GC (proposed) algorithm shows superior performance over competing benchmarks.}
\label{tabel: GC on syntetic REE, HE and RE.}
\end{table}

\begin{figure}
 \centering
 \begin{subfigure}[b]{0.32\textwidth}
 \centering
 \includegraphics[width=\textwidth]{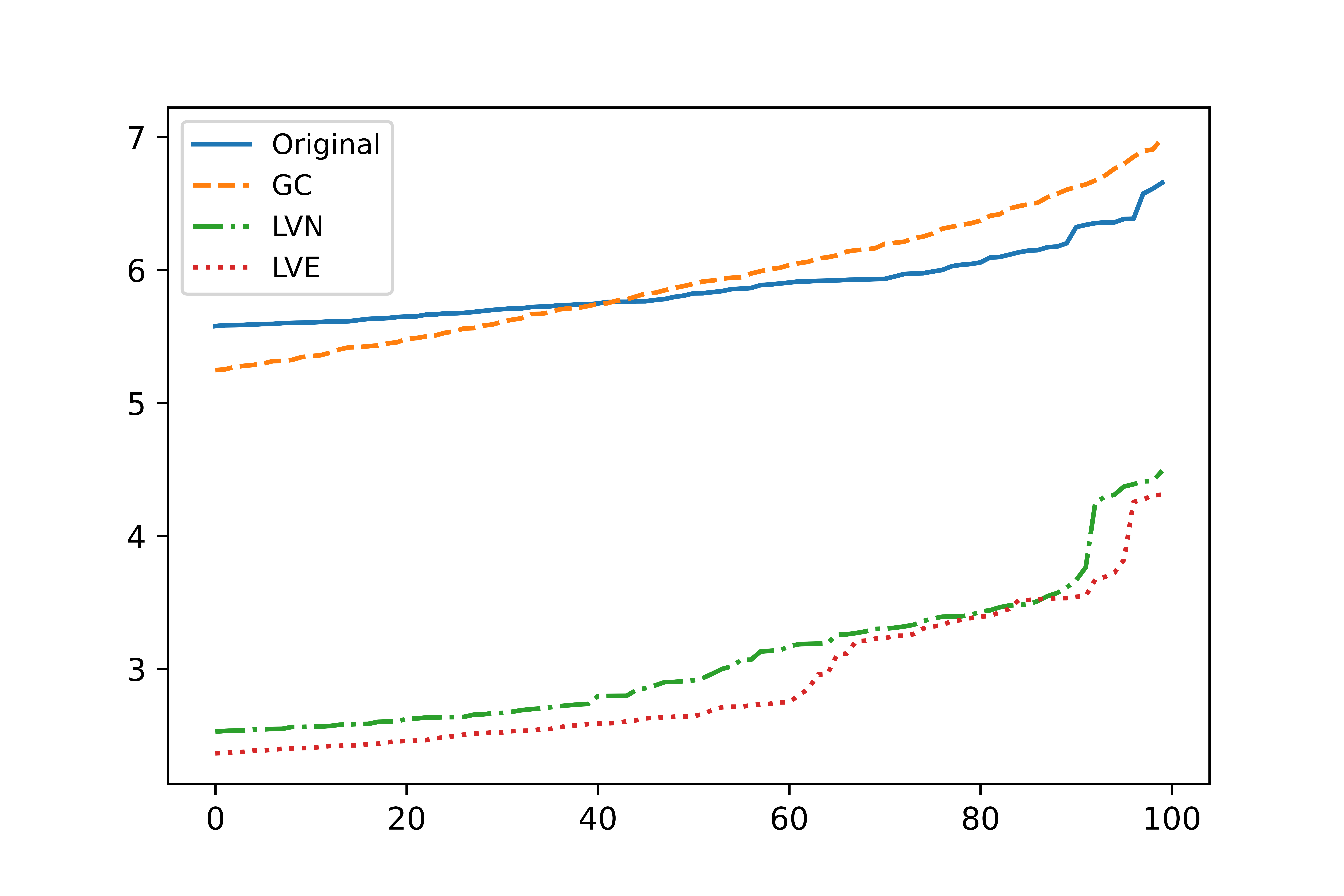}
 \label{GC eigen values Minnesota_Eigen_3}
 \caption{Minnesota($r= 0.3$)}
 \end{subfigure}
 \begin{subfigure}[b]{0.32\textwidth}
 \centering
 \includegraphics[width=\textwidth]{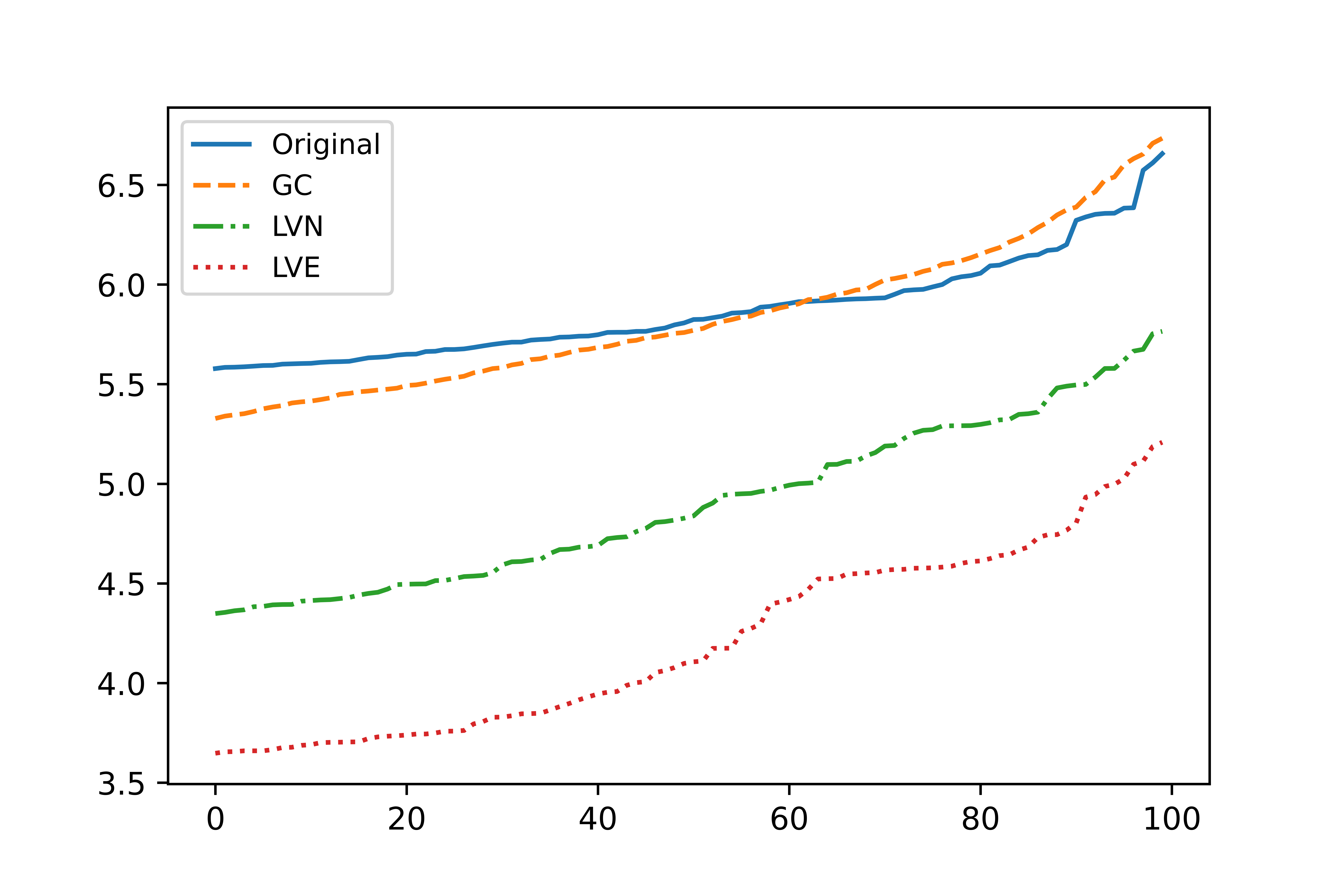}
 \label{GC eigen values Minnesota_Eigen_5}
 \caption{Minnesota($r$= 0.5)}
 \end{subfigure}
 \begin{subfigure}[b]{0.32\textwidth}
 \centering
 \includegraphics[width=\textwidth]{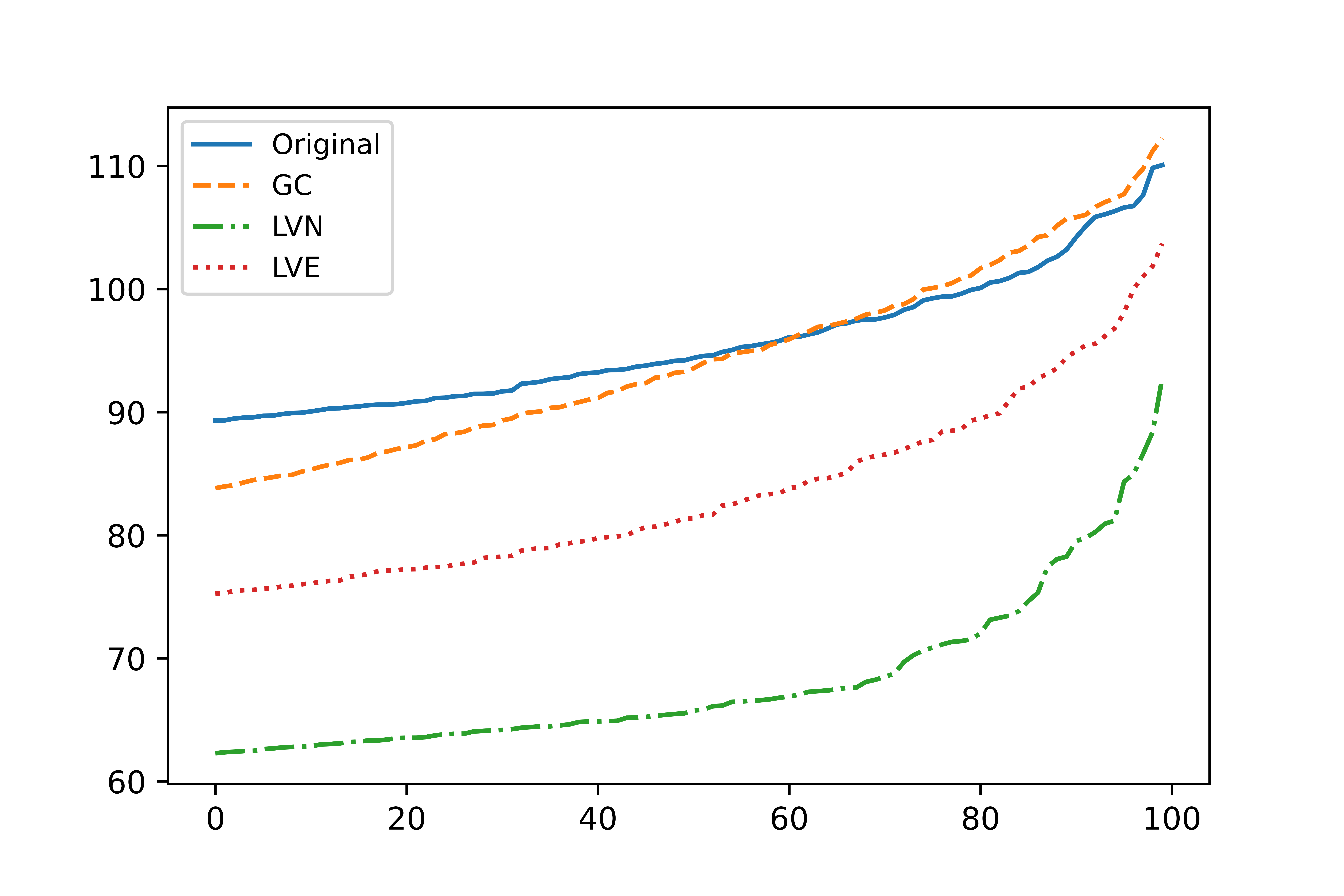}
 \label{GC eigen values Bunny_Eigen_3}
 \caption{Bunny(r= 0.3)}
 \end{subfigure}
 \begin{subfigure}[b]{0.32\textwidth}
 \centering
 \includegraphics[width=\textwidth]{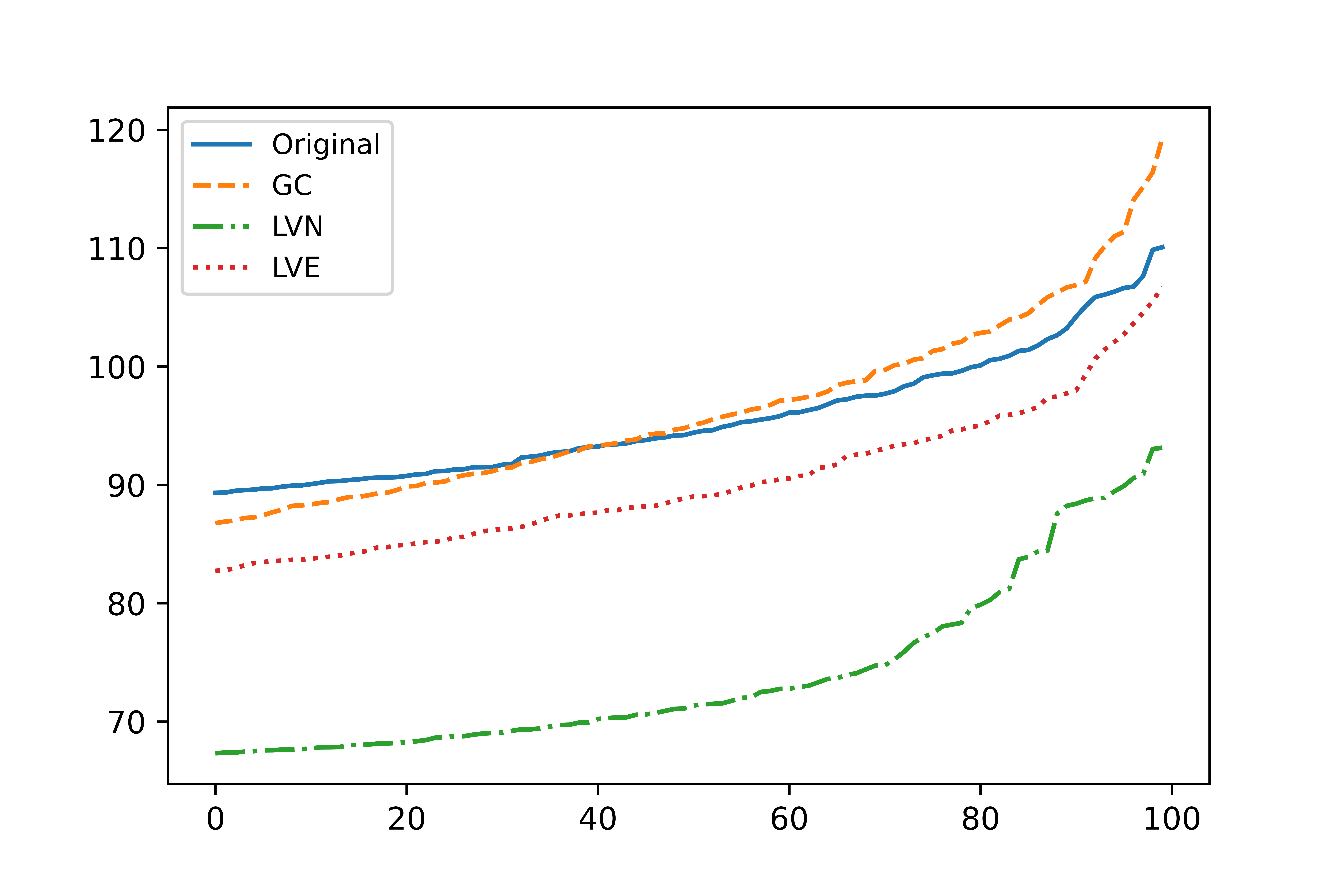}
 \label{GC eigen values Bunny_Eigen_5}
 \caption{Bunny($r$= 0.5)}
 \end{subfigure}
 \begin{subfigure}[b]{0.32\textwidth}
 \centering
 \includegraphics[width=\textwidth]{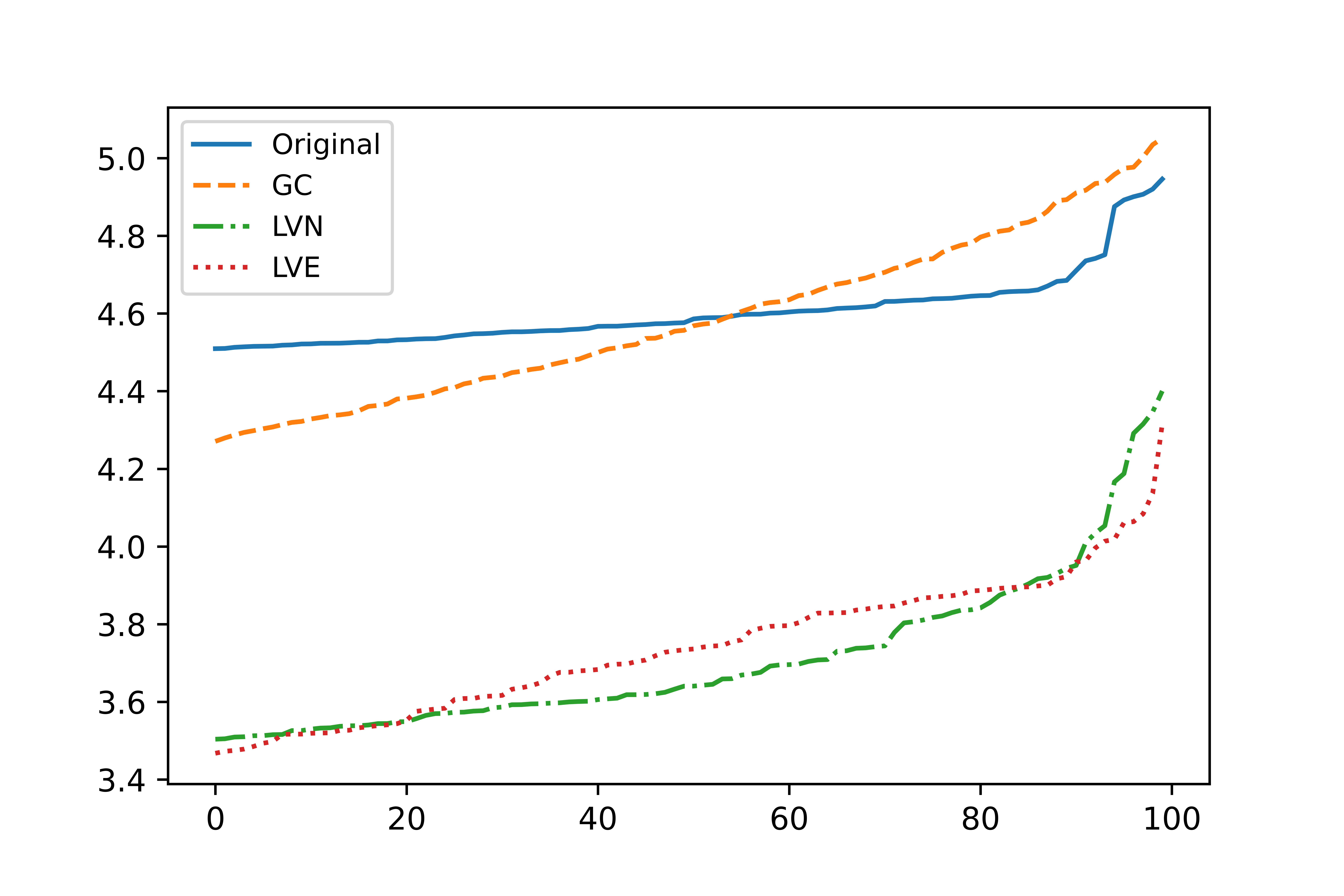}
 \label{GC eigen values Airfoil_Eigen_3}
 \caption{Airfoil($r$= 0.3)}
 \end{subfigure}
 \begin{subfigure}[b]{0.32\textwidth}
 \centering
 \includegraphics[width=\textwidth]{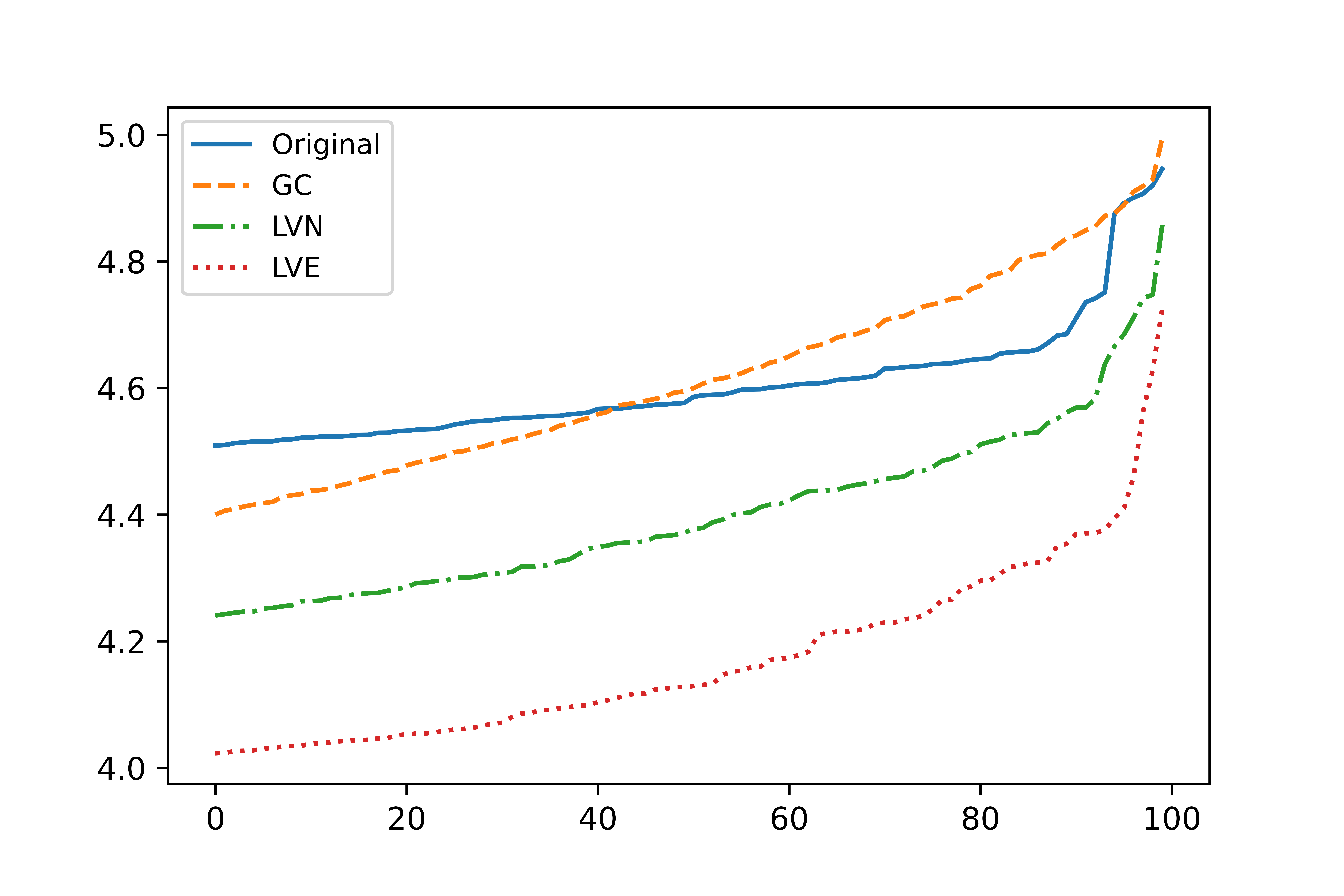}
 \label{GC eigen values Airfoil_Eigen_5}
 \caption{Airfoil($r$= 0.5)}
 \end{subfigure}
 \caption{This figure shows the top-100 eigen value plots for original graph and coarsened graphs obtained by GC (proposed), LVN and LVE for real datasets.}
 \label{GC eigen values}
\end{figure}

\begin{figure}
 \centering
 \begin{subfigure}[b]{0.32\textwidth}
 \centering
 \includegraphics[width=\textwidth]{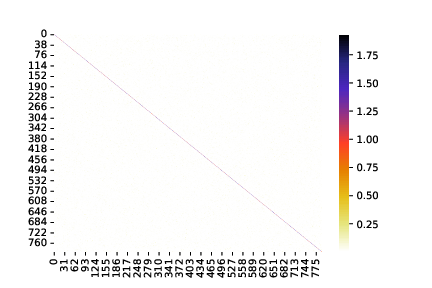}
 \label{GC for heat maps Minnesota}
 \caption{Minnesota}
 \end{subfigure}
 \begin{subfigure}[b]{0.32\textwidth}
 \centering
 \includegraphics[width=\textwidth]{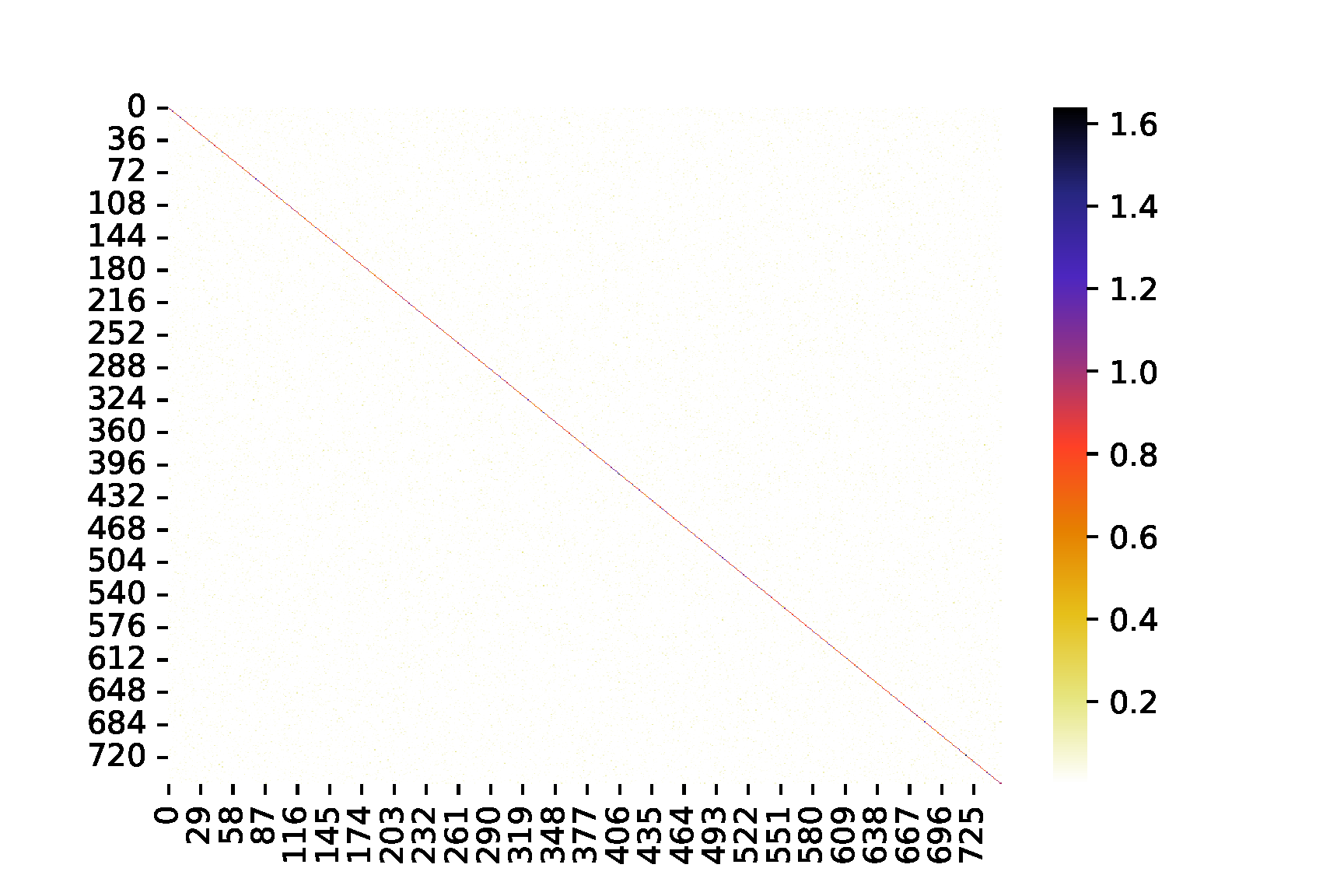}
 \label{GC for heat maps bunny}
 \caption{Bunny}
 \end{subfigure}
 \begin{subfigure}[b]{0.32\textwidth}
 \centering
 \includegraphics[width=\textwidth]{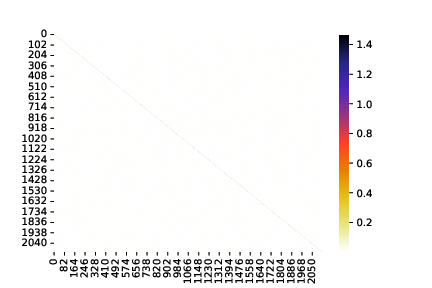}
 \label{GC for heat maps airfoil}
 \caption{Airfoil}
 \end{subfigure}
\caption{This figure presents the heat maps of $C^TC$ for real datasets which do not have the feature matrix. It is evident from Figure \ref{GC for heat map1} that the loading matrix $C$ obtained by the proposed GC algorithm results in a balanced mapping}
\label{GC for heat map1}
\end{figure}

\noindent \textbf{Spectral Similarity:}
We compare the spectral similarity of the GC (proposed) framework against LVN, and LVE. We plot the top 100 eigenvalues of original and coarsened graph Laplacian matrices with the following coarsening ratios$r$= 0.3, 0.5, and 0.7. It is evident from Figure \ref{GC eigen values} that eigenvalues of coarsened graph matrix learn from the proposed GC algorithm are almost similar to the original graph Laplacian matrix as compared to other state-of-the-art algorithms.
\noindent \textbf{Heat Maps of $C^TC$:} The heatmap of $C^TC$ is shown in Figure \ref{GC for heat map1}.


\subsubsection{Two Stage Featured Graph Coarsening}
Two-stage featured graph coarsening method is an extension of the GC (proposed) algorithm. In the first step, we obtain $C$ matrix using GC (proposed), and then in the second step, we learn the smooth feature matrix of coarsened graph $X_c$ using \eqref{eqn:X}. The computational complexity of GC algorithm is less as compared to the FGC algorithm. Two stage Featured graph coarsening algorithm can be used for large datasets. The performance comparison of FGC and two stage graph coarsening algorithm is in Table \ref{tabel: GC DE real}.
\begin{table}
\begin{center}
\begin{tabular}{ |m{5em} | m{2em} | m{4em}| m{6em}|} 
 \hline
 \multirow{2}{*}{Dataset} & \multirow{2}{*}{$r$=$\frac{k}{p} $} & 
 \multicolumn{2}{c|}{$\text{HE}$}\\
 \cline{3-4}
 & & FGC & Two stage GC\\
 \hline
 Cora & 0.7 0.5 0.3 & \textbf{1.71} \textbf{\newline} \textbf{1.18} \textbf{\newline} \textbf{0.72} & 1.84\textbf{\newline} 1.40\textbf{\newline} 0.85 \\ 
 \hline
 Citeseer & 0.7 \textbf{\newline}0.5\textbf{\newline} 0.3 & \textbf{1.80} \textbf{\newline} \textbf{1.05} \textbf{\newline}\textbf{0.85} & 2.08 \textbf{\newline}1.09\textbf{\newline} 0.902\\ 
 \hline
 Polblogs & 0.7\textbf{\newline} 0.5 \textbf{\newline}0.3 & 2.89 \textbf{\newline}2.70 \textbf{\newline}\textbf{1.73} & \textbf{2.64} \textbf{\newline} \textbf{2.38}\textbf{\newline} 2.14 \\ 
 \hline
 ACM & 0.7\textbf{\newline} 0.5 \textbf{\newline}0.3 & \textbf{1.86} \textbf{\newline}\textbf{0.98} \textbf{\newline}\textbf{0.45} & 2.21 \textbf{\newline}1.84 \textbf{\newline}1.095 \\ 
 \hline
\end{tabular}
\end{center}
 \caption{This table summarizes the HE value obtained by FGC (proposed) and two stage graph GC (proposed) on different coarsening ratios ($r$) for standard real graph datasets.}
\label{tabel: GC DE real}
\end{table}

\subsection{Performance Evaluation for the FGCR Algorithm}
\begin{figure}
 \centering
 \begin{subfigure}[b]{0.24\textwidth}
 \centering
 \includegraphics[width=\textwidth]{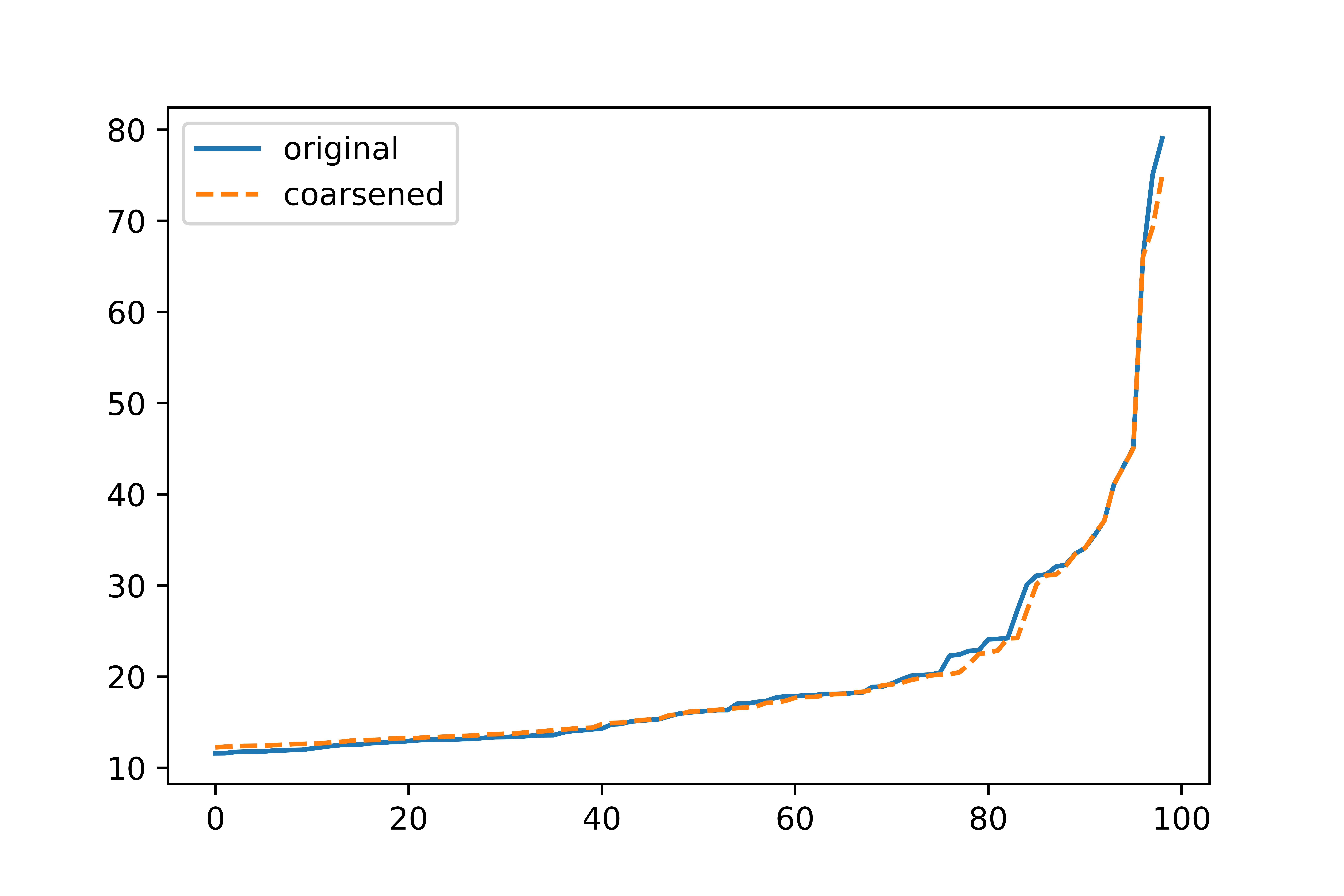}
 \label{FGC eigen values Improvised_cora_Eigen_3_3}
 \caption{Cora}
 \end{subfigure}
 \begin{subfigure}[b]{0.24\textwidth}
 \centering
 \includegraphics[width=\textwidth]{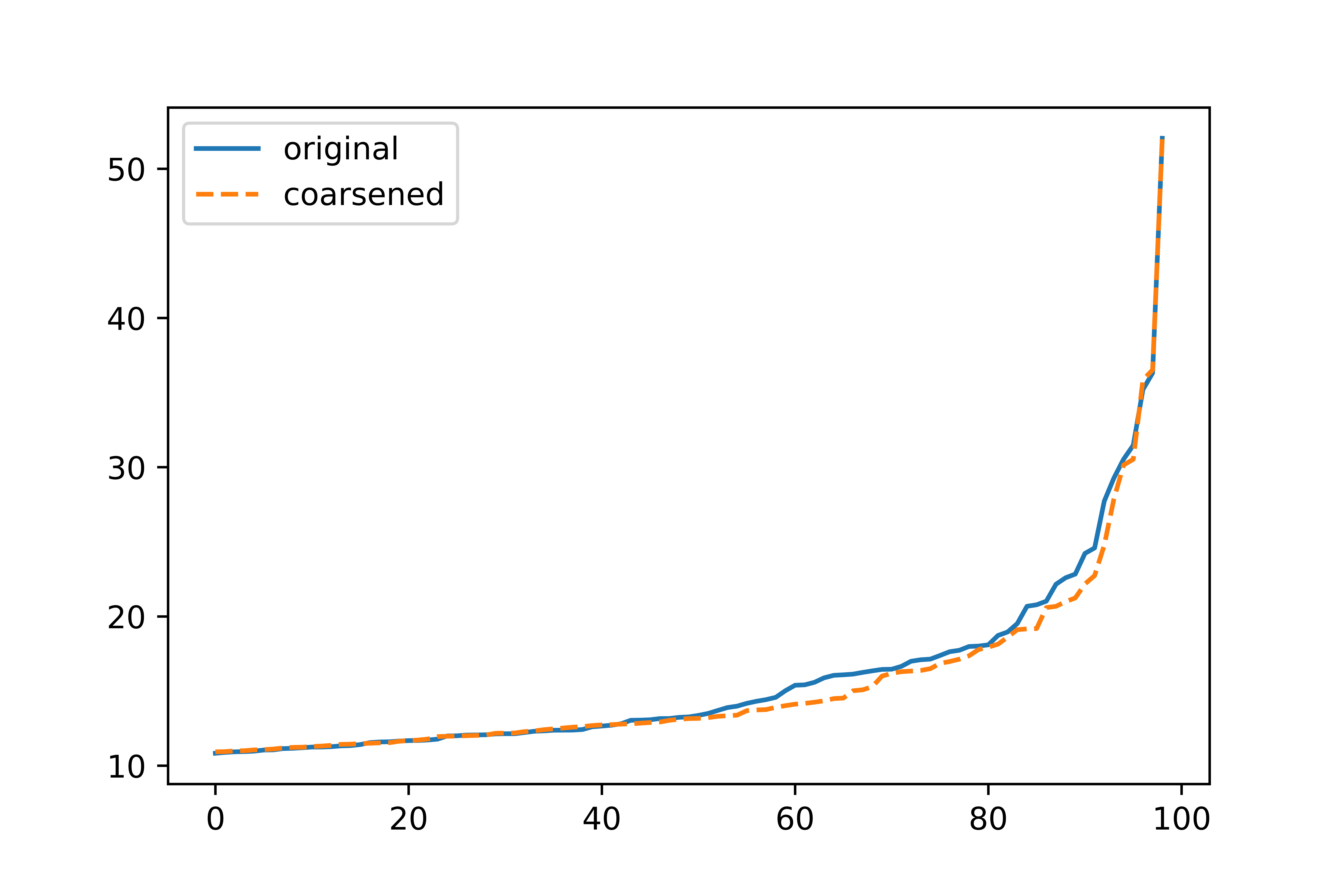}
 \label{FGC eigen values Improvised_citeseer_Eigen_3_7}
 \caption{Citeseer}
 \end{subfigure}
 \begin{subfigure}[b]{0.24\textwidth}
 \centering
 \includegraphics[width=\textwidth]{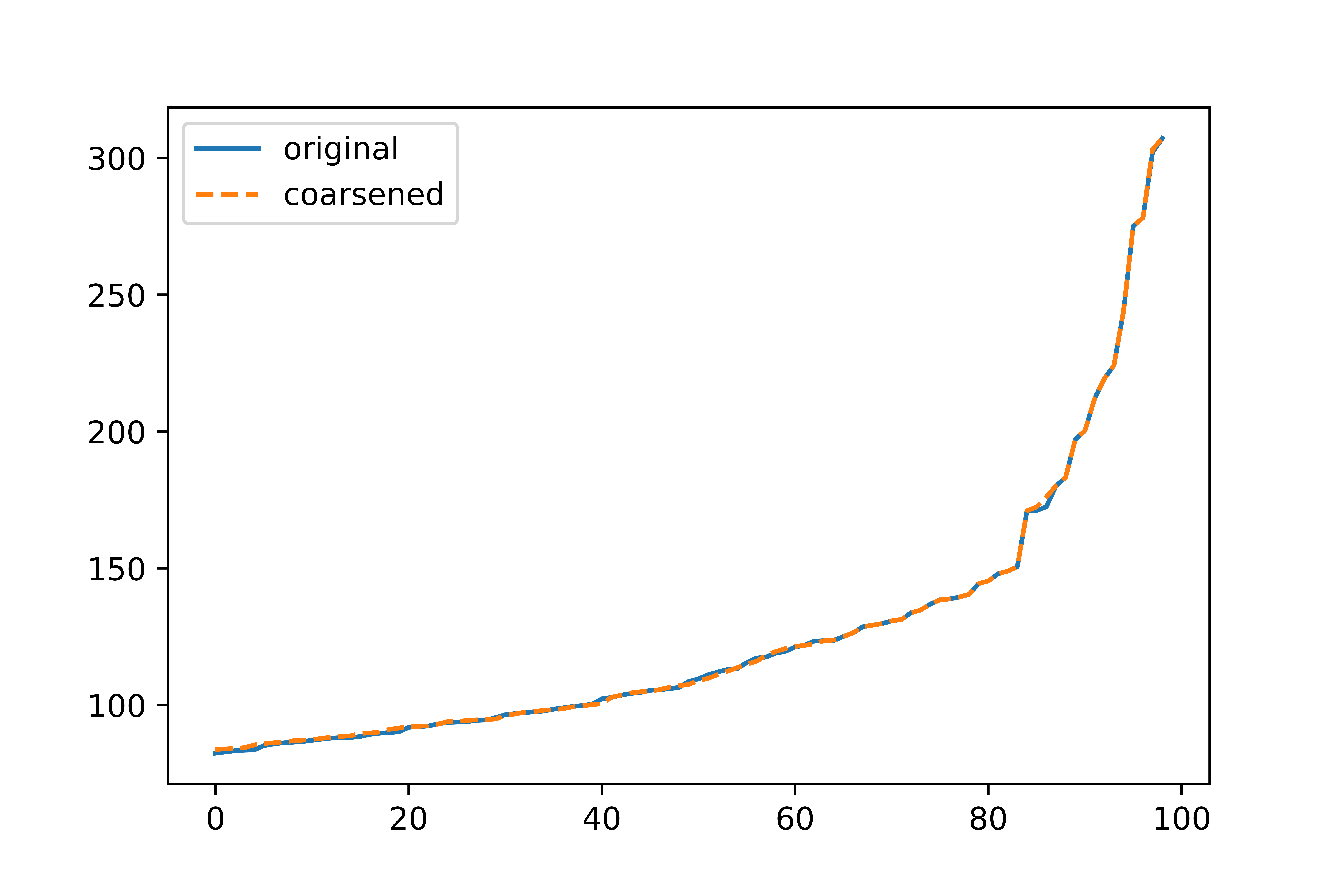}
 \label{FGC eigen values Improvised_polblogs_Eigen_3_3}
 \caption{Polblogs}
 \end{subfigure}
 \begin{subfigure}[b]{0.24\textwidth}
 \centering
 \includegraphics[width=\textwidth]{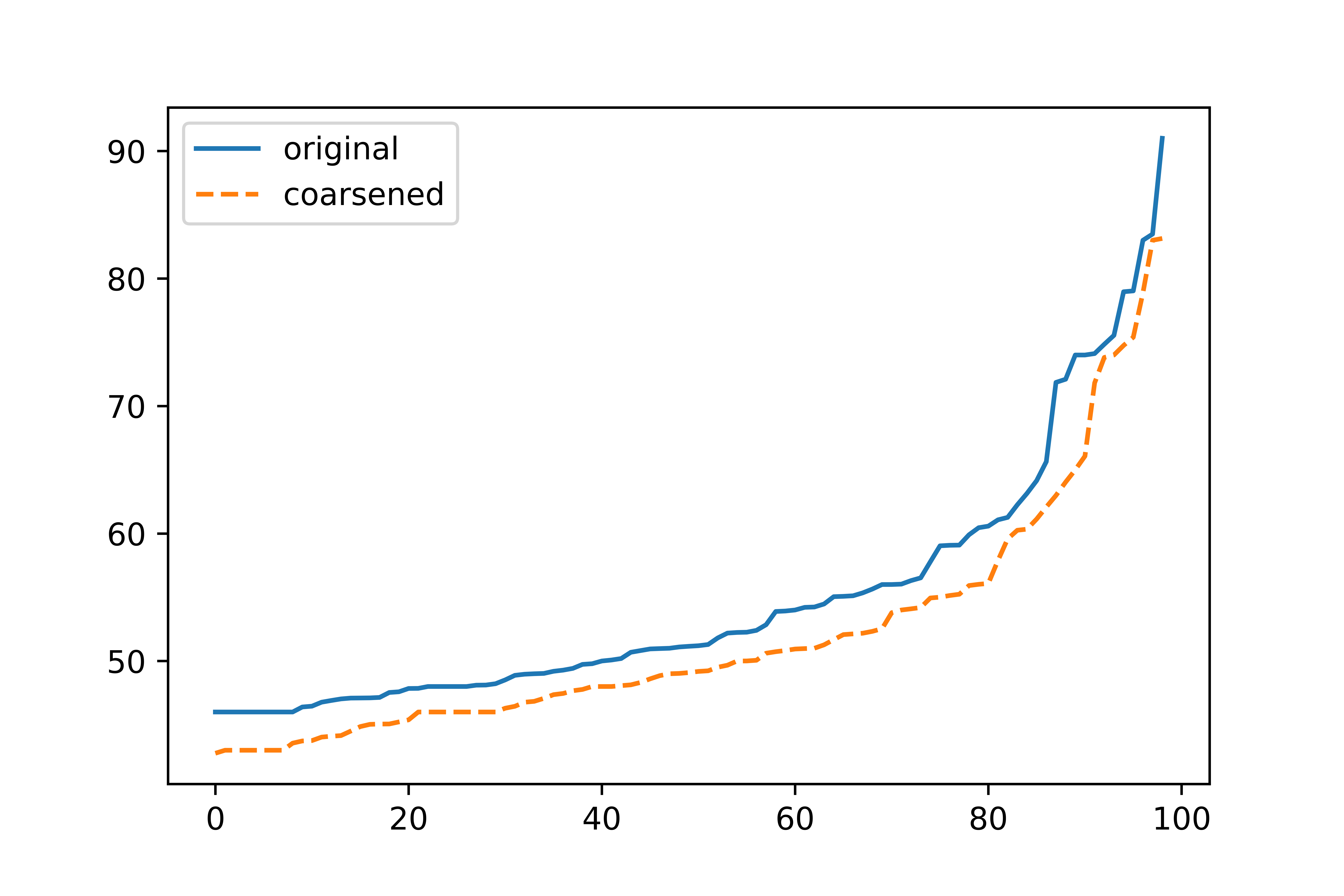}
 \label{FGC eigen values Improvised_acm_Eigen_3_7}
 \caption{ACM}
 \end{subfigure}
 \caption{This figure plots the top-100 eigenvalues of the coarsened graph obtained by FGCR against the original graph for Cora, Citeseer, Polblogs, and ACM datasets.} 
 \label{FGCR eigen values}
\end{figure}

\noindent \textbf{REE, HE and RE analysis:} To show the experimental correctness of the FGCR algorithm, we computed REE, HE, and RE on coarsening ratio $r=$0.7 for different reduction ratios $(rr) =$ 0.3, 0.5, and 0.7 respectively, which is defined by $rr=\frac{d}{n}$, where $n$ is the feature dimension corresponding to each node in the original graph data and $d$ is the feature dimension corresponding to each super-node in the coarsened graph. It is evident from Table \ref{tabel: FGCR HE, REE and RE} that the FGCR algorithm has a similar performance to the FGC algorithm in ensuring the properties of the original graph in the coarsened graph while reducing the feature dimension of each super-node of the coarsened graph. 
\begin{table}
\begin{center}
\begin{tabular}{ |m{4em} | m{4em} | m{4em}| m{4em} |m{4em}|m{4em}|} 
 \hline
 \multirow{1}{*}{Dataset} & $r$=$\frac{k}{p} $ & $rr$ & REE & HE & RE \\ 
 \hline
 Cora & 0.7 \textbf{\newline}0.7 \textbf{\newline}0.7 & 0.3\textbf{\newline} 0.5\textbf{\newline} 0.7 & 0.02 \textbf{\newline}0.02\textbf{\newline} 0.02 & 0.98\textbf{\newline} 0.92 \textbf{\newline}0.77 & 2.2\textbf{\newline} 2.17 \textbf{\newline}1.84 \\
 \hline
 Citeseer & 0.7 \textbf{\newline}0.7 \textbf{\newline}0.7 & 0.3 \textbf{\newline}0.5 \textbf{\newline}0.7 & 0.023 \textbf{\newline}0.02\textbf{\newline} 0.02 & 1.27\textbf{\newline} 1.09\textbf{\newline} 1.04& 2.140 \textbf{\newline}1.79 \textbf{\newline} 1.78\\ 
 \hline
 Polblogs & 0.7\textbf{\newline} 0.7\textbf{\newline} 0.7 & 0.3\textbf{\newline} 0.5 \textbf{\newline}0.7 & 0.09 \textbf{\newline}0.07 \textbf{\newline}0.03 & 2.195 \textbf{\newline}2.187\textbf{\newline} 2.067 & 6.30\textbf{\newline} 6.26 6.095 
 \\ 
 \hline
 ACM & 0.7\textbf{\newline} 0.7\textbf{\newline} 0.7 & 0.3\textbf{\newline} 0.5 \textbf{\newline}0.7 & 0.081 \textbf{\newline} 0.069 \textbf{\newline} 0.051& 0.85\textbf{\newline} 0.78 \textbf{\newline}0.69 & 4.30\textbf{\newline} 4.120 \textbf{\newline}3.95 
 \\ 
 \hline
 \end{tabular}
\end{center}
 \caption{REE, HE, and RE values for FGCR (proposed) algorithm on Cora, Citeseer, Polblogs, and ACM.}
\label{tabel: FGCR HE, REE and RE}
\end{table}
\noindent \textbf{Spectral Similarity:}
In Figure \eqref{FGCR eigen values}, we have shown the spectral similarity of the original graph and coarsened graph learned by FGCR. It is evident that the eigenvalue plot for coarsening ratio $r$=0.7 and the reduction ratio $rr=$0.7 is close to the original graph eigenvalues.
\subsection{Graph Coarsening under Adversarial Attack}
Graph Coarsening techniques rely on the edge connectivity of the graph network while reducing its data size. But in the real world, we may come across adversarial attacked graph data with poisoned edges or node features creating noise to it and therefore can mislead to poor graph coarsening of our graph data. Most of the adversarial attacks on the graph data are done by adding, removing, or re-wiring its edges \cite{runwal2022robust, dai2018adversarial}\\
As the state-of-the-art graph coarsening methods takes only the graph matrix as input to learning the coarsened graph. However, the FGC (proposed) method will use node features along with a graph matrix to learn the coarsened graph. Now, in adversarial attacked graph data, consider a poisoned edge $E_{ij}$ between the nodes $i$ and $j$ with features $X_{i}$ and $X_{j}$ respectively. Poisoned edge $E_{ij}$ wants to map $i$-th and $j$-th node into the same super-node as having an edge between them. But, actually, there is no edge between $i$-th and $j$-th node in the original graph (without any noise). However, if in the original graph(without noise) $X_{i}$ and $X_{j}$ are not having similar features, then the FGC algorithm reduces the probability of mapping $i$-th and $j$-th node into the same super-node, due to the smoothness property i.e., if two nodes are having similar features then there must be an edge between them. Finally, the smoothness or Dirichlet energy term in \eqref{Main formulation FGC1} opposes the effect of an extra edge due to an adversarial attack while doing the coarsening. Furthermore, we performed experiments on Cora, Citeseer, and ACM datasets by adding noise (extra edges) to their original graph structures. The results shown in Table \ref{tabel: Adversarial attack REE and DE} show that FGC (proposed) performs well even on noisy graph datasets as compared to other state-of-the-art algorithms of graph coarsening.
\noindent \textbf{REE and DE analysis}
We have attacked the real datasets by perturbation rate ($pr$) of 10\% and 5\%, i.e., the number of extra edges added to perturb the real datasets are 10\% or 5\% of the total number of edges in original graphs. We have compared the FGC (proposed), LVN, and LVE for REE and DE values. It is evident that FGC outperforms the existing state-of-the-art algorithm.

\begin{figure}
 \centering
 \begin{subfigure}[b]{0.32\textwidth}
 \centering
 \includegraphics[width=\textwidth]{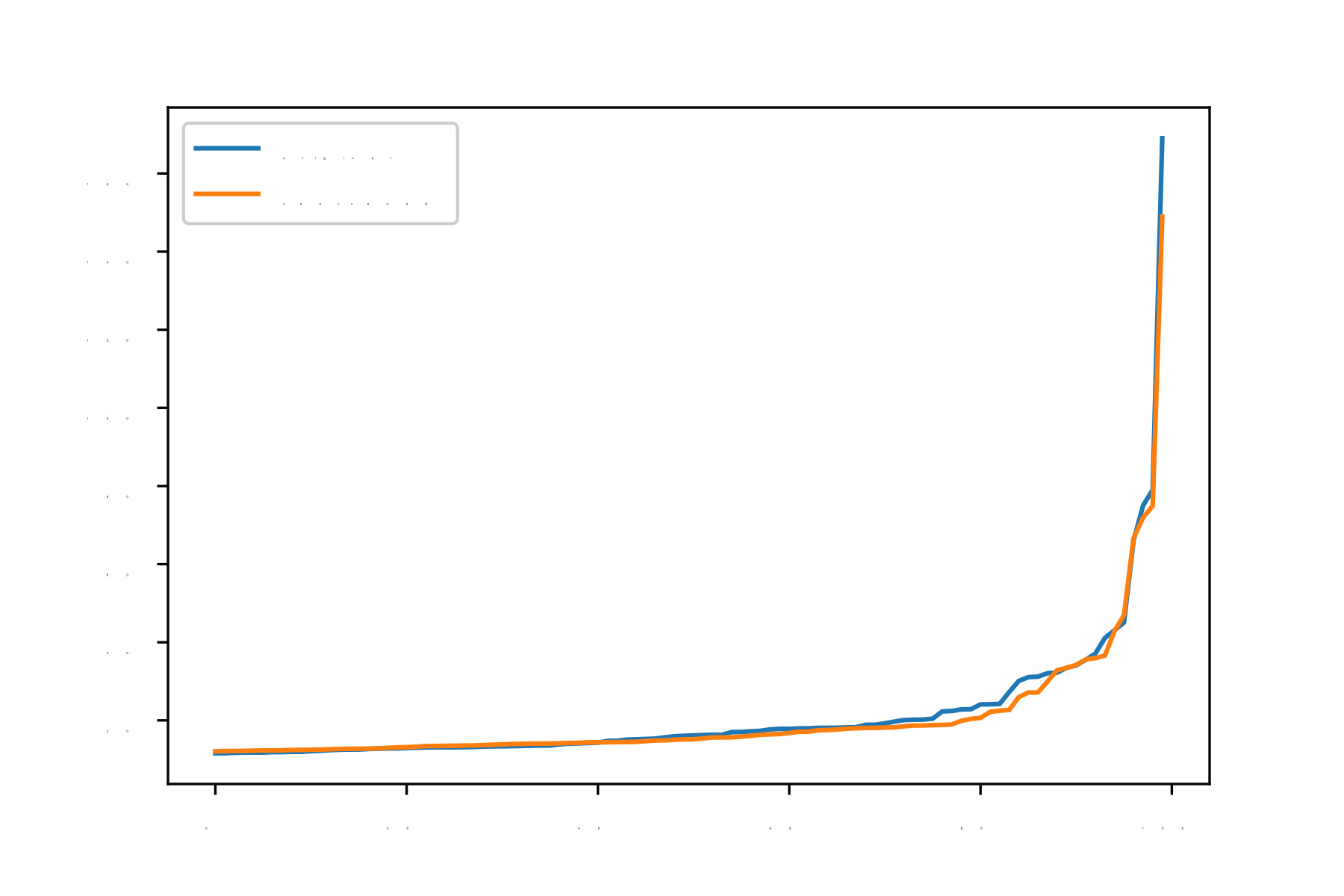}
 \label{adversarial attack eigen values Random_FGC_cora_Eigen_5_1}
 (a) Cora($pr=$ 10\%)
 \end{subfigure}
 \begin{subfigure}[b]{0.32\textwidth}
 \centering
 \includegraphics[width=\textwidth]{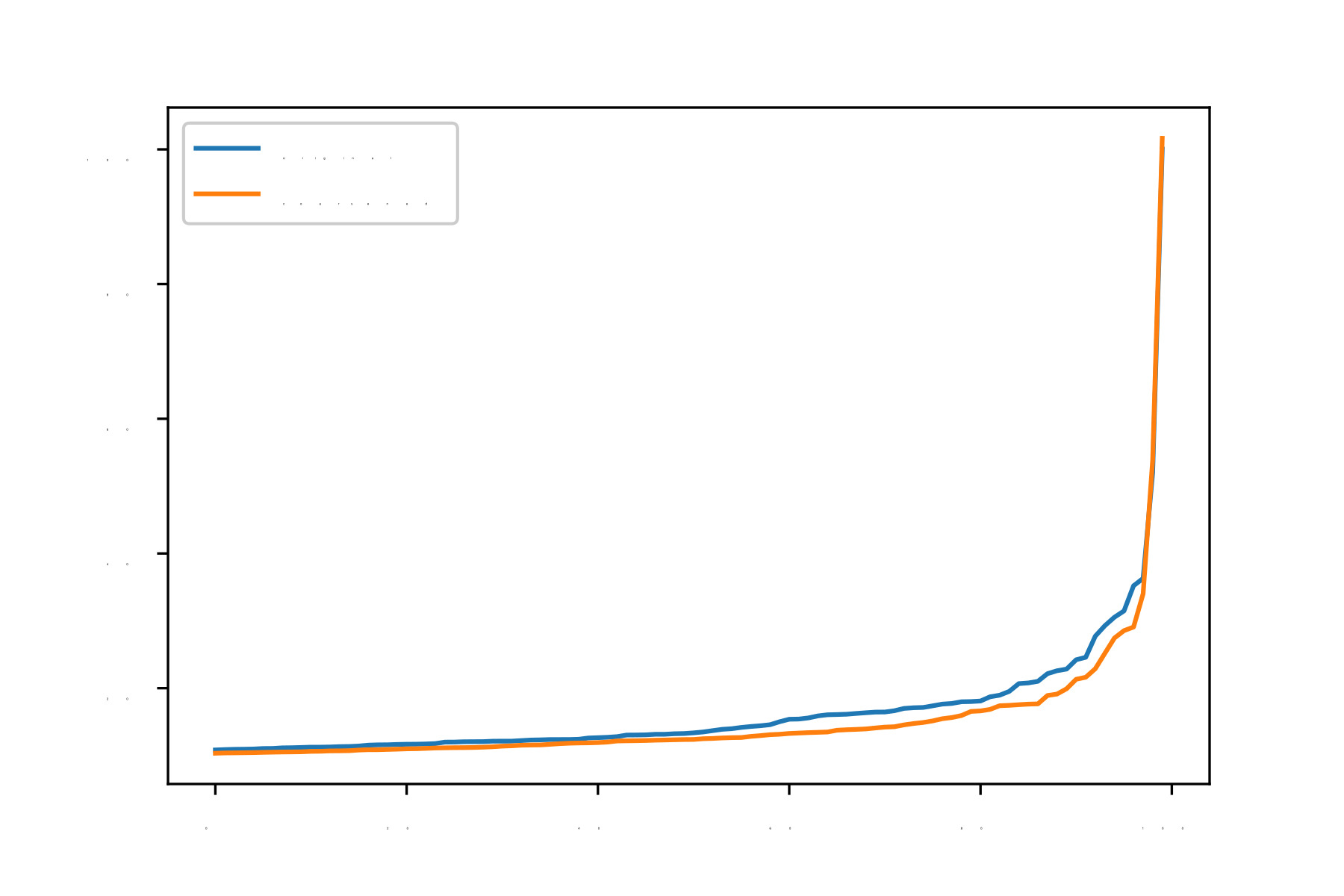}
 \label{adversarial attack eigen values Random_FGC_citeseer_Eigen_5_1}
 (b) Citeseer($pr=$ 10\%)
 \end{subfigure}
 \begin{subfigure}[b]{0.32\textwidth}
 \centering
 \includegraphics[width=\textwidth]{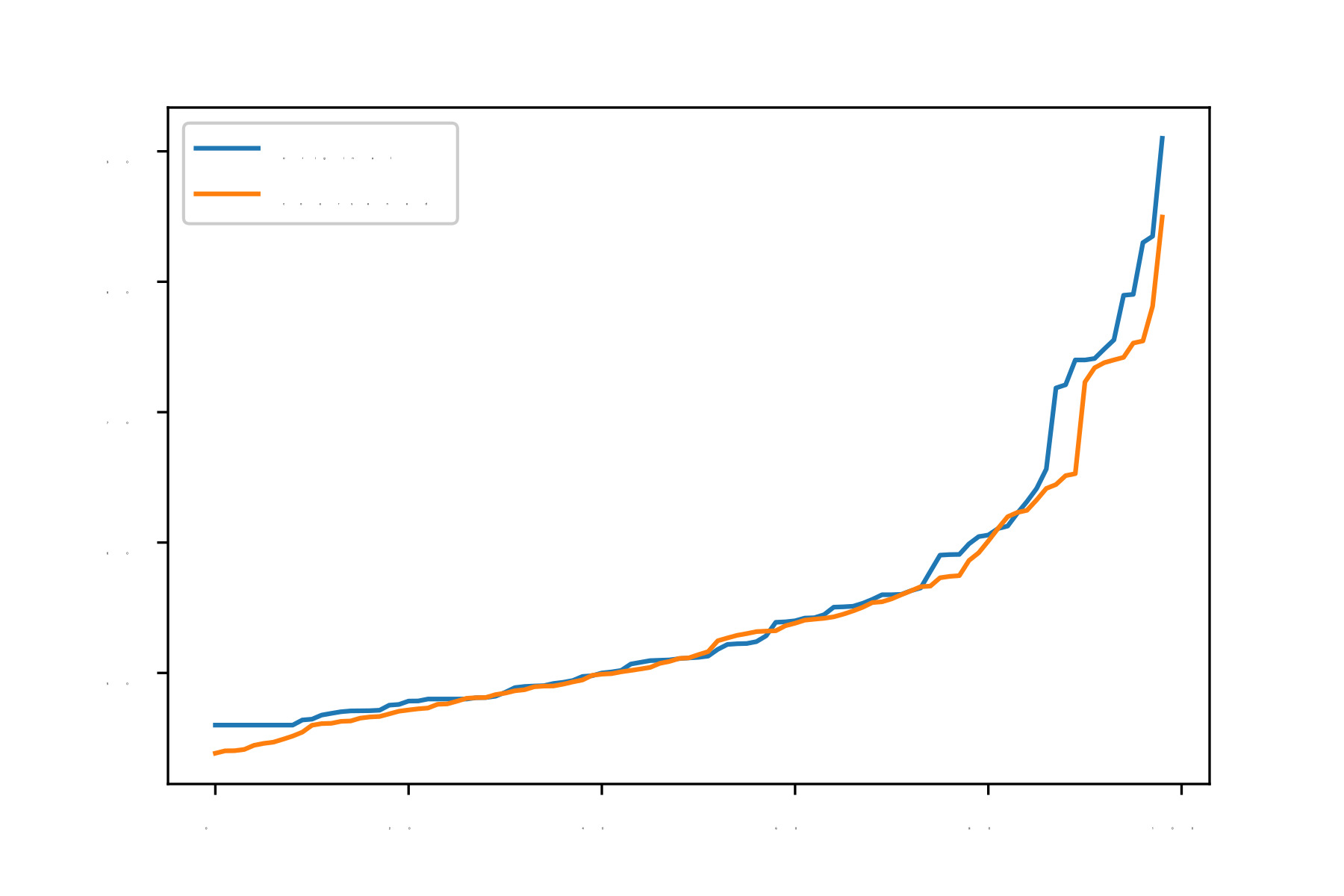}
 \label{adversarial attack eigen values Random_FGC_acm_Eigen_5_1}
 (c) ACM($pr=$ 10\%)
 \end{subfigure}
 \begin{subfigure}[b]{0.32\textwidth}
 \centering
 \includegraphics[width=\textwidth]{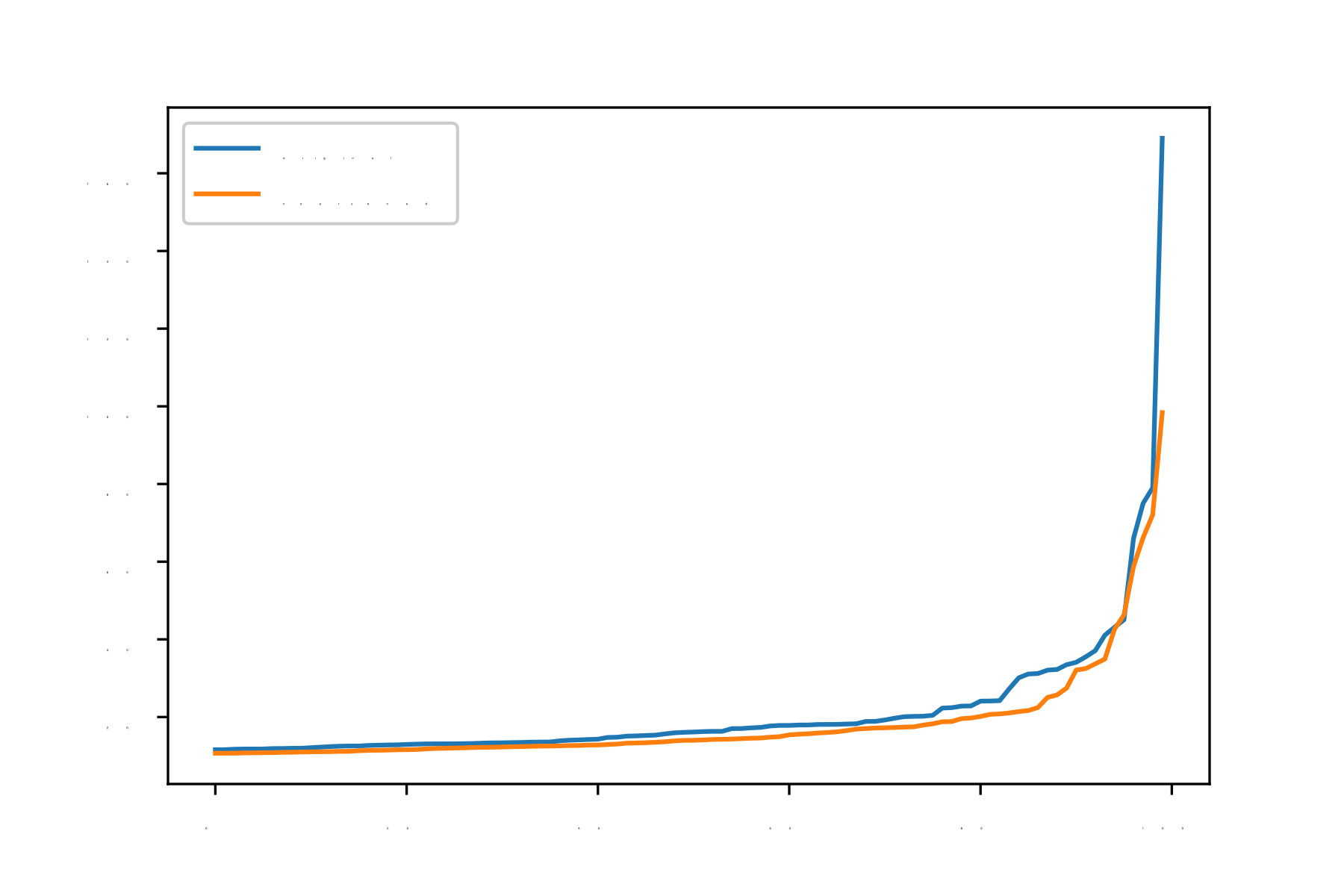}
 \label{adversarial attack eigen values Random_FGC_cora_Eigen_5_3}
 (d) Cora($pr=$ 5\%)
 \end{subfigure}
 \begin{subfigure}[b]{0.32\textwidth}
 \centering
 \includegraphics[width=\textwidth]{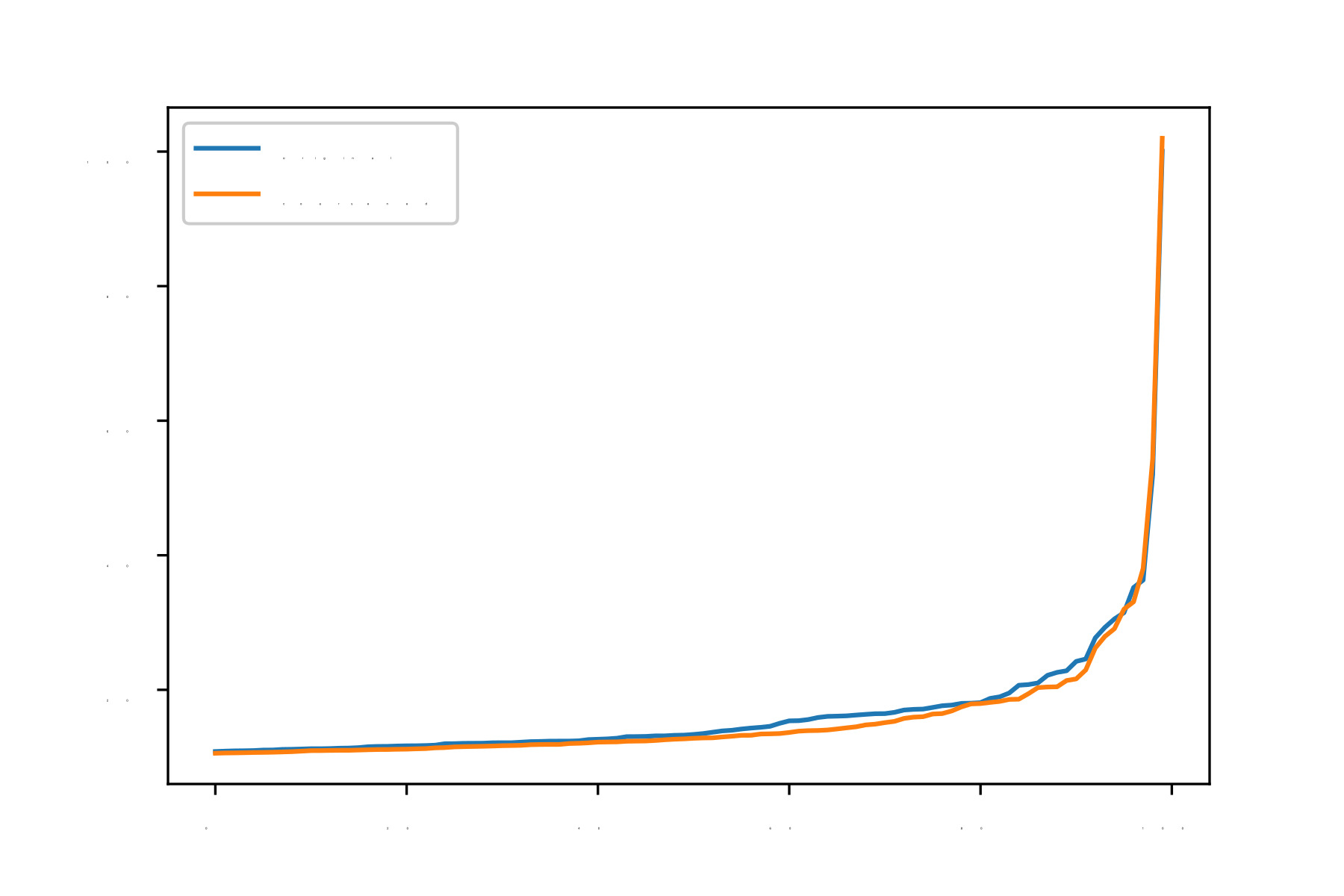}
 \label{adversarial attack eigen values Random_FGC_citeseer_Eigen_5_3}
 (e) Citeseer($pr=$ 5\%)
 \end{subfigure}
 \begin{subfigure}[b]{0.32\textwidth}
 \centering
 \includegraphics[width=\textwidth]{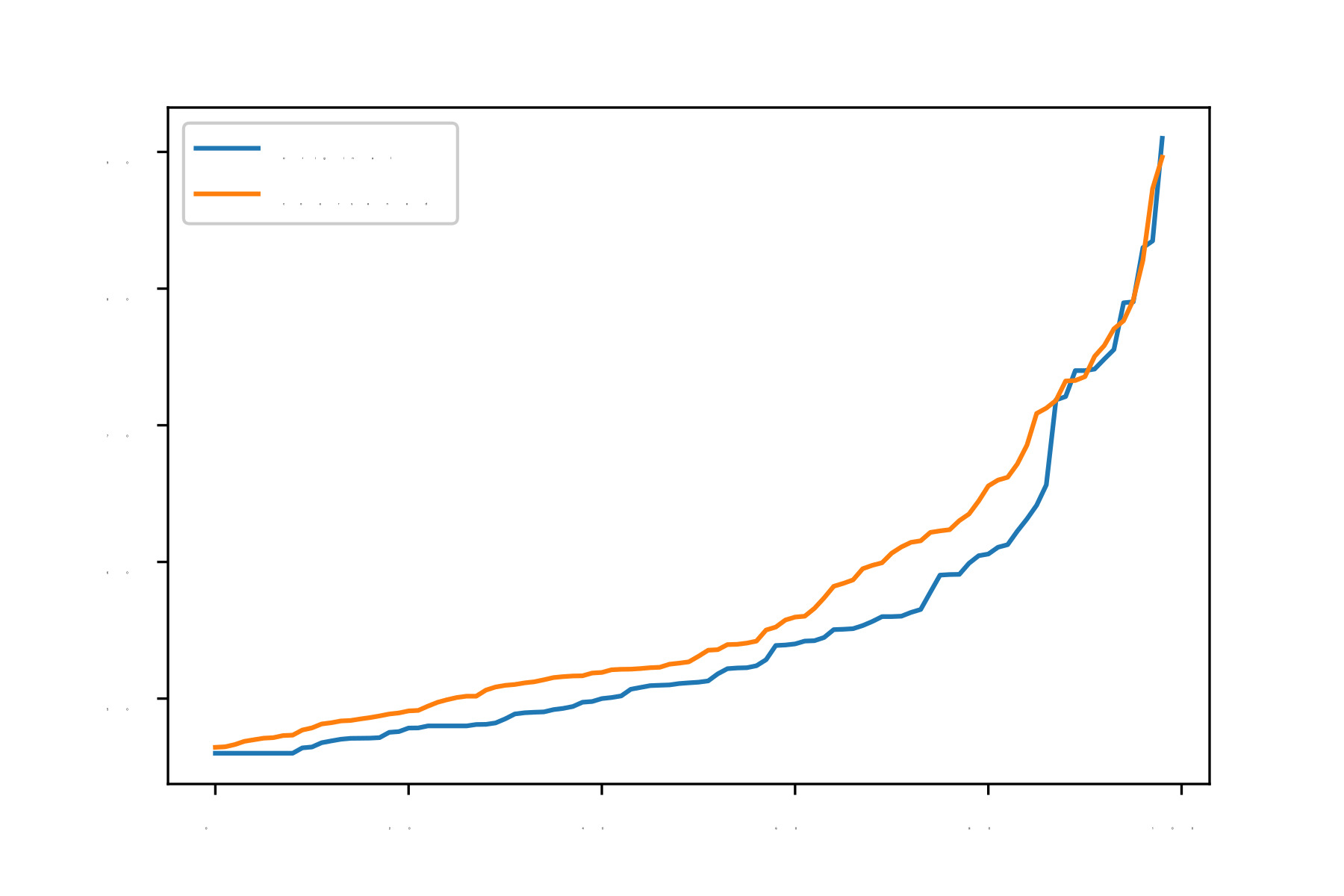}
 \label{adversarial attack eigen values Random_FGC_acm_Eigen_5_3}
 (f) ACM($pr=$ 5\%)
 \end{subfigure}
 \caption{Top-100 eigenvalues plots of original graph and coarsened graph obtained by FGC(proposed) algorithm and it is evident that FGC outperforms state of the art algorithm on 10\% and 5\% perturbation attack.}
 \label{adversarial attack eigen values}
\end{figure}

\begin{table}
\begin{center}
\begin{tabular}{ |m{1.5cm}|m{2em} | m{3em} | m{3em}| m{3em} |m{3em} | m{3em}|m{4em}|m{4em}|} 
 \hline
 \multirow{2}{*}{Dataset} & \multirow{2}{*}{$r$=$\frac{k}{p} $} & 
 \multicolumn{4}{c|}{$\text{REE}(\Theta,\Theta_c,100)$} & \multicolumn{3}{c|}{$\text{DE}$}\\
 \cline{3-9}
 & &$pr(\%)$ & FGC & LVN & LVE & FGC & LVN & LVE\\ 
 \hline
 Cora & 0.3 0.3 0.5 0.5 & 10 \textbf{\newline} 5 \textbf{\newline} 10 \textbf{\newline} 5 & \textbf{0.084 0.069 0.048 0.047} & 0.615 0.614 0.483 0.482 & 0.668 0.693 
 0.470 0.498 & \textbf{6724 6310 7734 7336} & 39575 36686 70348 66447 & 35719 32137 69263 65606 \\ 
 \hline
 Citeseer & 0.3 0.3 0.5 0.5 & 10 \textbf{\newline} 5 \textbf{\newline} 10 \textbf{\newline} 5 & \textbf{0.084 0.063 0.088 0.072} & 0.715 0.718 0.539 0.520 & 0.710 0.728 0.493 0.507 & \textbf{6759 5895 6239 6565} & 41730 36611 90022 84303 & 41300 34897 92593 84476\\ 
 \hline
 ACM & 0.3 0.3 0.5 0.5 & 10 \textbf{\newline} 5 \textbf{\newline} 10 \textbf{\newline} 5 & \textbf{0.027 0.029 0.017 0.011 }& 0.812 0.872 0.643 0.594 & 0.650 0.720 0.367 0.357 & \textbf{12741 11822 15563 16239 } & 180816 110788 418559 442640 & 244215 177451 580695 557645 \\ 
 \hline
\end{tabular}
\end{center}
 \caption{REE and DE results for Cora, Citeseer, and ACM for 10\% and 5\% perturbation by FGC (proposed), LVN, and LVE.}
\label{tabel: Adversarial attack REE and DE}
\end{table}
\noindent \textbf{Spectral Similarity:}
In this section, we have shown in Figure \ref{adversarial attack eigen values} the spectral similarity of the coarsened graph obtained by FGC and the original graph using eigenvalues plots on different datasets for coarsening ratio $r$=0.3 and perturbation rate $pr =$ 10\% and 5\%.
\subsection{Application of FGC in Classification}
\begin{figure}
 \centering
 \begin{subfigure}[b]{1.0\textwidth}
 \centering
 \includegraphics[width=\textwidth]{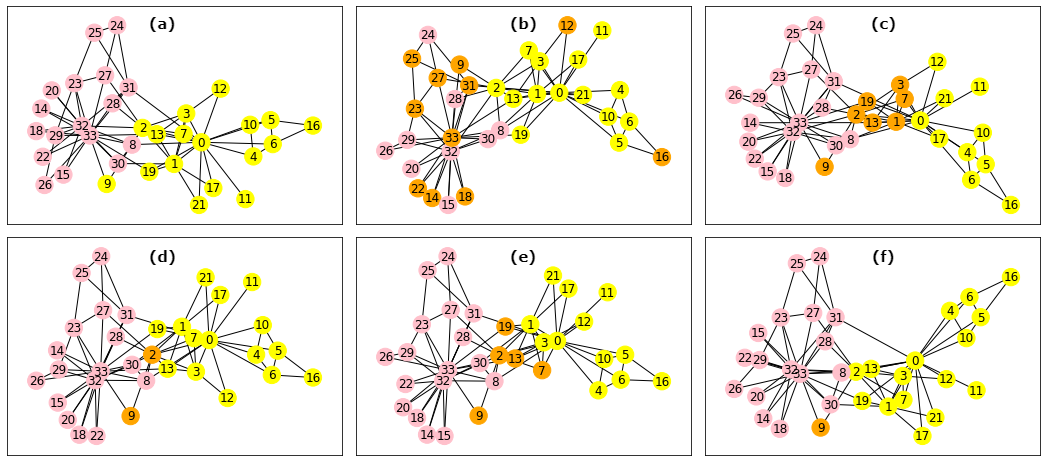}
 \label{karate final}
 \end{subfigure}
 \caption{ This figure evaluates the classification performance of the FGC algorithm on the classic Zachary's karate club dataset \cite{zachary1977information} into 2 classes: (a) Ground truth, (b) Graclus\cite{dhillon2007weighted}, (c) spectral clustering ratio cut \cite{ng2001spectral} (d) spectral clustering normalized cut \cite{ng2001spectral} (e) LVN \cite{loukas2019graph} and (f) FGC (Proposed). Orange nodes indicate misclassified points, FGC demonstrates a better performance, it resulted in only 1 misclassified point, while the number of misclassified points for (b), (c), (d), and (e) are 11, 7, 2 and 5, respectively.} 
 \label{Karate 2}
\end{figure}

In this section, we present one of the many applications where FGC can be used which is node-based classification. Zachary's karate club is a social network that consists of friendships among members of a university-based karate club. This dataset consists of 34 nodes, 156 edges, and 2 classes. We aim to classify these nodes into two groups. Its graph is shown in Figure \ref{Karate 2}(a) with two classes where class-1 is colored in pink and class-2 is colored in yellow. We performed experiments on FGC, graph clustering techniques, and state-of-the-art graph coarsening method, i.e., LVN, to classify these 34 nodes into 2 classes or two super-node. FGC (proposed) classification performance also validates the importance of features of graph data during graph coarsening and till now none of the pre-existing coarsening or clustering methods have been taken into account. For the FGC algorithm feature matrix, $X$ of size ${34 \times n}$ is generated by sampling from $X \sim \mathcal{N}(\mathbf{0}, \Theta^\dagger)$, where $\Theta$ is the Laplacian matrix of the given network.
Below, we have shown the node classification results of multiple techniques by coloring members of each group, i.e., super-node, with the same color i.e. members of super-node 1 or group 1 are colored in pink and members of super-node 2 or group 2 are colored in yellow. Moreover, the nodes sent to the wrong groups are colored in orange which means they are misclassified. The results of FGC below are shown for $n=600$, however, an $n$ of the order of $5* 34$ has been observed to perform well.

\begin{figure}
 \centering
 \begin{subfigure}[b]{0.45\textwidth}
 \centering
 \includegraphics[width=\textwidth]{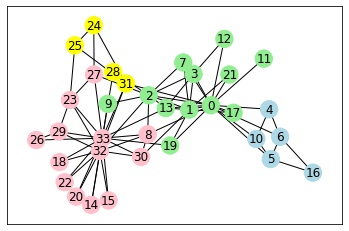}
 \caption{Ground truth} 
 \label{karate 4 true}
 \end{subfigure}
 \begin{subfigure}[b]{0.45\textwidth}
 \centering
 \includegraphics[width=\textwidth]{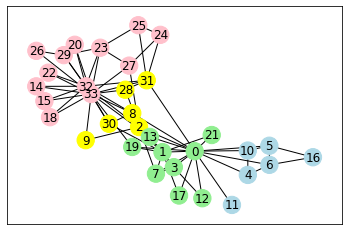}
 \caption{Spectral clustering ratio cut}
 \label{karate 4 SCC}
 \end{subfigure}
 \begin{subfigure}[b]{0.45\textwidth}
 \centering
 \includegraphics[width=\textwidth]{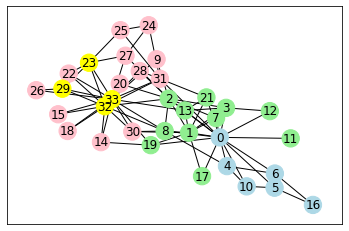}
 \caption{Spectral clustering normalized cut} 
 \label{karate 4 SCR}
 \end{subfigure}
 \begin{subfigure}[b]{0.45\textwidth}
 \centering
 \includegraphics[width=\textwidth]{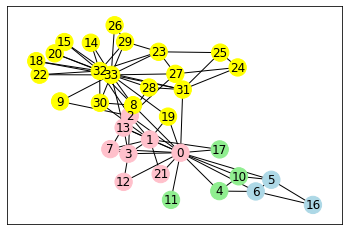}
 \caption{LVN} 
 \label{karate 4 LVN}
 \end{subfigure}
 \begin{subfigure}[b]{0.45\textwidth}
 \centering
 \includegraphics[width=\textwidth]{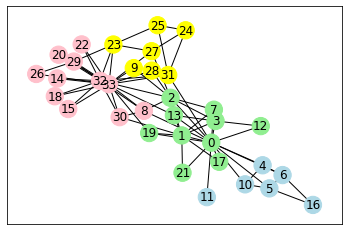}
 \caption{FGC(proposed)} 
 \label{karate 4 FGC}
 \end{subfigure}
 \caption{This figure evaluates the classification performance of the FGC algorithm on the classic Zachary's karate club dataset \cite{zachary1977information} into 4 classes: (a) Ground truth, (b) spectral clustering ratio cut \cite{ng2001spectral} (c) spectral clustering normalized cut \cite{ng2001spectral} (d) LVN \cite{loukas2019graph} and (e) FGC (Proposed). It is evident that FGC demonstrates a better performance, it resulted in 4 misclassified point, while the number of misclassified points for (b), (c) and (d) are 7, 11, and 24 respectively.}
 \label{Karate 4}
\end{figure}

We have also performed classification of these 34 nodes of Karate club dataset into 4 groups or 4 supernode using spectral clustering ratio cut, spectral clustering normalized cut, LVN and FGC (proposed) and it is evident in Figure \ref{Karate 4} that classification accuracy of FGC(proposed) is highest as compared to other state of the art algorithms.
Similarly, we have performed a classification of polblogs dataset into 2 classes. Here, the input is a political blog consisting of 1490 nodes, the goal is to classify the nodes into two groups. For the FGC algorithm, the feature matrix $X$ of size ${1490 \times 5000}$ is generated by sampling from $X \sim \mathcal{N}(\mathbf{0}, \Theta^\dagger)$, where $\Theta$ is the Laplacian matrix of the given network. The FGC algorithm and Graclus correctly classify 1250 and 829 nodes respectively. However, the performance of LVN and spectral clustering are not competent. The FGC result also demonstrates that the features may help in improving the graph-based task, and for some cases like the one presented here the features can also be artificially generated governed by the smoothness and homophily properties.

\subsection{Effect of Hyperparameters}
\begin{figure}
\centering
\begin{subfigure}[b]{0.19\textwidth}
\centering
\includegraphics[width=\textwidth]{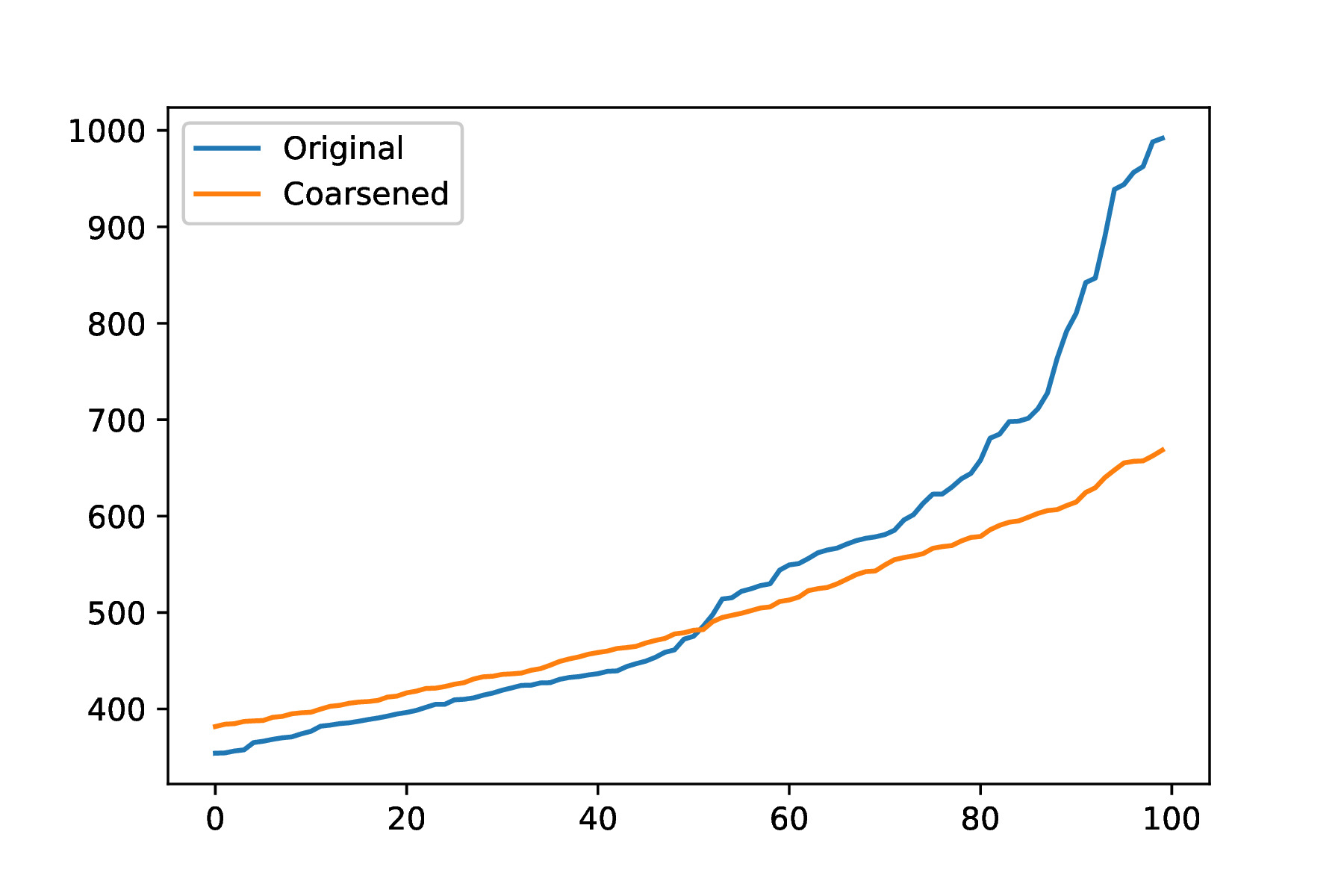}
\label{g_20_gamma0}
\caption{$\gamma$ = 200}
\end{subfigure}
\begin{subfigure}[b]{0.19\textwidth}
\centering
\includegraphics[width=\textwidth]{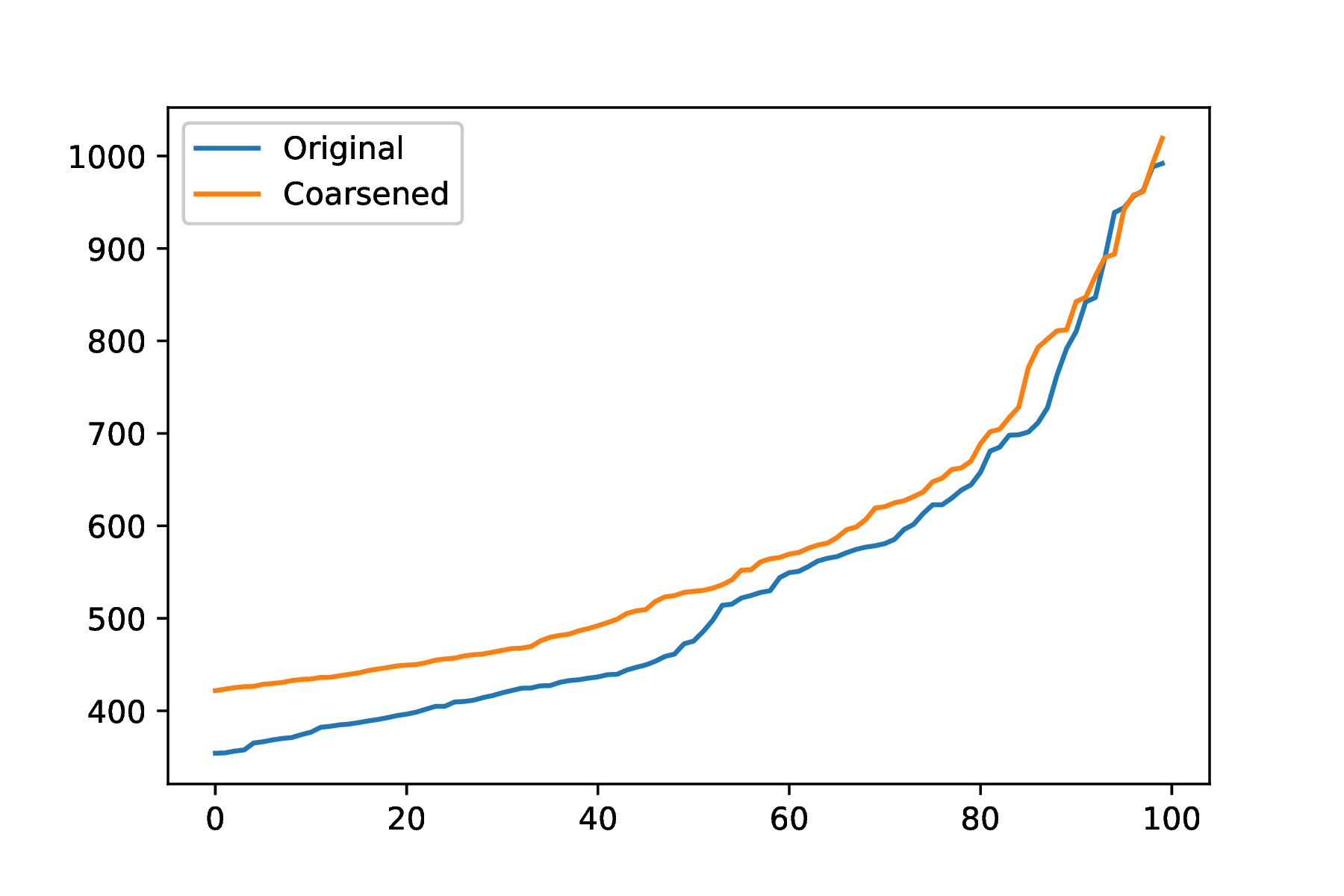}
\label{g_800_gamma}
\caption{$\gamma$ = 800}
\end{subfigure}
\begin{subfigure}[b]{0.19\textwidth}
\centering
\includegraphics[width=\textwidth]{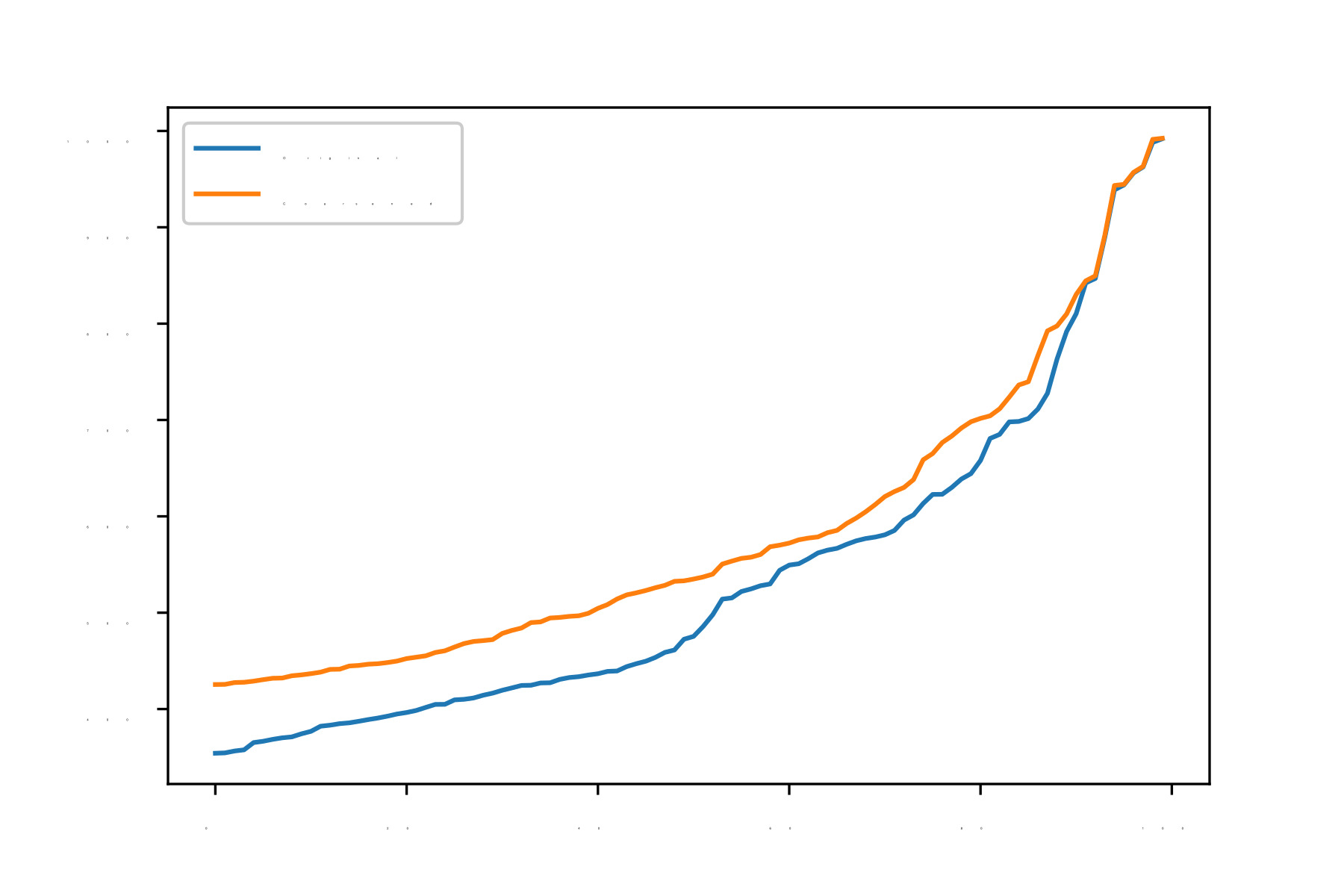}
\label{g_1000_gamma}
\caption{$\gamma$ =1000}
\end{subfigure}
\begin{subfigure}[b]{0.19\textwidth}
\centering
\includegraphics[width=\textwidth]{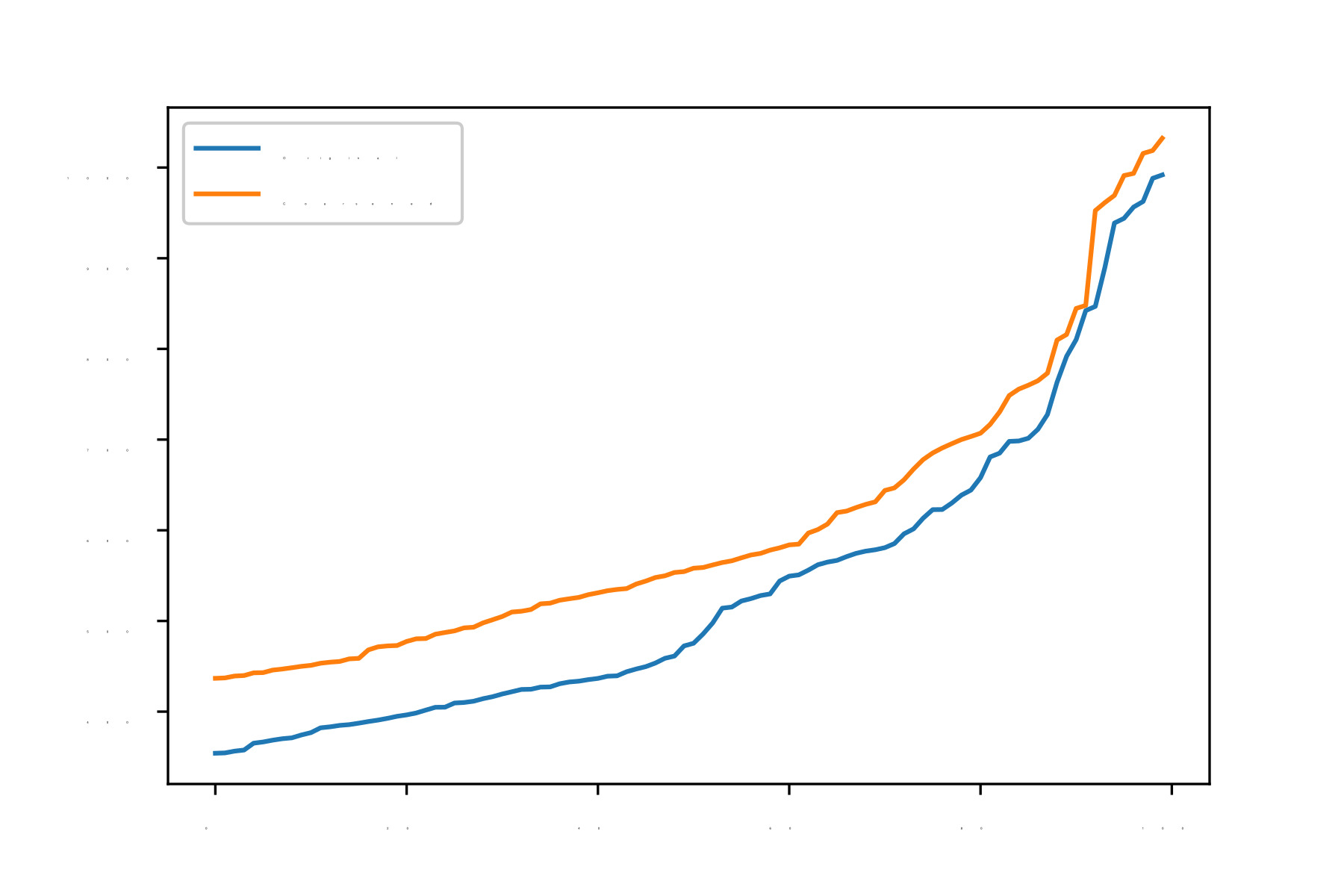}
\label{g_2000_gamma}
\caption{$\gamma$ = 2000}
\end{subfigure}
\begin{subfigure}[b]{0.19\textwidth}
\centering
\includegraphics[width=\textwidth]{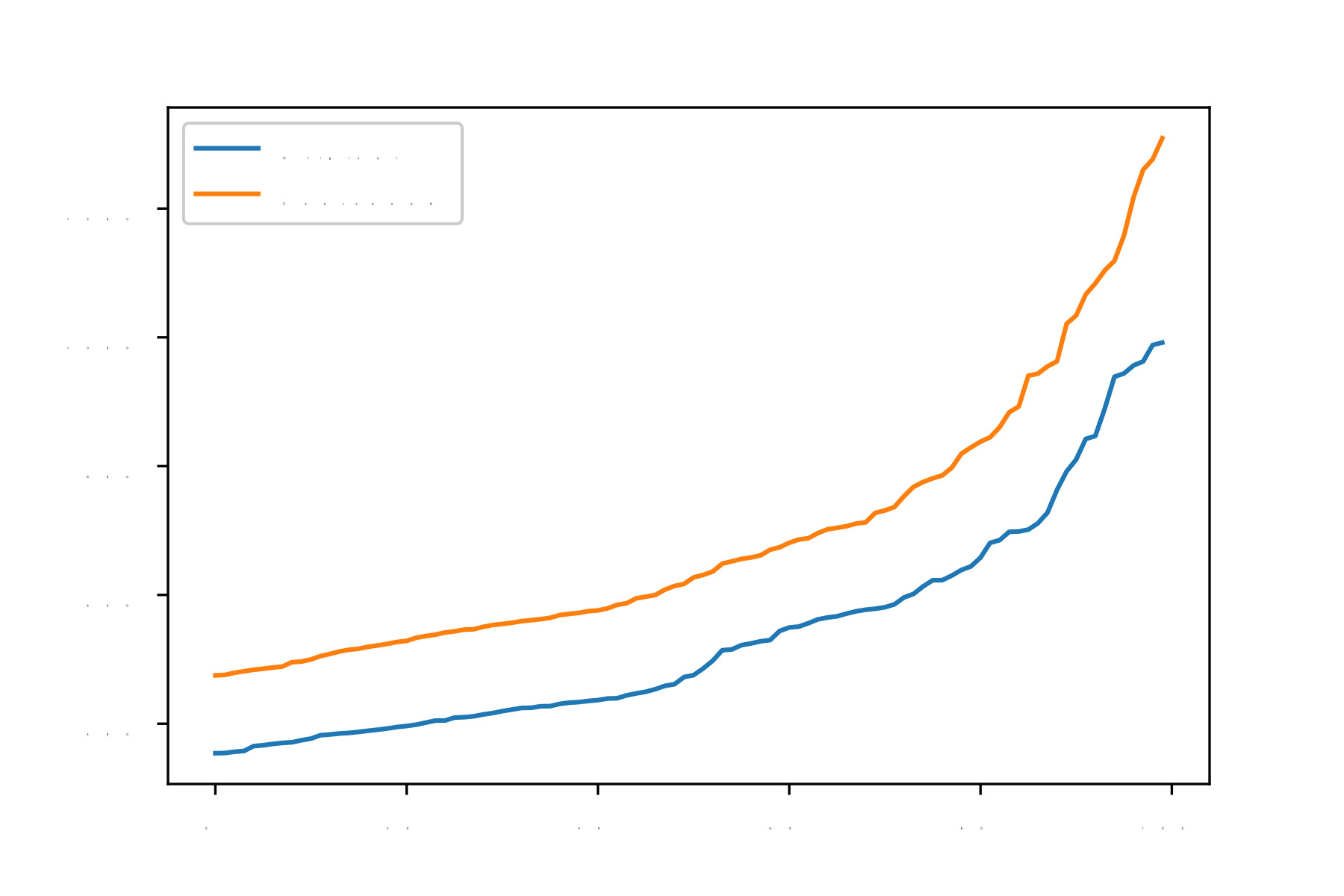}
\label{g_5000_gamma}
\caption{$\gamma$ = 5000}
\end{subfigure}
\caption{Fig.(a-e) shows the eigenvalue plot of original graph and coarsened graph obtained by FGC using hyperparameters $\alpha=500$, $\lambda=1000$ and varying $\gamma$ in between (100-50000). It is evident that for a moderate $\gamma$ i.e., between 200 to 2000, the REE is almost similar and our algorithm is consistent.}
 \label{Visualization gamma}
\end{figure}

\begin{figure}[H]
\centering
\begin{subfigure}[b]{0.19\textwidth}
\centering
\includegraphics[width=\textwidth]{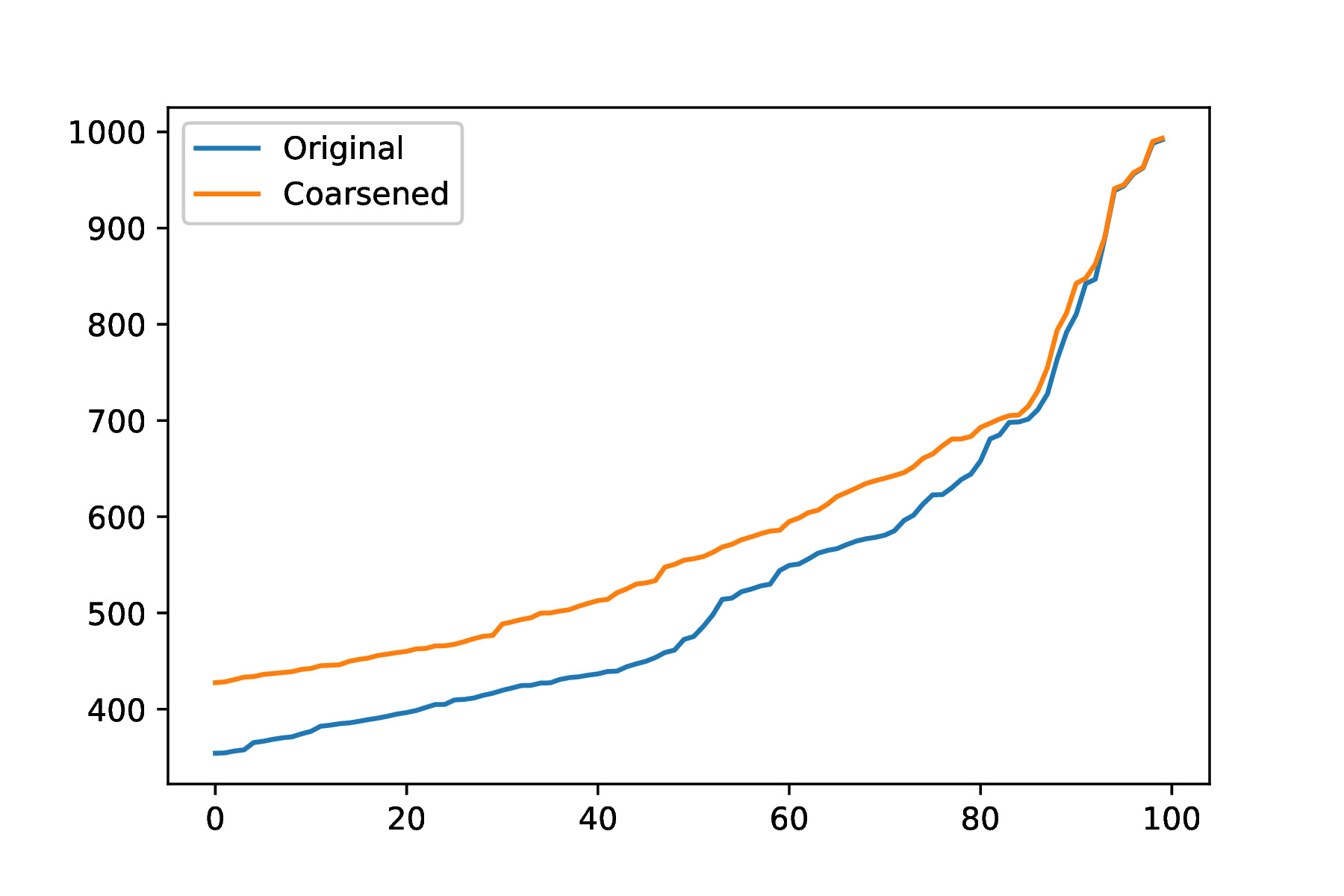}
\label{g_200_alpha}
\caption{$\alpha$ = 200}
\end{subfigure}
\begin{subfigure}[b]{0.19\textwidth}
\centering
\includegraphics[width=\textwidth]{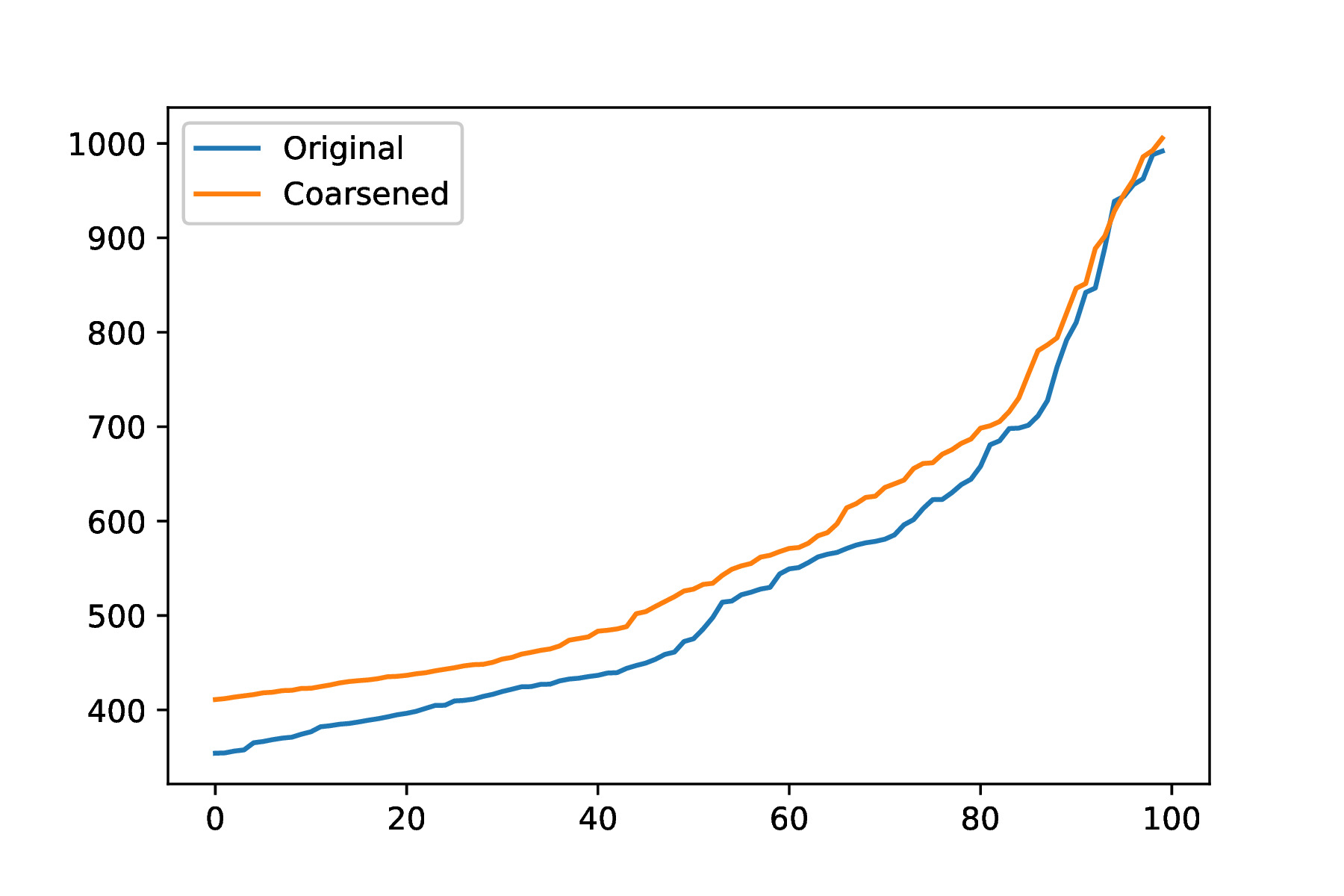}
\label{g_800_alpha}
\caption{$\alpha$ = 800}
\end{subfigure}
\begin{subfigure}[b]{0.19\textwidth}
\centering
\includegraphics[width=\textwidth]{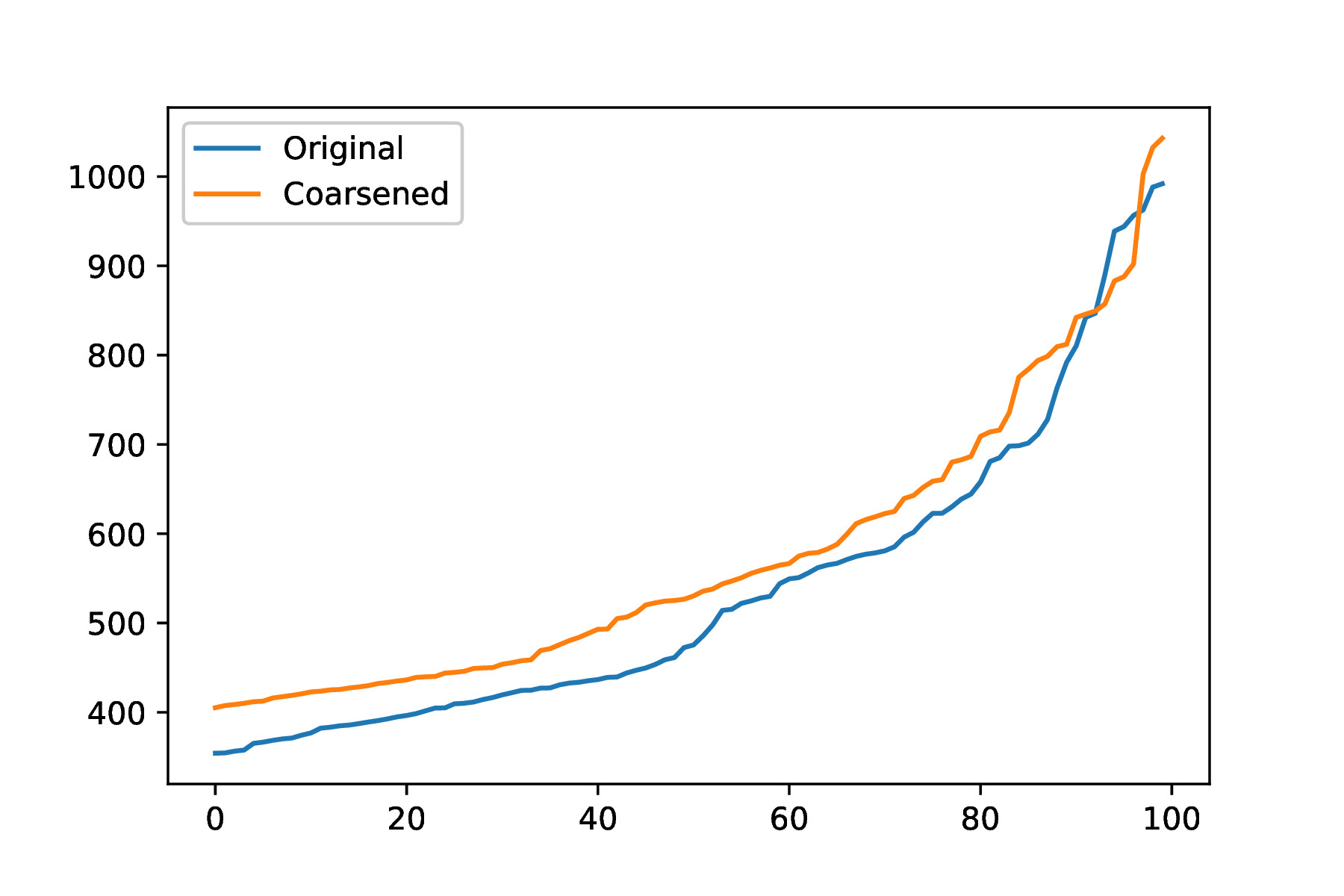}
\label{g_1000_alpha}
\caption{$\alpha$ = 1000}
\end{subfigure}
\begin{subfigure}[b]{0.19\textwidth}
\centering
\includegraphics[width=\textwidth]{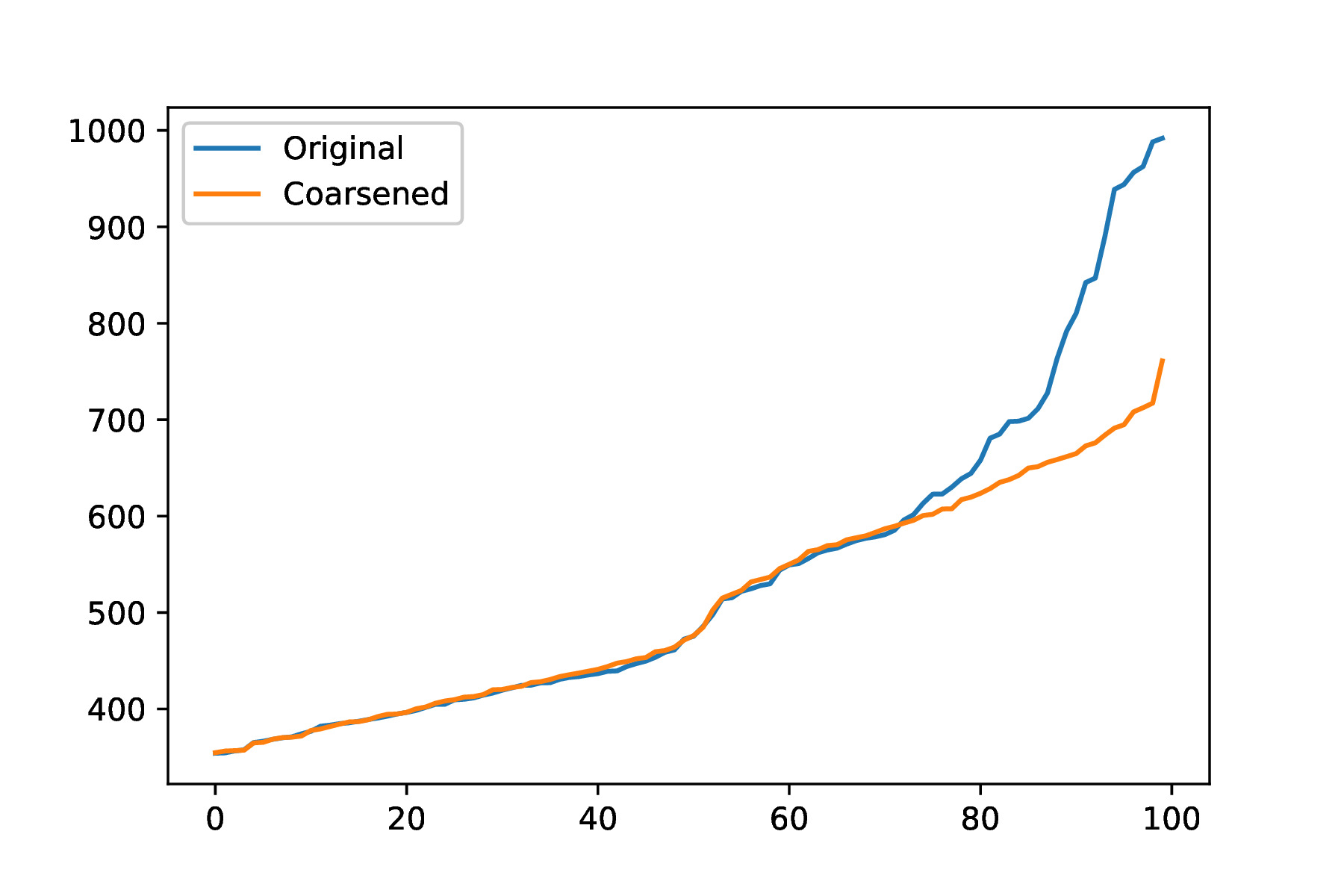}
\label{g_2000_alpha}
\caption{$\alpha$ = 2000}
\end{subfigure}
\begin{subfigure}[b]{0.19\textwidth}
\centering
\includegraphics[width=\textwidth]{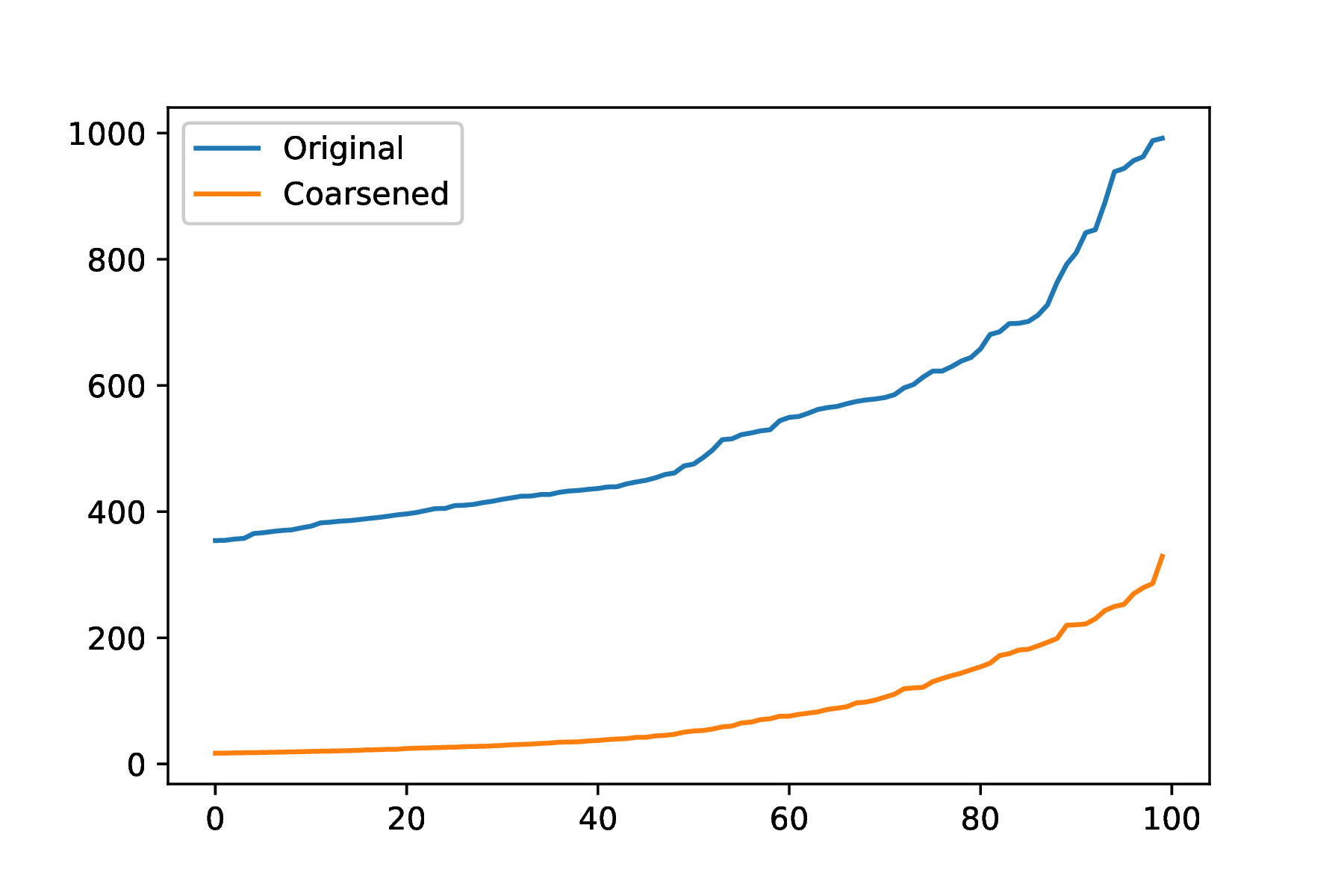}
\label{g_5000_alpha}
\caption{$\alpha$ = 5000}
\end{subfigure}
\caption{Fig.(a-e) shows the eigen value plot of original graph and coarsened graph obtained by FGC using hyperparameters $\lambda=1000$, $\gamma=600$ and varying $\alpha$ in between (100-50000). It is evident that for a moderate $\alpha$ i.e., between 100 to 2000, the REE is almost similar and our algorithm is consistent.}
 \label{Visualization alpha}
\end{figure}

\begin{figure}
\centering
\begin{subfigure}[b]{0.19\textwidth}
\centering
\includegraphics[width=\textwidth]{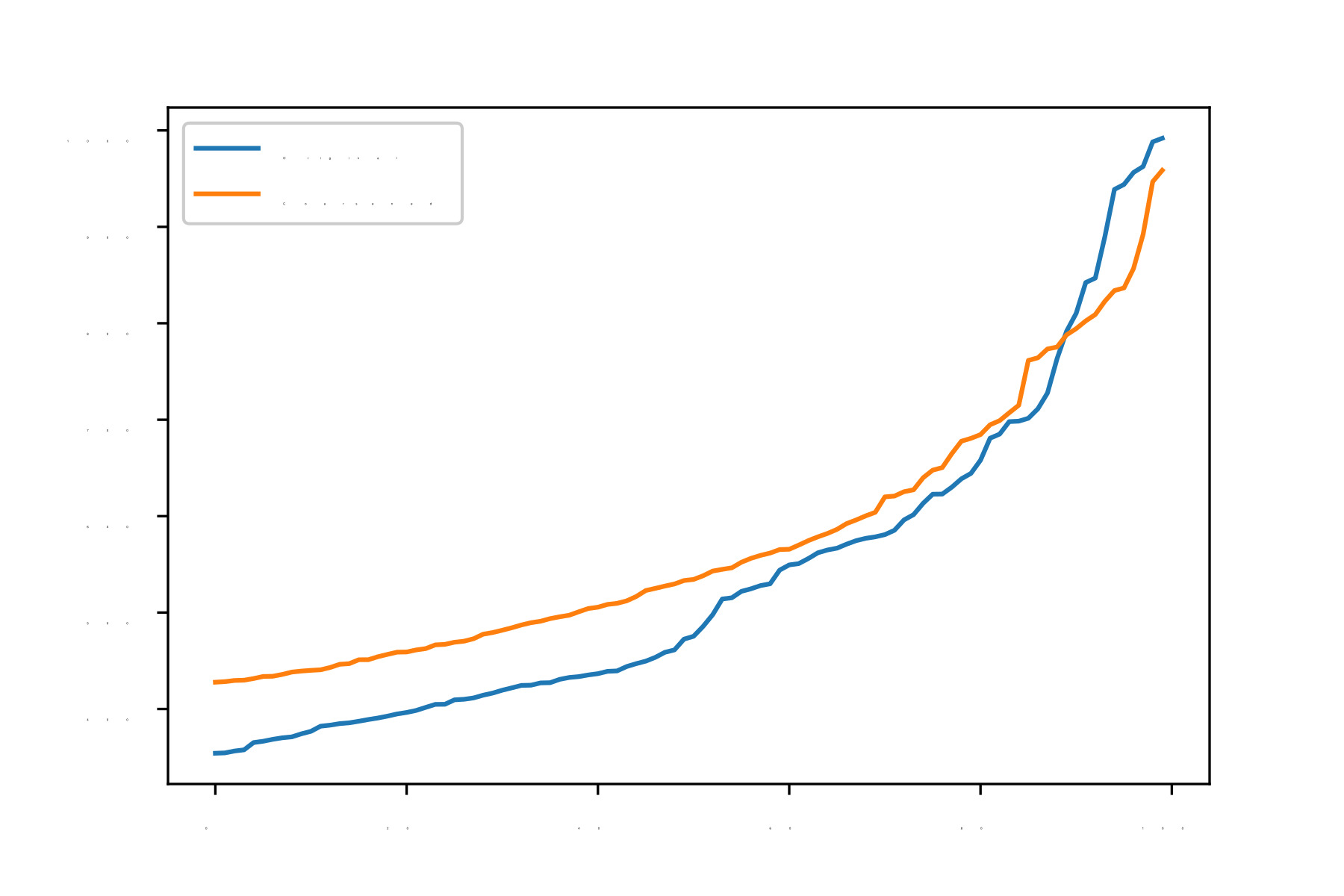}
\label{g_200_lambda}
\caption{$\lambda$ = 200}
\end{subfigure}
\begin{subfigure}[b]{0.19\textwidth}
\centering
\includegraphics[width=\textwidth]{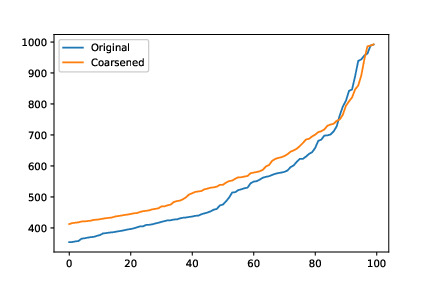}
\label{g_800_lambda}
\caption{$\lambda$ = 800}
\end{subfigure}
\begin{subfigure}[b]{0.19\textwidth}
\centering
\includegraphics[width=\textwidth]{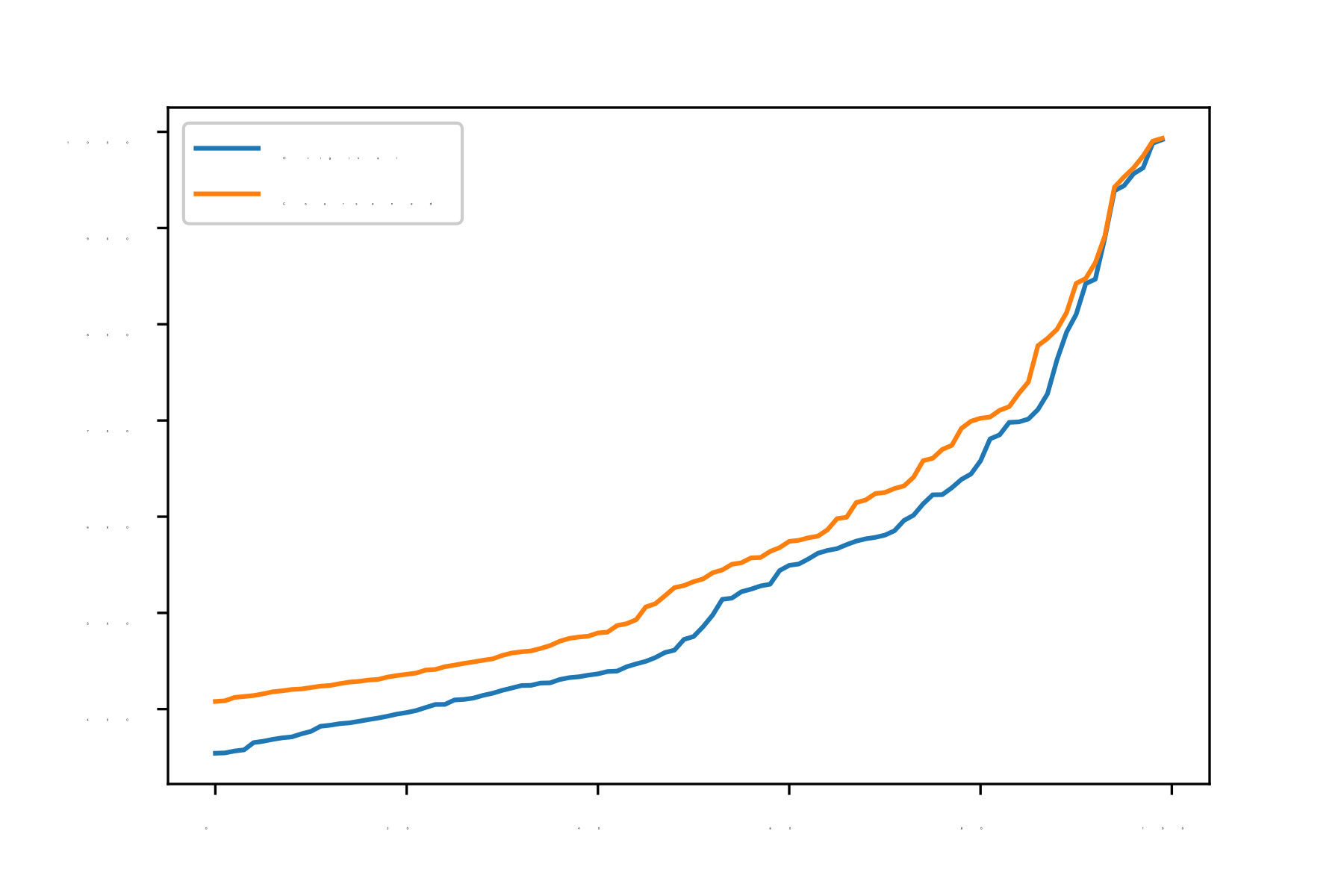}
\label{g_1000_lambda}
\caption{$\lambda$ = 1000}
\end{subfigure}
\begin{subfigure}[b]{0.19\textwidth}
\centering
\includegraphics[width=\textwidth]{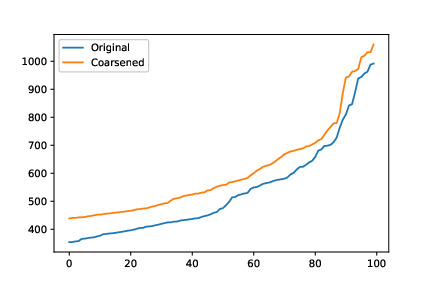}
\label{g_2000_lambda}
\caption{$\lambda$ = 2000}
\end{subfigure}
\begin{subfigure}[b]{0.19\textwidth}
\centering
\includegraphics[width=\textwidth]{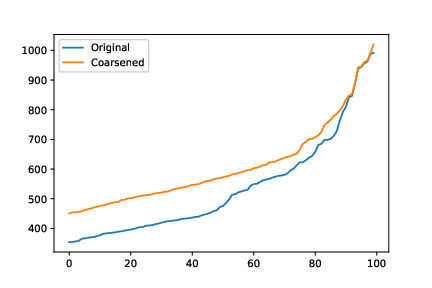}
\label{g_5000_lambda}
\caption{$\lambda$ = 5000}
\end{subfigure}
\caption{Fig.(a-e) shows the eigenvalue plot of original graph and coarsened graph obtained by FGC using hyperparameters $\alpha=500$, $\gamma=1000$ and varying $\lambda$ in between (100-50000). It is evident that for a moderate $\lambda$ i.e., between 100 to 2000, the REE is almost similar and our algorithm is consistent.}
 \label{Visualization lambda}
\end{figure}

The FGC algorithm has 3 hyperparameters:(i) $\gamma$ for ensuring the coarsen graph is connected, (ii) $\alpha$ to learn $\tilde{X}$ correctly (iii) $\lambda$ to enforce sparsity and orthogonality on loading matrix $C$. From figures \ref{Visualization gamma}, \ref{Visualization alpha}, and \ref{Visualization lambda}, it is observed that the algorithm is not sensitive to the hyperparameters $(\lambda, \gamma, \alpha)$ any moderate value of can be used for the FGC algorithm.

\subsection{Affect of features on Coarsening: Toy Example}
Here we demonstrate that the feature plays an important role in obtaining a coarsened graph matrix. Consider two given graph data $\mathcal{G}(\Theta, X1)$
and $\mathcal{G}(\Theta, X2)$ have the same graph matrices but with different associated features. The coarsened graph matrices obtained with the FGC algorithm for these two datasets will be different, while the methods like LVN and LVE which do not consider the features while doing coarsening will provide the same coarsening graph matrix for these two different datasets. See the below figures for the demonstration.
\begin{figure}
 \centering
 \begin{subfigure}[b]{0.45\textwidth}
 \centering
 \includegraphics[width=0.7\textwidth]{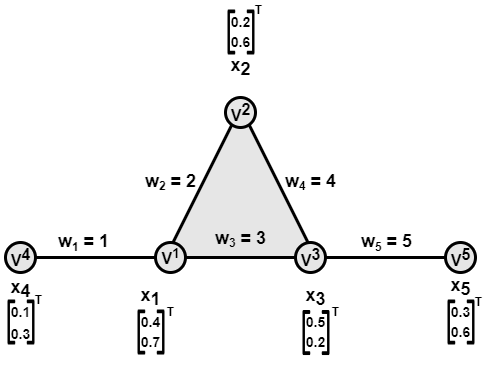}
 \caption{Original graph $\mathcal{G}(\Theta, X1)$}
 \end{subfigure}
 \hfill
 \begin{subfigure}[b]{0.45\textwidth}
 \centering
 \includegraphics[width=0.9\textwidth]{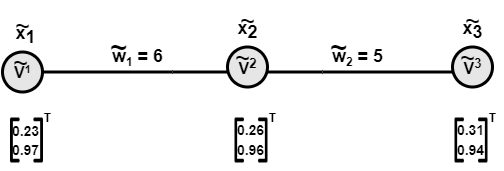}
 \caption{Coarsened graph $\mathcal{G}_c(\Theta_c, \tilde{X}1)$}
 \label{coarsened graph2}
 \end{subfigure}
 \hfill
 $$
C=
\begin{bmatrix}
 0 & 1 & 0\\
 1 & 0 & 0\\
 0 & 1 & 0\\
 0 & 1 & 0\\
 0 & 0 & 1\\ 
\end{bmatrix}
 \quad \text{,} \quad \Theta_c=
\begin{bmatrix}
 6 & -6 & 0 \\
 -6 & 11 & -5\\ 
 0 & -5 & 5 \\
\end{bmatrix}
 \quad \text{,} \quad W_c=
\begin{bmatrix}
 0 & 6 & 0 \\
 6 & 0 & 5\\ 
 0 & 5 & 0 \\
\end{bmatrix}
 \quad \text{,} \quad \tilde{X}=
\begin{bmatrix}
 0.23 & 0.97 \\
 0.26 & 0.96\\ 
 0.31 & 0.94 \\
\end{bmatrix}
$$
FGC on toy example having feature matrix $X1$
 \label{FGC toy 2}
\end{figure}
\begin{figure}
 \centering
 \begin{subfigure}[b]{0.45\textwidth}
 \centering
 \includegraphics[width=0.7\textwidth]{FGC_Original_Diagram.png}
 \caption{Original graph $\mathcal{G}(\Theta, X2)$}
 \end{subfigure}
 \hfill
 \begin{subfigure}[b]{0.45\textwidth}
 \centering
 \includegraphics[width=0.9\textwidth]{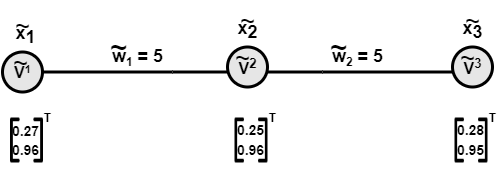}
 \caption{Coarsened graph $\mathcal{G}_c(\Theta_c, \tilde{X}2)$}
 \label{coarsened graph1}
 \end{subfigure}
 \hfill
 $$
C=
\begin{bmatrix}
 1 & 0 & 0\\
 0 & 1 & 0\\
 0 & 1 & 0\\
 1 & 0 & 0\\
 0 & 0 & 1\\ 
\end{bmatrix}
 \quad \text{,} \quad \Theta_c=
\begin{bmatrix}
 5 & -5 & 0 \\
 -5 & 10 & -5\\ 
 0 & -5 & 5 \\
\end{bmatrix}
 \quad \text{,} \quad W_c=
\begin{bmatrix}
 0 & 5 & 0 \\
 5 & 0 & 5\\ 
 0 & 5 & 0 \\
\end{bmatrix}
 \quad \text{,} \quad \tilde{X}=
\begin{bmatrix}
 0.27 & 0.96 \\
 0.25 & 0.96\\ 
 0.28 & 0.95 \\
\end{bmatrix}
$$
\label{FGC toy 3}
FGC on toy example having feature matrix $X2$.
\end{figure}

\section{Conclusion}
We introduced a novel and general framework for coarsening graph data named as Featured Graph Coarsening (FGC) which considers both the graph matrix and feature matrix jointly. In addition, for the graph data which do not have a feature matrix, we introduced the graph coarsening(GC) algorithm. Furthermore, as the graph size is reducing, it is desirable to reduce the dimension of features as well, hence we introduced FGCR algorithm. We posed FGC as a multi-block non-convex optimization problem which is an efficient algorithm developed by bringing in techniques from alternate minimization, majorization-minimization, and $\log$ determinant frameworks. The developed algorithm is provably convergent and ensures the necessary properties in the coarsen graph data like $\epsilon$-similarity and spectral similarity. Extensive experiments with both real and synthetic datasets demonstrate the superiority of the proposed FGC framework over existing state-of-the-art methods. The proposed approach for graph coarsening will be of significant interest to the graph machine learning community. 
\newpage
\section{Appendix} 
\subsection{Proof of Lemma \ref{LipshitzforC}}\label{Lipschitz}
Consider the function $-\gamma \text{log det}(C^T\Theta C +J)$. The Lipschitz constant $L_1$ of the function $-\gamma \text{log det}(C^T\Theta C +J)$ is related to the smallest non zero eigenvalue of coarsened Laplacian matrix $C^T\Theta C=\Theta_c$, which is bounded away from $\frac{\delta}{(k-1)^2}$ \cite{rajawat2017stochastic}, where $\delta$ is the minimum non zero weight of coarsened graph. However, for
practical purposes, the edges with very small weights can be ignored and set to be zero, and
we can assume that the non-zero weights of the coarsened graph $\mathcal{G}_c$ are bounded by some constant $\delta \geq 0$. On the other hand, we do not need a tight Lipschitz constant
$L_1$. In fact, any $L_1^'\geq L_1$ makes the function $g(C|C^{(t)})$
satisfy \eqref{majorizedfunction}.

Now, consider the $\text{tr}(\cdot)$ term:
\begin{align}
\begin{split}\label{eq21}
\Big|\text{tr}(\tilde{X}^{T}C1^T\Theta C1\tilde{X})-& \text{tr}(\tilde{X}^{T}C2^T\Theta C2\tilde{X}) \Big|
= \Big|\text{tr}(\tilde{X}^{T}C1^T\Theta C1\tilde{X})-\text{tr}(\tilde{X}^TC2^T\Theta C1\tilde{X})\\
&+\text{tr}(\tilde{X}^TC2^T\Theta C1\tilde{X})-\text{tr}(\tilde{X}^{T}C2^T\Theta C2\tilde{X})\Big| 
\end{split}\\
&\leq \Big|\text{tr}(\tilde{X}^T(C1-C2)^T\Theta C1\tilde{X})\Big|+ \Big|\text{tr}(\tilde{X}^TC2^T \Theta (C1-C2)\tilde{X}\Big| \label{eq22}\\
&\leq \|\text{tr}\|\|\tilde{X}^T(C1-C2)^T\Theta C1\tilde{X}\|_F+\|\text{tr}\|\|\tilde{X}^TC2^T \Theta (C1-C2)\tilde{X}\|_F \label{eq23}\\
&\leq ||\text{tr}||||\tilde{X}||_F^2||\Theta||||C1-C2||_F(||C1||_F+||C2||_F) \label{eq24}\\
&\leq L_2||C1-C2||_F \label{eq25}
\end{align}

We applied the triangle inequality after adding and subtracting $\text{tr}(\tilde{X}^TC2^T\Theta C1\tilde{X})$ in \eqref{eq21} to get \eqref{eq22}. Using the property of the norm of the trace operator i.e. $\|\text{tr}\|=\underset{A \neq 0}{{\sup}}\frac{|\text{tr(A)}|}{||A||_F}$ from $\mathbb{R}^{n \times n}$ to $\mathbb{R}$ in \eqref{eq22} to get \eqref{eq23}. Applying the Frobenius norm property i.e. $\|AB\|_F\leq \|A\|_F\|B\|_F$ in \eqref{eq23} to get $\eqref{eq24}$. Since, in each row of $C$ is having only one non zero entry i.e. 1 and rest entries are zero so, $\|C1\|_F=\|C2\|_F = \sqrt{p}$ and putting this in \eqref{eq24}, we get \eqref{eq25} where, $L_2=2\sqrt{p}\|\text{tr}\|\|\tilde{X}\|_F^2||\Theta||_F$.

Next, consider the function $\frac \alpha 2 \|C\tilde{X} - X\|^2_F$:
\begin{align}
\frac \alpha 2 \|C\tilde{X} - X\|^2_F 
&= \frac \alpha 2 \text{tr}((C\tilde{X} - X)^T(C\tilde{X} - X)) \\
&= \frac \alpha 2 \text{tr}(\tilde{X}^TC^TC\tilde{X}-X^TC\tilde{X}+X^TX-\tilde{X}^TC^TX)\\
&= \frac \alpha 2 (\text{tr}(\tilde{X}^TC^TC\tilde{X})-\text{tr}(\tilde{X}^TC^TX)- \text{tr}(X^TC\tilde{X})+\text{tr}(X^TX)) 
\end{align}
With respect to $C$, $\text{tr}(X^TX)$ is a constant and $\text{tr}(\tilde{X}^TC^TC\tilde{X}), \text{tr}(\tilde{X}^TC^TX), \text{tr}(X^TC\tilde{X})$ are Lipschitz continous function and proof is very similar to the proof of tr($\cdot$) as in \eqref{eq21}-\eqref{eq25}, and sum of Lipschitz continuous function is Lipschitz continuous so $\frac \alpha 2 \|C\tilde{X} - X\|^2_F$ is $L_3$ Lipschitz continuous.\\

Finally, consider the function $\frac \lambda 2 \|C^T\|_{1,2}^2$. Note that we have $C \geq 0$ means all the elements of $C$ are non-negative, $|C|_{ij}=C_{ij}\geq 0$. With this the $\ell_{1}$-norm becomes summation, and we obtain the following: 
\begin{align}
\|C^T\|_{1,2}^2
 &=\sum_{i=1}^p\big(\sum_{j=1}^kC_{ij}\big)^2 \label{eqC}\\
 &=\sum_{i=1}^p([C^T]_{i}\textbf{1})^2\\
 &=\|C\textbf{1}\|_F^2\\
 &=\text{tr}(\textbf{1}^TC^TC\textbf{1})
\end{align}
where $\textbf{1}$ is a vector having all entry 1, $[C^T]_i$ is $i$-th row of loading matrix C and since each entry of C is $C_{ij}\geq 0$. $\text{tr}(\textbf{1}^TC^TC\textbf{1})$ is Lipschitz continuous function and proof is similar to proof of tr($\cdot$) as in \eqref{eq21}-\eqref{eq25} so $\|C^T\|_{1,2}^2$ is $L$-4 Lipschitz continuous function.\\
Addition of Lipschitz continuous functions is Lipschitz continuous so we can say that $f(C)$ in \eqref{UpdateC123} is $L$- Lipschitz continuous function where $L=\text{max}(L_1, L_2, L_3, L_4)$.
 
\subsection{Proof of Lemma \ref{KKTcondition}} \label{LagrangianKKT}
The Lagrangian function of \eqref{eqn:C} is:
\begin{align}
L(C, \tilde{X}, \bm \mu_1)= & \frac{1}{2}C^TC-C^TA - \bm \mu_1^\top C + \bm \mu_2^T\Big[ \|C_1^T\|_2^2 \quad \|C_2^T\|_2^2 \ldots \|C_p^T\|_2^2 \Big]^T
\end{align}
where $\bm \mu_1$ is the dual variable. The KKT conditions of \eqref{eqn:C} is
\begin{align}
C-A-\bm \mu_1+2\Big[\mu_{21}C^T_1, \ldots \mu_{2p}C^T_p]^T=0, \label{107} \\
\bm \mu_2^T\Big[ \|C_1^T\|_2^2 \quad \|C_2^T\|_2^2 \ldots \|C_p^T\|_2^2 \Big]^T =0, \\
\bm \mu_1^\top C =0,\label{108}\\
\quad C \geq 0,\label{109}\\
\quad \bm \mu_1 \geq 0 \label{110} \\
\|[C^T]_i\|_2^2\leq 1\\
\quad \bm \mu_2 \geq 0 \label{199}
\end{align}
The optimal solution of $C$ that satisfies all KKT conditions \eqref{107}-\eqref{199} is
\begin{align}
C^{t+1}
&=\frac{(A)^{+}}{\|[A^T]_i\|_2}
\end{align}
where $A=\left(C^{(t)} - \frac{1}{L}\nabla f\left(C^{(t)}\right)\right)^+$ and $\|[A^T]_i\|$ is the $i$-th row of matrix A.
This concludes the proof.

\subsection{Proof of Theorem \ref{convergence}}\label{convergence proof}
We show that each limit point $(C^{t},\tilde{X}^{t})$ satisfies KKT condition for \eqref{Main formulation FGC}. Let $(C^\infty, \tilde{X}^\infty)$ be a limit point of the generated sequence.\\
The Lagrangian function of \eqref{Main formulation FGC} is

\begin{dmath}
L(C, \tilde{X},\bm \mu)= -\gamma \text{log det}(C^T\Theta C +J)+\text{tr}(\tilde{X}^{T}C^T\Theta C\tilde{X}) +\frac{\alpha}{2}||X-C\tilde X||_F^2 +\frac \lambda 2 \sum \limits_{i=1}^p \|[C^T]_i\|_1^2 - \bm \mu_1^\top C + \bm \mu_2^T\Big[ \|C_1^T\|_2^2 \quad \|C_2^T\|_2^2 \ldots \|C_p^T\|_2^2 \Big]^T
\end{dmath}

where $\bm \mu_1$ and $\bm \mu_2$ are the dual variables.\\ 
$(1)$ The KKT condition with respect to $C$ is

\begin{align}
\hspace{-3cm}-2\gamma \Theta C(C^T\Theta C+J)^{-1}+\alpha \left(C\tilde{X} - X\right)\tilde{X}^{\intercal} +2\Theta C\tilde{X}\tilde{X}^T+ \lambda C\pmb{1}_{k \times k}- \bm \mu_1 + 2\Big[\mu_{21}C^T_1, \ldots \mu_{2p}C^T_p]^T=0, \label{m1} \\
\bm \mu_2^T\Big[ \|C_1^T\|_2^2 \quad \|C_2^T\|_2^2 \ldots \|C_p^T\|_2^2 \Big]^T =0,\\
\bm \mu_1^\top C =0, \\ \bm \mu_1 \geq 0, \\ \quad C \geq 0, \\ \bm \mu_2 \geq 0,\\ \|[C^T]_i\|_2^2\leq 1
\end{align}
where $\textbf{1}_{k\times k}$ is a $k \times k$ matrix whose all entry is one. ${C}$ is derived by using KKT condition from \eqref{eqn:C}:
\begin{equation}\label{eqns:C}
\hspace{-2cm}C^{\infty}-C^{\infty} + \frac{1}{L}\left(-2\gamma \Theta C^\infty(C^\infty)^{T} \Theta C^{\infty}+J)^{-1}+\alpha (C^{\infty}\tilde{X}^{\infty} - X)(\tilde{X}^{\infty})^{T} +2\Theta C^{\infty}\tilde{X}^{\infty}(\tilde{X}^{\infty})^{T} + \lambda C^{\infty}\pmb{1}_{k \times k}\right)=0
\end{equation}

\begin{equation}\label{eqns1:C}
\hspace{-2cm}-2\gamma \Theta C^\infty ((C^\infty)^{T} \Theta C^\infty+J)^{-1}+\alpha (C^{\infty}\tilde{X}^{\infty} - X)(\tilde{X}^{\infty})^{T} +2\Theta C^\infty\tilde{X}^\infty(\tilde{X}^\infty)^T+ \lambda C^\infty\pmb{1}_{k \times k}=0
\end{equation}
Therefore scaling $\bm \mu_1=0$ and $\bm \mu_2=0$ , we can conclude that $C^\infty$ satisfies KKT condition.\\
$(2)$ The KKT condition with respect to $\tilde{X}$ is
\[2C^T\Theta C \tilde{X} + \alpha C^T(C\tilde{X} - X) = 0\]
This concludes the proof.

\subsection{Proof of Theorem \ref{epsilon-thm}}\label{Dritchletenergy}
We have $\|X\|_{\Theta}=\sqrt{\text{tr}(X^T\Theta X)}$ and $\|\tilde{X}\|_{\Theta}=\sqrt{\text{tr}(\tilde{X}^T\Theta_c \tilde{X})}$. Taking the absolute difference between $\|X\|_{\Theta}$ and $\|\tilde{X}\|_{\Theta_c}$, we get:
\begin{align}\label{eq-1}
\Big|\|X\|_\Theta-\|\tilde{X}\|_{\Theta_c}\Big|
&= \Big|\sqrt{\text{tr}(X^T\Theta X)}- \sqrt{\text{tr}(\tilde{X}^T\Theta_c \tilde{X})}\Big| 
\end{align}

As $\Theta$ is a positive semi-definite matrix using Cholesky's decomposition $\Theta=S^TS$ in \eqref{eq-1}, we get the following inequality:
\begin{align}
\Big|\|X\|_\Theta-\|\tilde{X}\|_{\Theta_c}\Big|
&= \Big|\sqrt{\tr{X^T\Theta X}}- \sqrt{\tr{\tilde{X}^T\Theta_c \tilde{X}}}\ \Big|\\
&= \Big|\sqrt{\tr{X^TS^TS X}}- \sqrt{\tr{\tilde{X}^T C^TS^TSC \tilde{X}}}\Big|\\
&=\Big| \|SX\|_F-\|SP^{\dagger}PX\|_F\Big| \\
&\leq \|SX-SP^{\dagger}PX\|_F \\
&\leq \epsilon\|X\|_{\Theta}\label{less than}
\end{align}

From the optimality condition of the optimization problem and the update of $\tilde{X}$ in \eqref{updatetildeX} we have the following inequality:
$$\|\tilde{X}\|_{\Theta_c} \leq \|X\|_\Theta,$$
Using this in \eqref{less than} we get 
\begin{equation}\label{less}
\frac{\Big|\|X\|_\Theta-\|\tilde{X}\|_{\Theta_c}\Big|}{\|X\|_\Theta}\leq 1
\end{equation}
The equation \eqref{less than} and \eqref{less} implies that the range of $\epsilon\in[0,1].$ Next, by applying the property of the modulus function in \eqref{less than}, we obtain the following inequality for all the $n$ samples: 
\begin{equation}{\label{rsp}}
(1-\epsilon)\|X\|_{\Theta} \leq \|\tilde{X}\|_{\Theta_c} \leq (1+\epsilon)\|X\|_{\Theta}
\end{equation}
where $\epsilon\in[0,1] $ and this concludes the proof.

\bibliographystyle{unsrt}
\bibliography{references}

\end{document}